\documentclass{article}

\usepackage{arxiv}
\usepackage[utf8]{inputenc} 
\usepackage[T1]{fontenc}    
\usepackage{hyperref}       
\usepackage{url}            
\usepackage{booktabs}       
\usepackage{amsfonts}       
\usepackage{amsmath}
\newcommand{\R}{\ensuremath{\mathbb{R}}}
\usepackage{nicefrac}       
\usepackage{microtype}      
\usepackage{lipsum}		
\usepackage{graphicx}
\usepackage{doi}
\usepackage{authblk}
\usepackage{soul,color}

\usepackage{nicematrix}
\usepackage[flushleft]{threeparttable}

\NiceMatrixOptions{cell-space-top-limit=0.1pt,cell-space-bottom-limit=0.1pt}

\usepackage[backend=biber,sorting=none,sortcites]{biblatex}
\title{Pediatric brain tumor classification using digital histopathology and deep learning: evaluation of SOTA methods on a multi-center Swedish cohort}

\author[1,2,*]{Iulian Emil Tampu}
\author[2,3]{Per Nyman}
\author[1,2]{Christoforos Spyretos}
\author[2,4]{Ida Blystad}
\author[5,6]{Alia Shamikh}
\author[5]{Gabriela Prochazka}
\author[5,6]{Teresita Díaz de Ståhl}
\author[5,6]{Johanna Sandgren}
\author[2,4,7]{Peter Lundberg}
\author[1,2]{Neda Haj-Hosseini}

\affil[1]{Department of Biomedical Engineering, Linköping University, Sweden}
\affil[2]{Center for Medical Image Science and Visualization, Linköping University, Sweden}
\affil[3]{Crown Princess Victoria Children’s Hospital and Department of Health, Medicine and Caring Sciences, Linköping University, Sweden}
\affil[4]{Department of Radiology and Department of Health, Medicine and Caring Sciences, Linköping University, Sweden}
\affil[5]{Department of Oncology-Pathology, Karolinska Institutet, Solna, Sweden}
\affil[6]{Department of Clinical Pathology and Cancer Diagnostics, Karolinska University Hospital, Sweden}
\affil[7]{Department of Radiation Physics and Department of Medical and Health Sciences, Linköping University, Sweden}
\affil[ ]{ }
\affil[*]{Corresponding author: \texttt{iulian.emil.tampu@liu.se}}

\usepackage{setspace}
\makeatletter
\let\oldaffillist\AB@affillist
\renewcommand{\AB@affillist}{\begin{flushleft}\begin{spacing}{1.2}\oldaffillist\end{spacing}\end{flushleft}}
\makeatother

\date{}

\renewcommand{\shorttitle}{Pediatric brain tumor classification using digital histopathology and DL foundation models}

\hypersetup{
pdftitle={Pediatric brain tumor classification using digital histopathology and deep learning: evaluation of SOTA methods on a multi-center Swedish cohort},
pdfsubject={q-bio.NC, q-bio.QM},
pdfauthor={Iulian E. Tampu},
pdfkeywords={},
}

\begin{document}
\maketitle

\begin{abstract}
Brain tumors are the most common solid tumors in children and young adults, but the scarcity of large histopathology datasets has limited the application of computational pathology in this group. This study implements two weakly supervised multiple-instance learning (MIL) approaches on patch-features obtained from state-of-the-art histology-specific foundation models to classify pediatric brain tumors in hematoxylin and eosin whole slide images (WSIs) from a multi-center Swedish cohort. WSIs from 540 subjects (age 8.5$\pm$4.9 years) diagnosed with brain tumor were gathered from the six Swedish university hospitals. Instance (patch)-level features were obtained from WSIs using three pre-trained feature extractors: ResNet50, UNI, and CONCH. Instances were aggregated using attention-based MIL (ABMIL) or clustering-constrained attention MIL (CLAM) for patient-level classification. Models were evaluated on three classification tasks based on the hierarchical classification of pediatric brain tumors: tumor category, family, and type. Model generalization was assessed by training on data from two of the centers and testing on data from four other centers. Model interpretability was evaluated through attention mapping. The highest classification performance was achieved using UNI features and ABMIL aggregation, with Matthew’s correlation coefficient of 0.76$\pm$0.04, 0.63$\pm$0.04, and 0.60$\pm$0.05 for tumor category, family, and type classification, respectively. When evaluating generalization, models utilizing UNI and CONCH features outperformed those using ResNet50. However, the drop in performance from the in-site to out-of-site testing was similar across feature extractors. These results show the potential of state-of-the-art computational pathology methods in diagnosing pediatric brain tumors at different hierarchical levels with fair generalizability on a multi-center national dataset.
\end{abstract}


\section*{Introduction}
\vspace*{-0.3cm}
Tumors of the central nervous system (CNS) are the most common solid neoplasms occurring in children and young adults (age <20 years), and account for 20\% of all pediatric tumors \cite{adel_fahmideh_pediatric_2021} with a global incidence of 1.2 per 100.000 children in 2022 \cite{erdmann_childhood_2021}. The prognosis of pediatric CNS tumors depends on several factors including tumor location, histology, age, and sex, with a 10-year survival of 72\% in western countries \cite{adel_fahmideh_pediatric_2021}. Timely and accurate diagnosis of pediatric brain tumors is essential to improve prognosis and reduce the collateral and long-term effects of treatment \cite{erdmann_childhood_2021}. Non-invasive imaging modalities, such as computed tomography (CT) and magnetic resonance imaging (MRI), are used for the initial diagnosis and to guide biopsies for histological and molecular analysis. A final diagnosis is attained by integrating radiological findings, histological examination, and molecular profiling, with the latter playing an increasingly larger role as genetic and epigenetic traits drive disease progression \cite{cole_neuropathology_2022,viaene_pediatric_2023}. Advancements in molecular analysis have enabled a refined diagnosis of pediatric brain tumors, identifying 22 distinct types \cite{cohen_alan_r_brain_2022,louis_2021_2021}. However, molecular profiling is costly, time-consuming, and not widely available today. Consequently, treatment sometimes begins after the preliminary diagnosis based on radiological and histological features. Histological examinations are labor-intensive, and in the case of pediatric brain tumors, the lack of expert pediatric neuropathologists (as low as one expert per 173.000 children \cite{craiu_training_2020} in Europe), emphasizes the need for decision-support tools to aid diagnosis. 
The digitization of histology glass slides into whole slide images (WSIs) has enabled the development of computational pathology (CPath) methods that automatically analyze the image data (with or without the combination of clinical metadata) for tissue sub-region segmentation, prognostication, cancer subtyping, and gene mutation prediction \cite{abels_computational_2019}. Currently, CPath takes advantage of artificial intelligence (AI) and deep learning algorithms (DL), with several studies showing the successful implementation of these methods for cell and gland detection/segmentation and classification \cite{salvi_impact_2021,niazi_digital_2019, bera_artificial_2019} and more recently for molecular biomarker detection from the image data \cite{echle_deep_2021}. 
In the context of slide and patient-level prediction tasks (tumor grading, subtyping, and prognostication), the state-of-the-art (SOTA) DL methods use variants of the multiple instance learning (MIL) framework \cite{maron_framework_1997,campanella_clinical-grade_2019} to address the unavailability of pixel-level annotations and the computational challenges that originate from the large size of the WSIs. This framework is composed of several customizable steps including segmentation of the tissue regions from the glass background, subdivision of the tissue region into non-overlapping patches, projection of each patch using pre-trained patch-encoders into a low-dimensional space and aggregation of the patch representations into a slide-level representation which can be used for the downstream task. A large corpus of research has been focused on obtaining histology-specific patch encoders that provide strong feature representations, that are task-agnostic and can be applied in low-data regime settings \cite{chen_scaling_2022, wang_transformer-based_2022, chen_towards_2024, lu_visual-language_2024}. Moreover, several aggregation methods have been developed \cite{gadermayr_multiple_2024}, with attention-based MIL (ABMIL) \cite{ilse_attention-based_2018} and its variants becoming the method of choice for CPath.

CPath methods for brain tumors have primarily focused on the adult population \cite{redlich_applications_2024}, with few studies exclusively investigating pediatric tumors for diagnosis or survival prediction \cite{bengs_medulloblastoma_2021, whitney_quantitative_2022, attallah_ai-based_2022, steyaert_multimodal_2023} mainly due to the unavailability of large datasets. Among these, Steyaer et at. proposed a deep learning-based approach for survival prediction of pediatric brain tumors (low-grade glioma, high-grade astrocytoma, high-grade ependymoma, and high-grade medulloblastoma) and adult (low-grade glioma and glioblastoma) using H\&E WSIs and genetic data \cite{steyaert_multimodal_2023}. The authors obtained a slide-level representation by averaging the patch features extracted using an ImageNet pre-trained ResNet50 \cite{he_deep_2016} model. Results show that H\&E WSIs alone could predict the survival of pediatric brain tumors with a composite score of (0.67$\pm$0.16) which improved by 6.5\% when fused with RNA data. The subtyping of medulloblastoma, the most common malignant brain tumor in the pediatric population, has also been investigated \cite{bengs_medulloblastoma_2021, attallah_ai-based_2022}. By using a pre-trained EfficientNet model fine-tuned on squared crops of medulloblastoma WSIs, Bengs et al., showed a classification F1 score of 80.1\% between classic and nodular-type medulloblastoma \cite{bengs_medulloblastoma_2021}. In another work, Whitney and colleagues first segmented the cells in squared WSI patches and then extracted nuclear and histomorphometric features describing the shape, architecture, and texture of the cell population \cite{whitney_quantitative_2022}. A machine learning model was then trained on the extracted features for the classification between SHH-activated, WNT-activated, and Group 3, 4 medulloblastoma subtypes obtaining a patient-level area under the receiver operator curve (AUROC) of 0.70. Additionally, survival prediction within each tumor subtype was performed, with the highest AUC of 0.92 obtained for medulloblastoma Group 3 survival prediction. Of note, both \cite{bengs_medulloblastoma_2021, whitney_quantitative_2022} extracted patches manually from the WSI to ensure that only tumor regions were available for feature extraction and training. Thus, both methods suffer from the need for fine detailed annotations and cannot be trained in an end-to-end fashion. 

In this work, we aim to implement and evaluate state-of-the-art weakly supervised DL methods on WSIs of pediatric brain tumors for the classification of tumor category, family, and type on a unique multi-center dataset that represents a population-based range of diagnosis. The contributions of this work are: (1) implement and evaluate two histopathology-specific feature extractors (UNI and CONCH) for the classification of pediatric brain tumors at patient level, and compare them with a baseline ImageNet pre-trained ResNet50, (2) evaluate and compare two established attention-based approaches (ABMIL) and clustering-constrained-attention MIL (CLAM) for the aggregation of the patch-level features into a patient level classification and (3) evaluate model generalization by training on data from two centers and testing on data from four other centers

\section*{Materials and Methods}
\subsection*{Dataset}
The dataset used in this study was made available from the Swedish Childhood Tumor Biobank (BTB), and is part of a national effort to generate and maintain multi-modal data on pediatric brain tumors (among other childhood solid malignancies) from the six Swedish university hospitals where tissue samples and linked sample information are collected to BTB after informed consent. Ethical approval was obtained from the Swedish Ethical Review Authority (Dnr 2021–03985 and Dnr 2022-00065-02). The study was additionally approved by BTB, as well as Karolinska University Hospital and Stockholm Medical Biobank, the medico-legal authority for BTB's personal data and tissue samples, respectively. 

Data included subjects from 2013 to 2023, and although it does not include all pediatric brain tumor patients, the selection is random and representative of the prevalent diagnoses in the country. Data from three centers (Linköping, Uppsala, and Umeå) were obtained as H\&E stained WSIs, while from the remaining centers (Stockholm, Gothenburg, and Lund) glass slides were retrieved and scanned using digital scanners (Hamamatsu-Nanozoomer-XR \& S360). In total, 1449 WSIs from 540 subjects (284 males, 253 females, three not specified, age in years 8.29$\pm$5.03, range [0.00, 20.00]) were available representing primary, metastatic, and recurrent tumors. Patient diagnoses were retrieved from the clinical health records, specifying tumor category, family, and type. Eleven subjects had multiple diagnoses corresponding to analysis of samples from metastatic and/or recurrent tumor regions (11 subjects with double diagnosis). Thus, 564 unique subject-diagnosis pairs, hereafter called cases, were identified and used for the analysis. Overall, the WSIs were scanned with five different scanners at magnifications of $\times$20 or $\times$40 resulting in a variability in image quality and color accuracy, among others. A total of 62 cases and 128 WSI were excluded. Of these, three WSIs were excluded since missing clinical information, two WSIs were removed since there was not enough material for diagnosis (one case and one female), 32 WSIs were excluded since diagnosis was either not a tumor or inflammation (19 cases, 11 males and eight females), five WSIs were removed since the diagnosis was of a non malignant tumor (two cases, two females), 14 WSIs were excludes since related to diagnosis of tumors outside the CNS (nine cases, six males and three females) and 22 WSIs were removed since diagnoses could not be matched to the latest 2021 WHO classification of brain tumors \cite{louis_2021_2021}. Additionally, 44 WSIs (from 10 cases, three males and 7 females) were removed due to severe scanning artifacts (e.g., out-of-focus scan). To ensure a sufficient number of cases for training and testing, only tumor diagnoses represented by at least 10 cases were included in the analysis, with the threshold of 10 set according to the few-shot results presented in \cite{chen_towards_2024}. Table \ref{tab:dataset_description} summarizes the dataset used for model training and testing, detailing the counts of cases and WSIs available for different tumor categories, families, and types. Additionally, Figures \ref{fig:tissue_area_stats_tumor_category}, \ref{fig:tissue_area_stats_tumor_family} and \ref{fig:tissue_area_stats_tumor_type} show the tissue area distribution for all the classes included in the analysis.

\begin{table}[ht]
\small
 \centering
 \caption{Summary of subjects and whole slide images (WSIs) used for model development and evaluation. Numbers reflect the remaining cases and WSIs after dataset curation.}\label{tab:dataset_description}
    \begin{NiceTabular}{m[c]{31mm} m[c]{15mm} m[c]{31mm} m[c]{15mm} m[c]{31mm} m[c]{15mm}}[]

    \toprule
    
    \Block{1-2}{\textbf{Tumor category}} & & \Block{1-2}{\textbf{Tumor Family}} & & \Block{1-2}{\textbf{Tumor Type}} \\
    
    \midrule
    
    \Block{1-1}{\texttt{class}} & \Block{1-1}{\texttt{Cases\\(slides)}} & \Block{1-1}{\texttt{class}} & \Block{1-1}{\texttt{Cases\\(slides)}} & \Block{1-1}{\texttt{class}} & \Block{1-1}{\texttt{Cases\\(slides)}} \\

    \midrule
    
    \Block{7-1}{Gliomas/glioneuronal\\/neuronal tumors} & \Block{7-1}{311\\(819)} & \Block{1-1}{Circumscribed astrocytic glioma} & \Block{1-1}{175\\(434)} & \Block{1-1}{Pilocytic astrocytoma} & \Block{1-1}{166\\(402)}  \\

    \cmidrule{3-6}
    
    & & \Block{2-1}{Glioneuronal and neuronal tumors} & \Block{2-1}{55\\(148)} & \Block{1-1}{Ganglioglioma} & \Block{1-1}{32\\(93)} \\
    
    \cmidrule{5-6}
    
    & & & & \Block{1-1}{Dysembryoplastic neuroepithelial tumor (DNET)} & \Block{1-1}{15\\(39)} \\

    \cmidrule{3-6}

    & & \Block{2-1}{Ependymal tumors} & \Block{2-1}{35\\(101)} & \Block{1-1}{Ependymoma grade 3} & \Block{1-1}{21\\(56)} \\

    \cmidrule{5-6}

    & & & & \Block{1-1}{Ependymoma grade 1-2} & \Block{1-1}{14\\(45)} \\

    \cmidrule{3-6}

    & & \Block{1-1}{Adult-type diffuse gliomas} & \Block{1-1}{20\\(65)} & \Block{1-1}{Glioblastoma} & \Block{1-1}{16\\(51)} \\
    
    \cmidrule{3-6}
    
    & & \Block{1-1}{Pediatric-type diffuse high-grade glioma} & \Block{1-1}{11\\(27)} & \Block{1-1}{/} & \Block{1-1}{/} \\
    
    \midrule

    \Block{3-1}{Embryonal tumors} & \Block{5-1}{118\\(330)} & \Block{2-1}{Medulloblastoma} & \Block{2-1}{84\\(238)} & \Block{1-1}{Medulloblastoma\\non-WNT/non-SHH} & \Block{1-1}{37\\(104)} \\

    \cmidrule{5-6}
    
    & & & & \Block{1-1}{Medulloblastoma\\WNT activated} & \Block{1-1}{10\\(25)} \\

    \cmidrule{3-6}

    & & \Block{1-1}{Other CNS embryonal tumors} & \Block{1-1}{26\\(77)} & \Block{1-1}{AT/RT} & \Block{1-1}{17\\(39)} \\

    \cmidrule{3-6}

    & & \Block{1-1}{Embryonal tumors NOS} & \Block{1-1}{10\\(15)} & \Block{1-1}{/} & \Block{1-1}{/} \\

    \midrule

    \Block{1-1}{Tumors of the sellar region} & \Block{1-1}{31\\(65)} & \Block{1-1}{Adamantinomatous craniopharyngioma} & \Block{1-1}{16\\(34)} & \Block{1-1}{/} & \Block{1-1}{/} \\

    \midrule

    \Block{1-1}{Meningiomas} & \Block{1-1}{11\\(32)} & \Block{1-1}{Meningioma} & \Block{1-1}{11\\(32)} & \Block{1-1}{/} & \Block{1-1}{/} \\

    \midrule

    \Block{1-1}{Germ cell tumors} & \Block{1-1}{11\\(30)} & \Block{1-1}{/} & \Block{1-1}{/} & \Block{1-1}{/} & \Block{1-1}{/} \\

    \midrule

    \Block{1-1}{Choroid plexus tumors} & \Block{1-1}{10\\(24)} & \Block{1-1}{/} & \Block{1-1}{/} & \Block{1-1}{/} & \Block{1-1}{/} \\

    \midrule
    
    \Block{1-1}{Mesenchymal, non-meningothelial tumors} & \Block{1-1}{10\\(21)} & \Block{1-1}{/} & \Block{1-1}{/} & \Block{1-1}{/} & \Block{1-1}{/} \\

    \midrule\midrule
    
    \Block{1-1}{\textbf{Total}} & \Block{1-1}{\textbf{502\\(1321)}} & \Block{1-1}{} & \Block{1-1}{\textbf{443\\(1171)}} & \Block{1-1}{} & \Block{1-1}{\textbf{328\\(854)}} \\

    \bottomrule
    \end{NiceTabular}
\end{table}

\begin{figure}[ht]
    \centering
        \includegraphics[width=\textwidth]{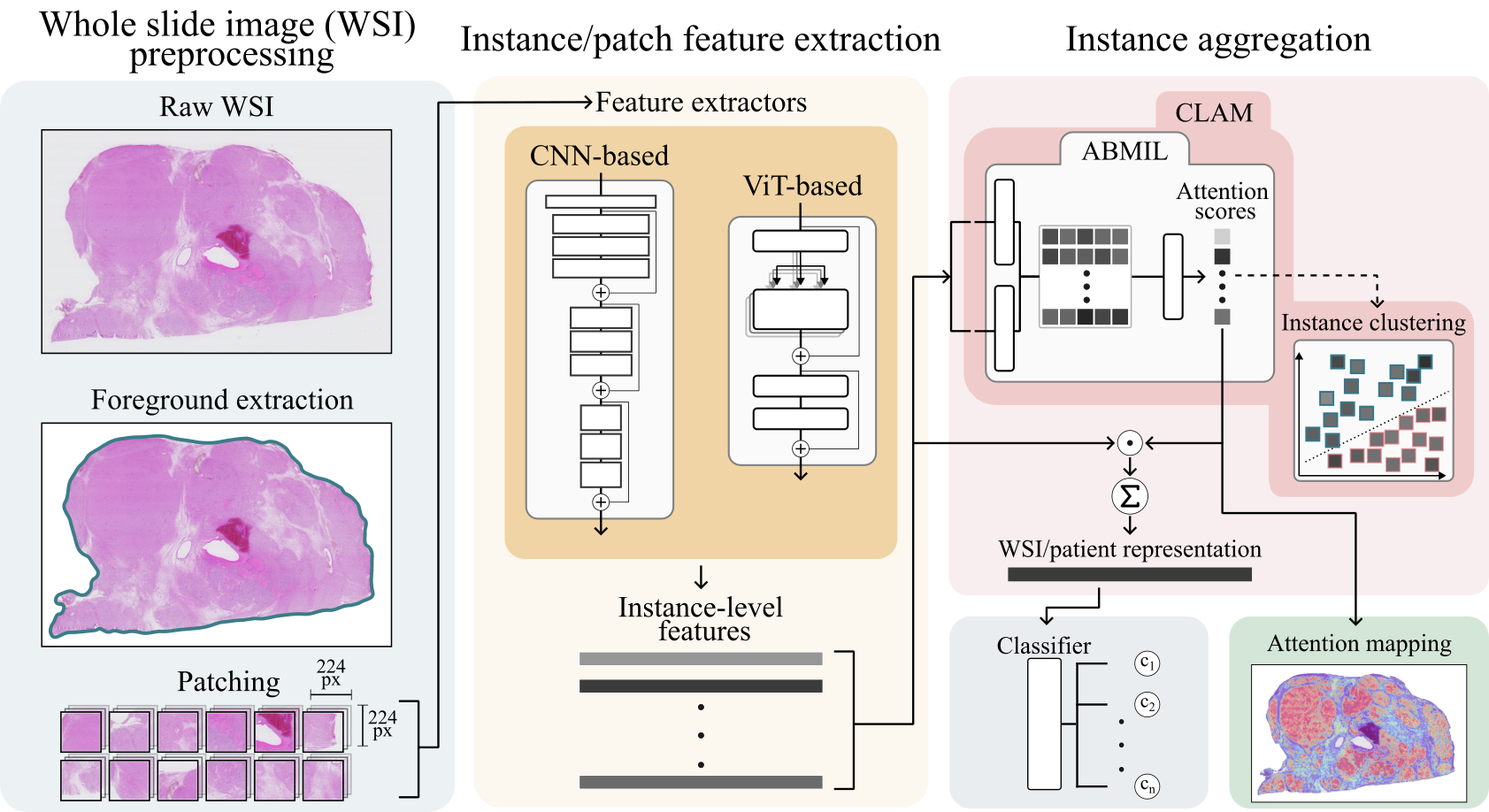}
    \caption{Overview of the whole slide image (WSI) preprocessing, classification and attention map generation. CNN: convolutional neural networks (in this work the ResNet50), ViT: vision transformer (in this work UNI and CONCH), ABMIL: attention-based multiple instance learning, CLAM: clustering-constrained attention multiple instance learning. }
    \label{fig:method}
\end{figure}

\subsection*{Deep-learning}
In this work, the weakly-supervised MIL framework was utilized to train models for patient-level prediction (Figure \ref{fig:method}). Initially, non-overlapping patches of size 224$\times$224 pixels at $\times$20 magnification were obtained from tissue-segmented regions of each WSI using the CLAM framework \cite{lu_data-efficient_2021}. Features from each patch (or instance) were then extracted using one of three pre-trained image encoders: ResNet50, UNI, and CONCH (see Instance-level feature extraction section). The instance-level features of all the WSIs of each patient were then aggregated into a single representation using two different MIL-based methods: ABMIL [20] and clustering-constrained-attention MIL (CLAM) \cite{lu_data-efficient_2021} (see \textit{Attention-based aggregation methods} section). The aggregated representation was then utilized for patient-level diagnosis via a fully connected layer with softmax activation.  

\subsubsection*{Instance-level feature extraction}
Tumors often exhibit heterogeneous patterns, meaning different regions of the tumor can have varying characteristics. Patch-level (or instance-level) feature extraction allows for a detailed analysis of these diverse regions, which can provide critical insights for accurate diagnosis and prognosis. Instance-level feature extraction was performed to obtain a compact representation of each 224$\times$224 pixel patch. This approach is advantageous in computational pathology since using features as input for end-to-end training of a patient or slide-level classification model has substantially lower computational requirements (both hardware and time) compared to using the raw pixel information. Thus, the choice of the feature extractor becomes crucial since the subsequent classification model does not have access to the raw pixel intensities. In this study, we investigated three different feature extractors, namely: (1) ResNet50 \cite{he_deep_2016}, a deep convolutional neural network pre-trained on ImageNet \cite{krizhevsky_imagenet_2017} that is used in several MIL-based approaches in CPath \cite{lu_data-efficient_2021}, (2) UNI \cite{chen_towards_2024}, a vision transformer (ViT) foundation model pre-trained on more than 100M H\&E histology patches from WSIs across 20 major tissue types using the self-supervised DINOv2 framework \cite{oquab_dinov2_2024} and (3) CONCH \cite{lu_visual-language_2024}, a vision-language foundation model pre-trained on more than 1.17M H\&E histology image-caption pairs. The three pre-trained feature extractors were implemented in the CLAM framework, and features were extracted from identical patch coordinates. The size of the feature encoding produced by the pretrained encoders was 1024 for ResNet50 and ViT\_UNI, and 512 for ViT\_CONCH. Additionally, the features obtained from all WSIs for each case were concatenated, obtaining a per-case bag of features on which the attention-based aggregation was applied. Note that this approach is equivalent to first stitching all the WSIs of one case into a single image and then performing patch and feature extraction.

\subsubsection*{Attention-based aggregation methods}
In tumor slides, not all regions are equally important. For example, some patches may contain benign tissue, while others may show aggressive cancer. Attention-based aggregation allows the model to focus more on the most relevant parts of the image, possibly leading to more accurate and clinically useful predictions. In this work, two attention-based aggregation methods were investigated to derive a case-level representation from the patch-level features: the well-established ABMIL and its variation, CLAM. ABMIL \cite{ilse_attention-based_2018} was introduced as a learnable pooling method, in contrast to max or average pooling, which borrows from the attention mechanism but is invariant to the number and order of the instances in a bag. It employs two fully connected layers to learn a weight for each of the instances in the bag, which is then used to compute the bag representation as the weighted average of the instances. The gated version of the attention mechanism \cite{ilse_attention-based_2018} was used in this work, which is described by
\begin{equation}
    a_k = \frac{\text{exp} \left( w\top \left( \left(\text{tanh} V h_k^\top \right) \bigodot \text{sign}\left( Uh_k^\top\right)\right) \right)}{\sum^K_{j=1}\text{exp} \left( w\top\left( \text{tanh}\left( Vh_k^\top \right) \bigodot \text{sign}\left( Uh_k^\top\right) \right) \right)}
\end{equation}
where $h_k$ is the projected instance feature vector $\in \R^{1 \times 255}$, and $w \in \R^{384 \times 1}$, $V \in \R^{381 \times 512}$, $U \in \R^{381 \times 512}$ are model parameters updated during training. The aggregated representation is then computed by
\begin{equation}
    z = \sum^K_{j=1} a_j h_j
\end{equation}
The second aggregation method, clustering-constrained attention multiple-instance learning
(CLAM) \cite{lu_data-efficient_2021}, is one of several attention MIL approaches available in the literature \cite{chen_benchmarking_2024} and used in several CPath implementations. CLAM builds upon ABMIL by adding a clustering layer parallel to the attention mechanism that is trained to differentiate between instances that are positive and negative evidence for the ground truth label. Namely, the instance-level clustering loss guides the model to attend more to instances that are positive evidence exclusively for the ground truth and not for the other labels. In this work, the k=8 instances with the highest and lowest attention scores were used for clustering \cite{lu_data-efficient_2021}. The single-branch and small-size version of CLAM was used in this work. 

\subsection*{Experimental setup}
\subsubsection*{\textit{Classification tasks}}
Utilizing the hierarchical classification system of CNS tumors, three classification tasks were designed to assess the model's performance at varying levels of classification granularity. In particular, at a coarse level, models were tasked to classify between seven tumor categories (choroid plexus tumors, embryonal tumors, gliomas/glioneuronal/neuronal tumors, meningiomas, sellar region tumors, germ cell tumors, and mesenchymal, non-meningothelial tumors). At a medium level, models were trained to distinguish between 10 tumor families (adamantinomatous craniopharyngioma, adult-type diffuse gliomas, circumscribed astrocytic gliomas, ependymal tumors, embryonal tumors not otherwise specified (NOS), glioneuronal and neuronal tumors, medulloblastoma, meningioma, other CNS embryonal tumors and pediatric-type diffuse high-grade gliomas). Finally, at the fine level, models were tasked to classify between nine classes (AT/RT, dysembryoplastic neuroepithelial tumor (DNET), ependymoma grade 3, ependymoma grade 1-2, ganglioglioma, glioblastoma, medulloblastoma WNT-activated, medulloblastoma non-WNT/non-SHH and pilocytic astrocytoma). Table 1 shows the number of subjects and WSIs for each of the classes. Classification performance for the three tasks was evaluated both when using the data from all the sites for model training and when investigating model generalization (see \textit{Model generalization}). Given the imbalance in the number of cases between the tumor entities used in the analysis, additional experiments with a balanced number of cases per class (n=10) and balanced tissue area were performed.

\subsubsection*{\textit{Model generalization}}
To evaluate the generalization of the classification models, data from two of the six sites (Stockholm and Uppsala, referred to as in-site data) were used for training, while the data from the remaining sites were used for testing (referred to as out-of-site data). The selection for the training and testing sites was guided by the number of cases, trying to obtain an equal number of cumulative cases in the training and testing sites. However, due to the smaller number of cases in the training set and the minimum requirement of 10 samples per class, the number of classes for each of the classification tasks was adjusted as follows: three classes for tumor category (embryonal tumors, gliomas/glioneuronal/neuronal tumors, tumors of the sellar region), five classes for tumor family (circumscribed astrocytic gliomas, ependymal tumors, glioneuronal and neuronal tumors, medulloblastoma, other CNS embryonal tumors), and four classes for tumor type (AT/RT, ganglioglioma, medulloblastoma non-WNT/non-SHH and pilocytic astrocytoma). Additional experiments with a minimum requirement of samples per class set to 8 for a larger inclusion are also performed. 

\subsubsection*{\textit{Training routine}}
The weights and biases of the attention-based aggregation layers and classification layer were randomly initialized and trained end-to-end using a batch size of one. For ABMIL, training was supervised using the cross-entropy loss computed between the model prediction and the case-level ground truth. In the case of CLAM, training was supervised by both the cross-entropy loss and the smooth-SVM loss obtained from the instance clustering. To account for class imbalance, a batch weighted sampling strategy was employed, where the probability of a sample being drawn is inversely proportional to the frequency of the number of samples in its class. For regularization, dropout was used on the input embeddings and after each intermediate layer in the model, with \textit{p}=0.1 and \textit{p}=0.25, respectively. Models were trained using the AdamW optimizer \cite{loshchilov_decoupled_2019} with a starting learning rate of 0.0001 and decreased during training using a cosine annealing scheduler. Training ran for a minimum of 10 and a maximum of 20 epochs along with early stopping on the validation loss with patience set to five epochs. Model training and evaluation ran on a workstation equipped with a 24 GB 4090 Nvidia GPU and a 24-core CPU. We adapted the CLAM framework to allow for additional feature extractors and aggregation methods. The entire codebase was implemented in Python (v.3.10.14) and PyTorch (v2.0.1, CUDA 11.7). Metrics were computed using the \texttt{scikit-learn} library in Python \cite{buitinck_api_2013}.

\subsubsection*{\textit{Evaluation and statistical analysis}}
Classification performance is reported in terms of Matthews correlation coefficient (MCC), balanced accuracy, weighted F1 score, and AUROC. MCC, balanced accuracy, and weighted F1 score were used to account for the class imbalance in the test set. Balanced accuracy was computed as the average recall across all classes. The weighted F1 score was calculated as the average of the F1 score for each class, weighted by the number of ground truth samples in each class. For the experiments investigating the classification performance using data from all the sites, metrics are reported as mean and standard deviation over 150 replicates obtained from a nonparametric bootstrapping. The number of replicates was obtained through a power calculation using statistics from a preliminary investigation. The split between training/validation/test to obtain each of the replicas was performed on a subject level and was class-stratified, with 30\% of samples set for testing, 20\% for validation, and 50\% for training. Statistical comparison was performed through a two-sided paired permutation test with 10,000 permutations using test values from the nonparametric bootstrapping replicates, for which the significance level was set to \textit{p}=0.05. Bonferroni correction was additionally utilized in the case of multiple comparisons. For the model generalization evaluation, metrics are reported as mean and standard deviation over five replicates obtained from a nonparametric bootstrapping, with replicates obtained using the same split strategy as before. No statistical analysis was performed between the performances obtained on the in-site and out-of-site testing.

\subsubsection*{\textit{Attention mapping}}
The attention scores for each instance in a bag can be used to highlight regions in the WSIs that contributed more to the class predicted by the model. In particular, attention score heatmaps can be overlaid on the WSIs to highlight the salient histological morphologies used for classification. In this work, attention maps were computed from the models whose MCC performance was closest to the median value across the 150 replications. For a few selected WSIs, an expert pediatric neuropathologist identified regions relevant for the diagnosis, describing the characteristics of the cell population (normal, tumor, and tumor grade) and stroma. A qualitative comparison was performed between the attention maps and these regions.

\section*{Results}
\begin{figure}[ht]
    \centering
        \includegraphics[width=\textwidth]{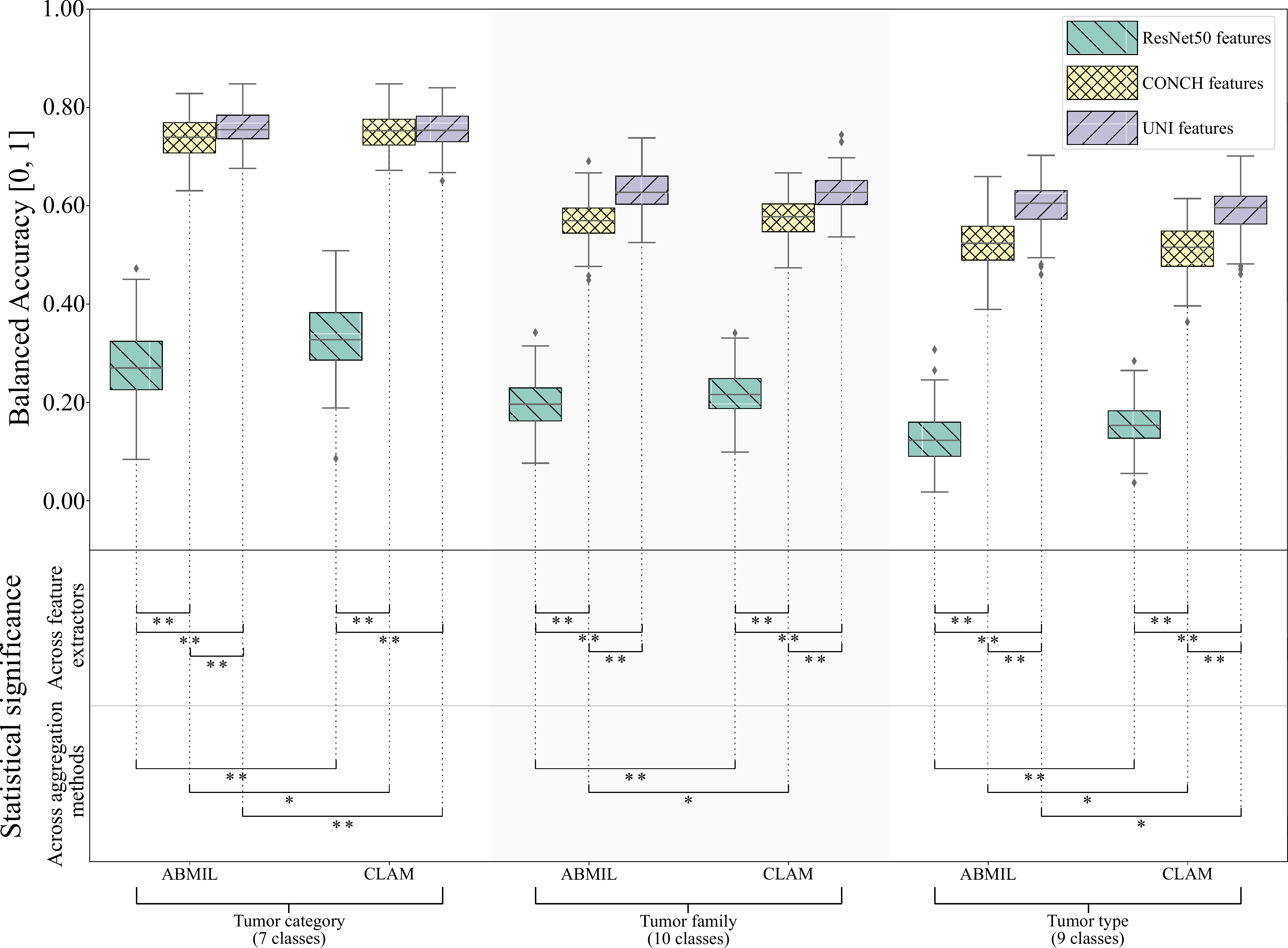}
    \caption{Matthew’s correlation coefficient on the test set for all model configurations and classification tasks, when training using data from all the available sites. Each boxplot summarizes the performance of 150 replicates of a non-parametric bootstrapping. Statistical significance is shown when comparing models within classification tasks, and separately across feature extractors and aggregation methods. (*) indicates statistical significance lower than 0.05, while (**) lower than 0.001. ABMIL, attention-based multiple-instance learning; CLAM, clustering-constrained attention multiple instance-learning; MCC, Matthew’s correlation coefficient. ResNet50, convolutional neural network pretrained on ImageNet, UNI and CONCH, vision transformer foundation models pretrained on histopathology data.}
    \label{fig:results}
\end{figure}

\begin{table}[ht]
\centering
    \caption{Classification performance when training on data from all contributing sites. Metrics are presented as mean and standard deviation with 95\% confidence intervals (CI) computed over 150 model runs.}\label{tab:all_classification_performance}
    \small
    \begin{NiceTabular}{p[c]{19mm} p[c]{19mm} p[c]{19mm} p[c]{19mm} m[c]{19mm} p[c]{19mm} p[c]{19mm}}[]
    \toprule
    \Block{1-1}{Classification\\granularity} & \Block{1-1}{Instance feature\\extractor} & \Block{1-1}{MIL aggregation\\method} & \Block{1-1}{MCC\\mean$\pm$std\\ [95\% CI]} & \Block{1-1}{Balanced\\accuracy\\ mean$\pm$std\\ [95\% CI]} & \Block{1-1}{F1 score\\ mean$\pm$std\\ [95\% CI]} & \Block{1-1}{AUROC\\ mean$\pm$std\\ [95\% CI]} \\

    \midrule
    \Block{6-1}{Tumor\\category} & \Block{2-1}{ResNet50} & \Block{1-1}{ABMIL$^\ast$} & \Block{1-1}{0.27$\pm$0.07\\ [0.26, 0.29]} & \Block{1-1}{0.34$\pm$0.06\\ [0.33, 0.35]} & \Block{1-1}{0.25$\pm$0.05\\ [0.24, 0.25]} & \Block{1-1}{0.71$\pm$0.05\\ [0.70, 0.72]} \\
    
    \cmidrule{3-7}
    
    & & \Block{1-1}{CLAM$^\diamondsuit$}& \Block{1-1}{0.33$\pm$0.07\\ [0.32, 0.34]} & \Block{1-1}{0.38$\pm$0.07\\ [0.37, 0.40]} & \Block{1-1}{0.29$\pm$0.05\\ [0.28, 0.30]} & \Block{1-1}{0.73$\pm$0.04\\ [0.72, 0.74]} \\

    \cmidrule{2-7}

    & \Block{2-1}{CONCH} & \Block{1-1}{ABMIL$^\ast$} & \Block{1-1}{0.74$\pm$0.04\\ [0.73, 0.75]} & \Block{1-1}{0.65$\pm$0.06\\ [0.64, 0.66]} & \Block{1-1}{0.63$\pm$0.06\\ [0.63, 0.64]} & \Block{1-1}{0.93$\pm$0.03\\ [0.93, 0.94]} \\

    \cmidrule{3-7}
    
    & & \Block{1-1}{CLAM$^\diamondsuit$} & \Block{1-1}{0.75$\pm$0.04\\ [0.75, 0.76]} & \Block{1-1}{0.65$\pm$0.06\\ [0.64, 0.66]} & \Block{1-1}{0.63$\pm$0.05\\ [0.63, 0.64]} & \Block{1-1}{0.93$\pm$0.02\\ [0.93, 0.94]} \\

    \cmidrule{2-7}

    & \Block{2-1}{\textbf{UNI}} & \Block{1-1}{\textbf{ABMIL$^\ast$}} & \Block{1-1}{\textbf{0.76$\pm$0.04\\ [0.75, 0.77]}} & \Block{1-1}{\textbf{0.57$\pm$0.07\\ [0.56, 0.58]}} & \Block{1-1}{\textbf{0.59$\pm$0.07\\ [0.58, 0.60]}} & \Block{1-1}{\textbf{0.94$\pm$0.02\\ [0.94, 0.94]}} \\

    \cmidrule{3-7}

    & & \Block{1-1}{CLAM$^\diamondsuit$} & \Block{1-1}{0.75$\pm$0.04\\ [0.75, 0.76]} & \Block{1-1}{0.55$\pm$0.07\\ [0.53, 0.56]} & \Block{1-1}{0.57$\pm$0.07\\ [0.56, 0.58]} & \Block{1-1}{0.94$\pm$0.02\\ [0.94, 0.94]} \\

    \midrule\midrule 
    \Block[fill=gray!10]{6-1}{Tumor\\family} & \Block[fill=gray!10]{2-1}{ResNet50} & \Block[fill=gray!10]{1-1}{ABMIL$^\ast$} & \Block[fill=gray!10]{1-1}{0.20$\pm$0.05\\ [0.19, 0.20]} & \Block[fill=gray!10]{1-1}{0.28$\pm$0.05\\ [0.28, 0.29]} & \Block[fill=gray!10]{1-1}{0.18$\pm$0.04\\ [0.17, 0.18]} & \Block[fill=gray!10]{1-1}{0.74$\pm$0.03\\ [0.73, 0.74]} \\

    \cmidrule{3-7}
    
    & & \Block[fill=gray!10]{1-1}{CLAM$^\diamondsuit$} & \Block[fill=gray!10]{1-1}{0.22$\pm$0.05\\ [0.21, 0.23]} & \Block[fill=gray!10]{1-1}{0.31$\pm$0.05\\ [0.30, 0.32]} & \Block[fill=gray!10]{1-1}{0.21$\pm$0.04\\ [0.20, 0.22]} & \Block[fill=gray!10]{1-1}{0.76$\pm$0.03\\ [0.75, 0.76]} \\

    \cmidrule{2-7}
     
    & \Block[fill=gray!10]{2-1}{CONCH} & \Block[fill=gray!10]{1-1}{ABMIL$^\ast$}  & \Block[fill=gray!10]{1-1}{0.57$\pm$0.04\\ [0.56, 0.58]} & \Block[fill=gray!10]{1-1}{0.55$\pm$0.05\\ [0.54, 0.56]} & \Block[fill=gray!10]{1-1}{0.53$\pm$0.05\\ [0.52, 0.53]} & \Block[fill=gray!10]{1-1}{0.90$\pm$0.02\\ [0.90, 0.90]} \\

    \cmidrule{3-7}
    
    & & \Block[fill=gray!10]{1-1}{CLAM$^\diamondsuit$} & \Block[fill=gray!10]{1-1}{0.58$\pm$0.04\\ [0.57, 0.58]} & \Block[fill=gray!10]{1-1}{0.55$\pm$0.05\\ [0.54, 0.56]} & \Block[fill=gray!10]{1-1}{0.53$\pm$0.05\\ [0.52, 0.54]} & \Block[fill=gray!10]{1-1}{0.90$\pm$0.02\\ [0.89, 0.90]} \\

    \cmidrule{2-7}
    
    & \Block[fill=gray!10]{2-1}{\textbf{UNI}} & \Block[fill=gray!10]{1-1}{\textbf{ABMIL$^\ast$}} & \Block[fill=gray!10]{1-1}{\textbf{0.63$\pm$0.04\\ [0.63, 0.64]}} & \Block[fill=gray!10]{1-1}{\textbf{0.55$\pm$0.05\\ [0.55, 0.56]}} & \Block[fill=gray!10]{1-1}{\textbf{0.56$\pm$0.06\\ [0.55, 0.57]}} & \Block[fill=gray!10]{1-1}{\textbf{0.91$\pm$0.02\\ [0.91, 0.91]}} \\

    \cmidrule{3-7}
    
    & & \Block[fill=gray!10]{1-1}{CLAM$^\ast$} & \Block[fill=gray!10]{1-1}{0.63$\pm$0.04\\ [0.62, 0.63]} & \Block[fill=gray!10]{1-1}{0.54$\pm$0.05\\ [0.53, 0.55]} & \Block[fill=gray!10]{1-1}{0.55$\pm$0.05\\ [0.54, 0.55]} & \Block[fill=gray!10]{1-1}{0.91$\pm$0.02\\ [0.90, 0.91]} \\

    \midrule\midrule 
    \Block{6-1}{Tumor\\type} & \Block{2-1}{ResNet50} & \Block{1-1}{ABMIL$^\ast$} & \Block{1-1}{0.13$\pm$0.05\\ [0.12, 0.14]} & \Block{1-1}{0.28$\pm$0.04\\ [0.27, 0.28]} & \Block{1-1}{0.14$\pm$0.05\\ [0.14, 0.15]} & \Block{1-1}{0.76$\pm$0.03\\ [0.76, 0.77]} \\

    \cmidrule{3-7}

    & & \Block{1-1}{CLAM$^\diamondsuit$} & \Block{1-1}{0.16$\pm$0.05\\ [0.15, 0.17]} & \Block{1-1}{0.31$\pm$0.05\\ [0.30, 0.31]} & \Block{1-1}{0.19$\pm$0.05\\ [0.18, 0.20]} & \Block{1-1}{0.78$\pm$0.03\\ [0.78, 0.79]} \\

    \cmidrule{2-7}
    
    & \Block{2-1}{CONCH} & \Block{1-1}{ABMIL$^\ast$} & \Block{1-1}{0.52$\pm$0.05\\ [0.51, 0.53]} & \Block{1-1}{0.56$\pm$0.06\\ [0.55, 0.57]} & \Block{1-1}{0.52$\pm$0.05\\ [0.51, 0.53]} & \Block{1-1}{0.92$\pm$0.02\\ [0.91, 0.92]} \\

    \cmidrule{3-7}
    
    & & \Block{1-1}{CLAM$^\diamondsuit$} & \Block{1-1}{0.51$\pm$0.05\\ [0.50, 0.52]} & \Block{1-1}{0.55$\pm$0.06\\ [0.55, 0.56]} & \Block{1-1}{0.52$\pm$0.05\\ [0.51, 0.52]} & \Block{1-1}{0.91$\pm$0.02\\ [0.91, 0.92]} \\

    \cmidrule{2-7}
    
    & \Block{2-1}{\textbf{UNI}} & \Block{1-1}{\textbf{ABMIL$^\ast$}} & \Block{1-1}{\textbf{0.60$\pm$0.05\\ [0.59, 0.61]}} & \Block{1-1}{\textbf{0.56$\pm$0.06\\ [0.55, 0.57]}} & \Block{1-1}{\textbf{0.56$\pm$0.06\\ [0.55, 0.57]}} & \Block{1-1}{\textbf{0.92$\pm$0.02\\ [0.92, 0.92]}} \\

    \cmidrule{3-7}
    
    & & \Block{1-1}{CLAM$^\diamondsuit$} & \Block{1-1}{0.59$\pm$0.05\\ [0.59, 0.60]} & \Block{1-1}{0.55$\pm$0.06\\ [0.54, 0.55]} & \Block{1-1}{0.55$\pm$0.06\\ [0.54, 0.56]} & \Block{1-1}{0.92$\pm$0.02\\ [0.92, 0.92]} \\
    \bottomrule
    
    \end{NiceTabular}
    \begin{minipage}{\textwidth}
    The best performance for each classification task in bold. Within a column, aggregation configurations for a given feature extractor without a common superscript differ significantly (two-sided permutation test (10000) iterations over 150 test values from 150 replicas of a non-parametric bootstrapping). Significance level set to 0.05. 
    Abbreviations: ABMIL, attention-based multiple-instance learning; CLAM, clustering-constrained attention multiple instance-learning; MCC, Matthew's correlation coefficient; ResNet50, convolutional neural network pretrained on ImageNet; UNI and CONCH, vision transformer foundation models pretrained on histopathology data.
    \end{minipage}

\end{table}

\subsection*{Classification performance}
As the granularity of tumor classification increased, from broader categories (seven classes) to families (10 classes) and finally types (nine classes), the overall model performance decreased, independently of the instance feature extractor or MIL aggregation method (Figure \ref{fig:results} and Table \ref{tab:all_classification_performance}). This is expected given that a finer categorization results in a smaller number of samples per class as well as less distinct classes overall. The largest decrease in performance is seen when transitioning from tumor category to family classification, with an average performance drop of 0.13 MCC across all model settings. When going from tumor families to types, the performance drop was less pronounced, with an average decrease of 0.05 MCC across all model settings. Overall, the models using UNI and ResNet50 extracted features experienced the smallest drop in performance compared to the models using CONCH features. 

When comparing feature extractors, models using UNI and CONCH extracted features obtained a significantly higher performance compared to ResNet50 features, across all classification tasks and independently of the aggregation method (Table \ref{tab:all_classification_performance}). Additionally, models using UNI features obtained a marginally higher classification performance for tumor category and type classification, with a larger difference on the tumor family classification. When considering the MIL aggregation method, the performance difference was marginal when comparing ABMIL and CLAM on the same features. Models trained with ResNet50 features benefited from using CLAM aggregation across all classification tasks, improving performance by 3.3\% on average. When using UNI features, ABMIL aggregation was better than CLAM for tumor category and type classification, while models using CONCH features showed differences between the two aggregation methods across all classification tasks. 

On the tumor category classification task (seven classes), the highest performance was achieved by models using UNI-extracted features and the ABMIL aggregation method, with an MCC of 0.76$\pm$0.04 (95\% CI [0.75, 0.77]). This performance was significantly higher than that of the best model using ResNet50 features (\textit{p}$<$0.001) (MCC of 0.33$\pm$0.07, 95\% CI [0.32, 0.34]), while there was no significant difference in performance compared to models using CONCH-extracted features (\textit{p}$>$0.01), (0.75$\pm$0.04, 95\% CI [0.75, 0.76]). A similar trend is seen for the tumor family classification task (10 classes), where CONCH and UNI feature extractors pre-trained on histopathology images perform better than ResNet50 pretrained on ImageNet. In particular, models using UNI features and ABMIL aggregation achieved an MCC of 0.62$\pm$0.04 (95\% CI [0.63, 0.64]), which was significantly higher (\textit{p}$<$0.001) compared to using CONCH-features (0.58$\pm$0.04, 95\% CI [0.57, 0.58]). Additionally, models using UNI or CONCH features both achieved a significantly higher performance compared to that of the best model using ResNet50 features (\textit{p}<0.0001) (MCC: 0.22$\pm$0.04, 95\% CI [0.21, 0.22]). Finally, on the tumor type classification (13 classes), the best-performing models were those using UNI features, achieving an MCC of 0.60$\pm$0.05 (95\% CI [0.59, 0.61]). This was significantly higher compared to models using either CONCH (0.52$\pm$0.05, 95\% CI [0.51, 0.53]) or ResNet50 (0.16$\pm$0.05, 95\% CI [0.15, 0.17]) extracted features (\textit{p}<0.001). 

Looking at the per-class F1 scores obtained by the best model for each classification task (Table \ref{tab:f1_score_all_classification} and Figures \ref{fig:cm_tumor_category}, \ref{fig:cm_tumor_family}, and \ref{fig:cm_tumor_type}), in the case of tumor category classification, germ cell tumors (2.2\% of training dataset) and mesenchymal, non-meningothelial tumors (2.0\% of training dataset) obtained the lowest F1 scores, 0.10$\pm$0.18 (95\% CI [0.07, 0.13]) and 0.11$\pm$0.017 (95\% CI [0.08, 0.14]), respectively, being mostly misclassified with the most representative tumor entities. The best-classified class was gliomas/glioneuronal/neuronal tumors, with an F1 score of 0.93$\pm$0.01 (95\% CI [0.93, 0.93]), which represented 62\% of all tumor category samples.\\
For tumor family classification, adult-type diffuse gliomas (4.5\% of training dataset), pediatric-type diffuse high-grade glioma (2,5\% of training dataset), and embryonal tumors NOS (2,3\% of training dataset) had the lowest F1 scores, with 0.14$\pm$0.14 (95\% CI [0.12, 0.16]), 0.15$\pm$0.21 (95\% CI [0.11, 0.18]), and 0.20$\pm$0.24 (95\% CI [0.16, 0.24]), respectively. The confusion matrix revealed that pediatric-type diffuse high-grade gliomas and adult-type diffuse gliomas are misclassified with the most represented tumor families in their tumor category, circumscribed astrocytic gliomas (39.5\% of training dataset) and glioneuronal/neuronal tumors (12\% of training dataset). Embryonal tumors NOS were instead mostly misclassified with medulloblastoma (19\% of training dataset), the most represented tumor family of its tumor category. Adamantinomatous craniopharyngioma achieved the highest F1 score of 0.96$\pm$0.08 (95\% CI [0.95, 0.98]), despite representing only 3.6\% of the training dataset. Additionally, when training classifiers specific for the tumor families of gliomas/glioneuronal/neuronal tumors and embryonal tumors, shows marginal improvements in classification scores (Tables \ref{tab:f1_score_ET_category_only} and \ref{tab:f1_score_GNN_category_only}, and Figures \ref{fig:cm_GGN_category_only} and \ref{fig:cm_ET_category_only}.\\
For tumor type classification, medulloblastoma WNT activated (3.0\% of training dataset), ependymoma grade 1-2 (4.3\% of training dataset), glioblastoma (4.9\% of training dataset) and ganglioglioma (9.8\% of training dataset) had F1 scores lower than 0.5, with the worst classified class being glioblastoma, which had a score of 0.27$\pm$0.19 (95\% CI [0.24, 0.30]). Looking at the confusion matrices, ependymomas grade 1-2 were mostly misclassified with grade 3 ependymoma, and medulloblastoma WNT activated with medulloblastoma non-WNT/non-SHH. Glioblastomas and gangliogliomas were primarily misclassified as the most represented tumor type, pilocytic astrocytoma. The best-classified classes were medulloblastoma non-WNT/non-SHH (11.3\% of training dataset), with an F1 score of 0.83$\pm$0.06 (95\% CI [0.81, 0.84]), and pilocytic astrocytoma (50.6\% of training dataset), with an F1 score of 0.84$\pm$0.03 (95\% CI [0.83, 0.84]).

Finally, when looking at the models' performance across classification tasks when training on a balanced dataset, in the case of class-balance (10 cases randomly selected for each tumor entity), results show that overall classification performance reduces on average by 0.13 MCC compared to when using all the data available. F1 scores for the most represented classes show a significant decrease, while scores for the poorly represented classes increase. A similar trend is also seen when using balanced tissue areas, with an overall decrease in performance of 0.22 MCC compared to when using all the data available. F1 scores for the most represented classes show a significant decrease, while scores for the poorly represented classes marginally increased. Detailed results are presented in Tables \ref{tab:classification_results_balanced_classes}, \ref{tab:f1_score_balanced_classes}, \ref{tab:classification_results_balanced_tissue_area} and \ref{tab:f1_score_balanced_tissue_area}. 

\subsection*{Model generalization}
When looking at the in-site test sets, the trend in classification performance across the classification tasks was comparable to that obtained when training and evaluating the entire dataset (see Tables \ref{tab:classification_performance_generalization_in_site_data_min_10}, \ref{tab:classification_performance_generalization_out_of_site_data_min_10}, \ref{tab:f1_score_generalization_in_site_data_min_10} and \ref{tab:f1_score_generalization_out_of_site_data_min_10} for the performance details). The difference between the models' performance on the in-site and the out-of-site test sets was not consistent among the classification tasks, as seen in Table \ref{tab:generalization_summary_min_10}. All models achieved higher classification performance on the hold-out sites compared to the training sites when looking at the tumor category classification, with models using Resnet50 features obtaining up to 0.09 higher MCC. For the tumor family and tumor type classification, the majority of models showed a drop in performance when tested on the holdout sites compared to the training sites, with a larger decrease observed for the tumor type classification. However, models using UNI features achieved marginally better performance on the tumor type classification on the hold-out sites compared to the training sites. No consistent trend in performance drop was observed between models using ResNet50, CONCH or UNI features, or ABMIL or CLAM aggregation. The generalization results, where the minimum number of classes was reduced to eight to include a larger number of tumor entities, are presented in Tables \ref{tab:classification_performance_generalization_in_site_data_min_8}, \ref{tab:classification_performance_generalization_out_of_site_data_min_8},
\ref{tab:generalization_summary_min_8}, \ref{tab:f1_score_generalization_in_site_data_min_8} and \ref{tab:f1_score_generalization_out_of_site_data_min_8}. These results show the performance on in-site testing closely mirroring the performance observed when training and evaluating the entire dataset. Additionally, the change in performance on out-of-site testing exhibits a similar trend to that seen when including only tumor entities with at least 10 cases.

\begin{table}[ht]
 \centering
 \caption{Classification performance when training on two sites and testing on the remaining four sites.}\label{tab:generalization_summary_min_10}

    \small
    \begin{NiceTabular}{m[c]{16mm} m[c]{16mm} m[c]{16mm} m[c]{16mm} m[c]{16mm} m[c]{16mm} m[c]{16mm} m[c]{16mm}}[]

    \toprule

    \Block{2-1}{Classification\\granularity} & \Block{2-1}{Instance\\feature\\extractor} & \Block{2-1}{Aggregation\\method} & \Block{1-2}{In-site testing} & & \Block{1-2}{Out-of-site testing} & & \Block{2-1}{Drop in\\performance\\(MCC difference)} \\

    \cmidrule{4-7}
    
    & & & \Block{1-1}{\textbf{MCC}\\mean$\pm$std\\ [95\% CI]} & \Block{1-1}{\textbf{Balanced}\\\textbf{accuracy}\\mean$\pm$std\\ [95\% CI]} & \Block{1-1}{\textbf{MCC}\\mean$\pm$std\\ [95\% CI]} & \Block{1-1}{\textbf{Balanced}\\\textbf{accuracy}\\mean$\pm$std\\ [95\% CI]} & \\

    \midrule

    \Block{6-1}{Tumor\\Category} & \Block{2-1}{ResNet50} & \Block{1-1}{ABMIL} & \Block{1-1}{0.42$\pm$0.09\\ [0.30, 0.55]} & \Block{1-1}{0.67$\pm$0.07\\ [0.58, 0.77]} & \Block{1-1}{0.51$\pm$0.08\\ [0.40, 0.61]} & \Block{1-1}{0.74$\pm$0.04\\ [0.68, 0.80]} & \Block{1-1}{0.09$\uparrow$} \\

    \cmidrule{3-8}

    & & \Block{1-1}{CLAM} & \Block{1-1}{0.45$\pm$0.06\\ [0.36, 0.53]} & \Block{1-1}{0.69$\pm$0.07\\ [0.59, 0.80]} & \Block{1-1}{0.54$\pm$0.07\\ [0.44, 0.64]} & \Block{1-1}{0.76$\pm$0.07\\ [0.66, 0.86]} & \Block{1-1}{0.09$\uparrow$} \\

    \cmidrule{2-8}
    
    & \Block{2-1}{CONCH} & \Block{1-1}{ABMIL} & \Block{1-1}{0.78$\pm$0.04\\ [0.73, 0.83]} & \Block{1-1}{0.88$\pm$0.03\\ [0.84, 0.93]} & \Block{1-1}{0.79$\pm$0.05\\ [0.73, 0.86]} & \Block{1-1}{0.89$\pm$0.04\\ [0.83, 0.94]} & \Block{1-1}{0.01$\uparrow$} \\

    \cmidrule{3-8}
    
    & & \Block{1-1}{CLAM} & \Block{1-1}{0.80$\pm$0.07\\ [0.70, 0.90]} & \Block{1-1}{0.89$\pm$0.05\\ [0.82, 0.95]} & \Block{1-1}{0.82$\pm$0.02\\ [0.79, 0.86]} & \Block{1-1}{0.92$\pm$0.02\\ [0.90, 0.94]} & \Block{1-1}{0.02$\uparrow$} \\

    \cmidrule{2-8}
    
    & \Block{2-1}{UNI} & \Block{1-1}{ABMIL} & \Block{1-1}{0.81$\pm$0.04\\ [0.75, 0.87]} & \Block{1-1}{0.87$\pm$0.02\\ [0.84, 0.90]} & \Block{1-1}{0.85$\pm$0.02\\ [0.82, 0.87]} & \Block{1-1}{0.92$\pm$0.01\\ [0.91, 0.93]} & \Block{1-1}{0.04$\uparrow$} \\
    \cmidrule{3-8}
    
    & & \Block{1-1}{CLAM} & \Block{1-1}{0.81$\pm$0.04\\ [0.76, 0.86]} & \Block{1-1}{0.87$\pm$0.02\\ [0.84, 0.91]} & \Block{1-1}{0.84$\pm$0.04\\ [0.79, 0.90]} & \Block{1-1}{0.91$\pm$0.03\\ [0.86, 0.95]} & \Block{1-1}{0.03$\uparrow$} \\

    \midrule\midrule

 \Block[fill=gray!10]{6-1}{Tumor\\family} & \Block[fill=gray!10]{2-1}{ResNet50} & \Block[fill=gray!10]{1-1}{ABMIL} & \Block[fill=gray!10]{1-1}{0.28$\pm$0.04\\ [0.23, 0.34]} & \Block[fill=gray!10]{1-1}{0.40$\pm$0.03\\ [0.36, 0.44]} & \Block[fill=gray!10]{1-1}{0.26$\pm$0.02\\ [0.24, 0.29]} & \Block[fill=gray!10]{1-1}{0.40$\pm$0.02\\ [0.38, 0.43]} & \Block[fill=gray!10]{1-1}{-0.02$\downarrow$} \\

    \cmidrule{3-8}

    & & \Block[fill=gray!10]{1-1}{CLAM} & \Block[fill=gray!10]{1-1}{0.31$\pm$0.10\\ [0.17, 0.44]} & \Block[fill=gray!10]{1-1}{0.43$\pm$0.09\\ [0.31, 0.54]} & \Block[fill=gray!10]{1-1}{0.29$\pm$0.06\\ [0.21, 0.37]} & \Block[fill=gray!10]{1-1}{0.41$\pm$0.06\\ [0.33, 0.49]} & \Block[fill=gray!10]{1-1}{-0.02$\downarrow$} \\

    \cmidrule{2-8}
    
    & \Block[fill=gray!10]{2-1}{CONCH} & \Block[fill=gray!10]{1-1}{ABMIL} & \Block[fill=gray!10]{1-1}{0.60$\pm$0.04\\ [0.54, 0.66]} & \Block[fill=gray!10]{1-1}{0.67$\pm$0.05\\ [0.61, 0.74]} & \Block[fill=gray!10]{1-1}{0.58$\pm$0.03\\ [0.54, 0.62]} & \Block[fill=gray!10]{1-1}{0.59$\pm$0.02\\ [0.57, 0.62]} & \Block[fill=gray!10]{1-1}{-0.02$\downarrow$} \\

    \cmidrule{3-8}
    
    & & \Block[fill=gray!10]{1-1}{CLAM} & \Block[fill=gray!10]{1-1}{0.65$\pm$0.02\\ [0.62, 0.68]} & \Block[fill=gray!10]{1-1}{0.72$\pm$0.02\\ [0.68, 0.75]} & \Block[fill=gray!10]{1-1}{0.59$\pm$0.02\\ [0.56, 0.62]} & \Block[fill=gray!10]{1-1}{0.64$\pm$0.01\\ [0.63, 0.65]} & \Block[fill=gray!10]{1-1}{-0.06$\downarrow$} \\

    \cmidrule{2-8}
    
    & \Block[fill=gray!10]{2-1}{UNI} & \Block[fill=gray!10]{1-1}{ABMIL} & \Block[fill=gray!10]{1-1}{0.67$\pm$0.05\\ [0.60, 0.74]} & \Block[fill=gray!10]{1-1}{0.72$\pm$0.05\\ [0.65, 0.79]} & \Block[fill=gray!10]{1-1}{0.66$\pm$0.02\\ [0.64, 0.67]} & \Block[fill=gray!10]{1-1}{0.65$\pm$0.01\\ [0.64, 0.67]} & \Block[fill=gray!10]{1-1}{-0.01$\downarrow$} \\

    \cmidrule{3-8}
    
    & & \Block[fill=gray!10]{1-1}{CLAM} & \Block[fill=gray!10]{1-1}{0.68$\pm$0.07\\ [0.58, 0.78]} & \Block[fill=gray!10]{1-1}{0.73$\pm$0.06\\ [0.64, 0.81]} & \Block[fill=gray!10]{1-1}{0.62$\pm$0.03\\ [0.58, 0.67]} & \Block[fill=gray!10]{1-1}{0.61$\pm$0.04\\ [0.55, 0.67]} & \Block[fill=gray!10]{1-1}{-0.06$\downarrow$} \\

    \midrule\midrule

    \Block{6-1}{Tumor\\type} & \Block{2-1}{ResNet50} & \Block{1-1}{ABMIL} & \Block{1-1}{0.40$\pm$0.13\\ [0.23, 0.58]} & \Block{1-1}{0.56$\pm$0.04\\ [0.51, 0.61]} & \Block{1-1}{0.26$\pm$0.14\\ [0.07, 0.45]} & \Block{1-1}{0.45$\pm$0.06\\ [0.36, 0.53]} & \Block{1-1}{-0.14$\downarrow$} \\

    \cmidrule{3-8}

    & & \Block{1-1}{CLAM} & \Block{1-1}{0.39$\pm$0.09\\ [0.27, 0.51]} & \Block{1-1}{0.58$\pm$0.07\\ [0.48, 0.67]} & \Block{1-1}{0.29$\pm$0.05\\ [0.22, 0.37]} & \Block{1-1}{0.51$\pm$0.04\\ [0.46, 0.57]} & \Block{1-1}{-0.10$\downarrow$} \\

    \cmidrule{2-8}
    
    & \Block{2-1}{CONCH} & \Block{1-1}{ABMIL} & \Block{1-1}{0.67$\pm$0.04\\ [0.62, 0.72]} & \Block{1-1}{0.78$\pm$0.03\\ [0.75, 0.82]} & \Block{1-1}{0.56$\pm$0.07\\ [0.46, 0.66]} & \Block{1-1}{0.63$\pm$0.02\\ [0.60, 0.65]} & \Block{1-1}{-0.11$\downarrow$} \\

    \cmidrule{3-8}
    
    & & \Block{1-1}{CLAM} & \Block{1-1}{0.69$\pm$0.06\\ [0.61, 0.77]} & \Block{1-1}{0.79$\pm$0.05\\ [0.72, 0.86]} & \Block{1-1}{0.58$\pm$0.04\\ [0.52, 0.63]} & \Block{1-1}{0.63$\pm$0.02\\ [0.60, 0.66]} & \Block{1-1}{-0.11$\downarrow$} \\

    \cmidrule{2-8}
    
    & \Block{2-1}{UNI} & \Block{1-1}{ABMIL} & \Block{1-1}{0.61$\pm$0.12\\ [0.45, 0.78]} & \Block{1-1}{0.73$\pm$0.10\\ [0.59, 0.86]} & \Block{1-1}{0.64$\pm$0.04\\ [0.58, 0.69]} & \Block{1-1}{0.63$\pm$0.03\\ [0.58, 0.68]} & \Block{1-1}{+0.03$\uparrow$} \\

    \cmidrule{3-8}
    
    & & \Block{1-1}{CLAM} & \Block{1-1}{0.60$\pm$0.10\\ [0.46, 0.74]} & \Block{1-1}{0.69$\pm$0.09\\ [0.56, 0.81]} & \Block{1-1}{0.62$\pm$0.03\\ [0.58, 0.67]} & \Block{1-1}{0.61$\pm$0.04\\ [0.55, 0.67]} & \Block{1-1}{+0.02$\uparrow$} \\

    \bottomrule
    \end{NiceTabular}
    \begin{minipage}{\textwidth}
    \textit{Note}: Metrics are presented as mean and standard deviation with 95\% confidence intervals (CI) computed over 5 model runs. The difference between testing on the in-site and out-of-site data is also presented, with arrows showing if performance increased ($\uparrow$) or decreased ($\downarrow$). Abbreviations: ABMIL, attention-based multiple-instance learning; CLAM, clustering-constrained attention multiple instance-learning; MCC, Matthew's correlation coefficient; ResNet50, convolutional neural network pretrained on ImageNet; UNI and CONCH, vision transformer foundation models pretrained on histopathology data.
    \end{minipage}
\end{table}

\subsection*{Attention mapping}
\begin{figure}[ht]
    \centering
        \includegraphics[width=\textwidth]{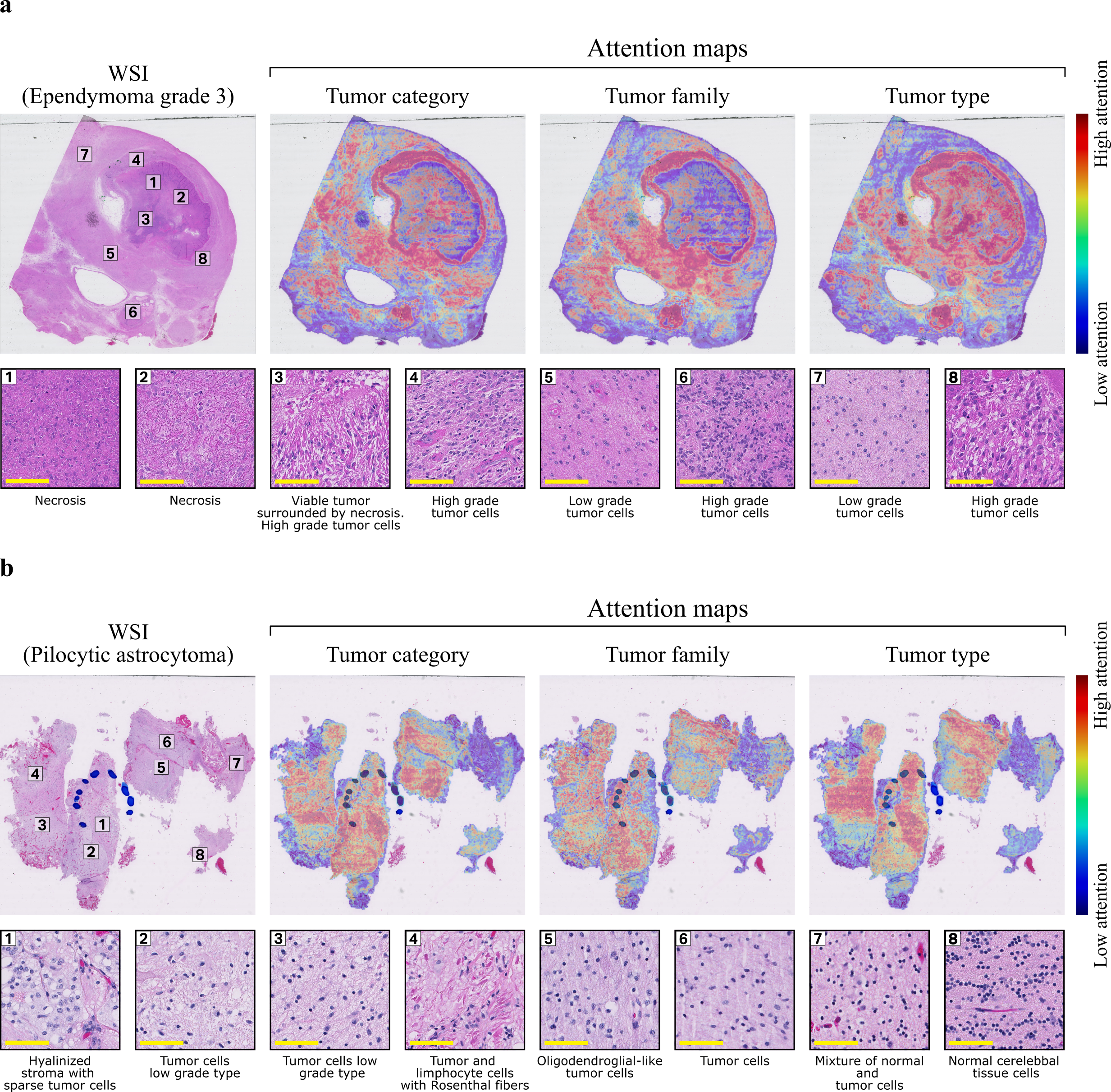}
    \caption{Comparison of attention maps with histopathology-defined regions. Cellular and extracellular details for eight regions across a whole slide image (WSI) are provided for two examples (a and b) by a specialist pediatric neuropathologist. Attention maps for the best-performing model configuration (UNI features with ABMIL aggregation) are shown for models trained on the three classification tasks. Warm colors identify regions with high attention scores. Scale bar: 50 micrometers for the zoomed-in regions. UNI, vision transformer foundation models pretrained on histopathology data; ABMIL, attention-based multiple-instance learning. The pen mark in (b) is from the original WSI and is ignored by the model in decision-making.}
    \label{fig:attention_mapping}
\end{figure}

A comparison between attention maps obtained from the best model configuration (UNI-features with ABMIL aggregation) across classification tasks with the regions identified and described by the neuropathologist is presented in Figure \ref{fig:attention_mapping}. Overall, the models' attention identified regions deemed as relevant for tumor diagnosis, encompassing high and low tumor grade areas, as well as necrotic regions and those where abnormal stroma could be found due to tumor presence. Additional examples of attention maps for all models and classification tasks are presented in Figures \ref{fig:sup_attention_ependymoma_g3}, \ref{fig:sup_attention_pilocytic_astrocytoma_1}, \ref{fig:sup_attention_pilocytic_astrocytoma_2}. 
The example diagnosed with ependymoma grade 3 (Figure \ref{fig:attention_mapping}.a) features a necrotic area encircled by a thin border of high-grade tumor cells, which are further surrounded by a broad region of low-grade tumor. Additionally, several other regions scattered across the WSI were described by the pathologist as high-grade tumor hot spots. Tumor grade heterogeneity is a common characteristic of high-grade ependymoma. Attention maps highlighted these regions differently depending on the classification task for which the model was trained. The necrotic region was partially attended by the model trained for tumor category and family classification, but it was attended by the model trained for tumor type classification. The low-grade tumor regions, while highlighted in part by all the models, were primarily attended to by the model trained for tumor category classification. The tumor family and type classification models attended selectively to low-grade tumor areas. Lastly, the high-grade tumor regions, including hot spots and viable tumor within the necrosis, were primarily highlighted by the tumor family and type classification models.  
In the case of the pilocytic astrocytoma case shown in Figure \ref{fig:attention_mapping}b, larger areas of low-grade tumor were identified in the WSI by the pathologist, along with regions of mixed normal-tumor tissue and areas showing stroma changes with hyalinization and Rosenthal fibers. However, the model trained for tumor type classification showed weaker attention compared to the model trained for tumor category and family classification on the regions with low-grade cells. All models attended to the areas with visible Rosenthal fibers, with higher attention scores from the tumor type classification. The mixed normal-tumor tissue was highlighted by the attention of models trained for tumor category classification, but was largely ignored by those trained for tumor family and type classification.

\section*{Discussion}
In this study, we investigated the classification of pediatric brain tumors on a national multi-center digital histopathology cohort with a wide range of diagnoses using an attention-based MIL approach on WSI features extracted by histology pre-trained foundation models to obtain a per-case classification. Our findings demonstrate the potential of leveraging DL for the classification of pediatric brain tumors at different levels of granularity.

\subsection*{\textit{Feature extractors and aggregation methods}}
Our results show that, regardless of the MIL aggregation method, there is a significant improvement in classification performance when training models on instance-level features obtained from in-domain pretrained feature extractors (CONCH and UNI) compared to when using features from an out-of-domain pre-trained feature extractor (ResNet50). The impact of feature extractors in CPath has been investigated in several studies \cite{wang_transformer-based_2022, chen_towards_2024, li_improving_2022}, demonstrating that self-supervised pretraining on in-domain histopathology data consistently results in performance improvements across a variety of tasks compared to ImageNet pretraining. The possibility of pretraining histology-specific and task-agnostic feature extractors has been facilitated by the curation of large histopathology datasets as well as methodological advancements in self-supervised pretraining. When considering models using CONCH and UNI features, a statistically significant difference in performance is observed between the two feature extractors when considering tumor family and type classification. Additionally, there is a difference in the performance drop when increasing tumor classification granularity. The difference in the sourcing and quality of the pretraining datasets can be used as an explanation for the higher classification performance and lower performance drop observed in models trained using UNI features. UNI was pretrained on a larger corpus of patches (100M vs 1.17M) obtained from WSIs of mostly cancer tissues, whereas the CONCH pretraining dataset was obtained by extracting histology images from publicly available research articles. Another possible reason can be found in the dimension of the feature space used for instance representations, with UNI outputting 1024 feature vectors while CONCH outputs 512. 

The difference in classification performance observed between aggregation methods partially agrees with the literature. In particular, the performance improvement of CLAM over ABMIL aggregation for models using ResNet50 aligns with the reported results \cite{wang_transformer-based_2022, zhang_gigapixel_2022, chen_benchmarking_2024, liang_interpretable_2023}. For example, Zhang et al. reported an average improvement of 1.3\% in classification performance across three classification tasks when using CLAM over ABMIL \cite{ zhang_gigapixel_2022}. Moreover, results presented here show no improvements or marginally worse performance when using UNI or CONCH features with CLAM aggregation. This was unexpected considering that CLAM is an extension of ABMIL. However, as shown by Chen et al. \cite{chen_benchmarking_2024} in the comparison of several foundation models and instance aggregation methods on nine CPath tasks, advanced aggregation approaches did not always outperform ABMIL and a case-by-case evaluation should be performed. 

Overall, several instance-level feature extractors and aggregation methods are available in the literature \cite{chen_scaling_2022, chen_benchmarking_2024, vorontsov_foundation_2024}. UNI and CONCH are among the recent and best-performing foundation models, designed for digital histopathology, but other feature extractors such as Virchow \cite{vorontsov_foundation_2024} could be evaluated given the comparable performance on adult brain tumors. Feature encoders such as HIPT \cite{chen_scaling_2022} which uses the hierarchical structure of WSIs to learn a slide-level representation could also be considered.

Classification performance reported in this study, even though not directly comparable given the differences in sample size and classes, shows similar trends to the available literature on brain tumor classification using WSI data when considering coarse and fine-grained classification. Using data from EBRAINS \cite{roetzer-pejrimovsky_digital_2022}, which is one of the few datasets available with a wide range of diagnoses from adult and pediatric brain tumors, Liu et al. evaluated tumor type classification when using UNI features and ABMIL aggregation \cite{chen_towards_2024}. The authors reported a balanced accuracy of 0.88 for the coarse-grained classification (12 classes) and 0.67 for the fine-grained classification (30 classes). The performance drop going from coarse to fine-grained classification is comparable to ours when going from tumor category to family classification; however, only a few tumor entities have overlap with the grouping in this paper.

\subsubsection*{\textit{Miss-classifications}}
When looking at the class-wise performance across classification tasks, germ cell tumors and mesenchymal, non-meningothelial tumors are misclassified as either CNS embryonal tumors or as gliomas/glioneuronal/neuronal tumors. Disregarding the class imbalance, misclassification as CNS embryonal tumors or as gliomas/glioneuronal/neuronal tumors can be related to the shared embryonic lineage between these tumor categories. Additionally, the variability in germinal layers of germ cell tumors, which drives cell differentiation, contributes to the challenges in histological classification.

In the case of tumor family classification, embryonal tumors NOS are 42\% of the time misclassified as medulloblastoma and 16\% of the time as other CNS embryonal tumors, highlighting the significant challenges in accurately diagnosing these tumors due to their overlapping histological features and the necessity for precise molecular testing to ensure correct classification \cite{ronsley2024pediatric}. Tumor family classes of pediatric-type diffuse high-grade gliomas and glioneuronal/neuronal tumors were primarily misclassified as circumscribed astrocytic gliomas, which belong to the same tumor category of gliomas/glioneuronal and neuronal tumors. For pediatric-type diffuse high-grade gliomas, the differential diagnosis includes glioneuronal tumors (e.g., gangliogliomas) and circumscribed astrocytic gliomas (e.g., pilocytic astrocytomas), and differentiation is challenging, requiring both histomorphological and molecular analysis \cite{gianno_paediatric-type_2022}. This again highlights that the features learned from the image data alone are insufficient to differentiate between tumors sharing similar histomorphological traits but different molecular profiles. Glioneuronal/neuronal tumors, which show neuronal-like cells concomitant with glia components, share features with pediatric-type diffuse low-grade gliomas, and both histomorphology and molecular information are needed for differentiation \cite{han_glial_2024}. Most of the glioneuronal/neuronal tumors available in the dataset belong to the ganglioglioma tumor type, which is less circumscribed. Thus, the misclassification with circumscribed astrocytic gliomas is less expected. A possible reason for this could be that diffuse low-grade gliomas were not among the classes included in the tumor family classification, and model classification was biased toward the class most represented (circumscribed astrocytic gliomas - 39\% of the dataset) which also, to a lower extent compared to pediatric type diffuse high-grade gliomas, shares features with glioneuronal/neuronal tumors (DNET, which are ~30\% of the tumor types in this family, are well-circumscribed).
Adult-type diffuse gliomas show a misclassification pattern similar to that of pediatric-type diffuse high-grade gliomas, with a large proportion being classified as circumscribed astrocytic gliomas and glioneuronal/neuronal tumors. As for the case of the pediatric-type gliomas, the misclassification of adult-type diffuse gliomas with circumscribed gliomas is unexpected considering the histomorphological differences.  A bias toward the most representative class could explain this, especially considering the balanced dataset analysis results, where the misclassification rate toward circumscribed gliomas decreases, shifting towards pediatric-type diffuse high-grade gliomas. Additionally, when looking at tumor type classification, glioblastoma misclassification as pilocytic astrocytoma drops from 27\% in the unbalanced dataset to 5\% in the balanced dataset.
At the tumor type level, the misclassification of medulloblastoma WNT-activated with medulloblastoma non-WNT non-SHH indicates once again that the features by the models from the image data are not sufficient to capture differences in molecular characteristics.

\subsubsection*{\textit{Model generalization}}
The generalization of DL-based CPath methods to out-of-distribution data has been investigated in a range of applications \cite{jahanifar_domain_2023, jarkman_generalization_2022, otalora_staining_2019}. Overall, these methods perform worse on out-of-distribution data compared to in-domain data due to domain shift, with research efforts focused on reducing the performance gap through domain adaptation and domain generalization methods  \cite{jarkman_generalization_2022}. Among the domain generalization methods, self-supervised pre-training of feature encoders on a large and diverse dataset has been shown to improve the generalization of learned representations \cite{chen_towards_2024}. The results on the out-of-site data in this study show that models trained using instance-level features from UNI and CONCH outperform those using ImageNet across all classification tasks. However, these results indicate only that the instance-level representations obtained from the histology pre-trained feature encoders are less affected by domain shift than those obtained when using a general ImageNet pre-trained model. Regardless of the feature extractor, no consistent benefit was observed in using CLAM over ABMIL when comparing classification performance between in-site and out-of-site testing.

\subsubsection*{\textit{Attention mapping}}
The qualitative analysis of model attention maps demonstrated that there is an overlap between the regions the models learn to be relevant for classification and those deemed important for diagnosis by the pathologist. Models attended to the tumor regions as well as other relevant features such as necrosis and abnormal stroma. Additionally, depending on the classification task granularity, models selectively focused on different diagnostic regions. This was particularly visible in the case of the ependymoma grade 3 sample, where the high-grade tumor regions were the most relevant for tumor type classification, whereas both high and low-grade tumor areas were utilized for tumor category classification. The importance of the necrotic area also changes from being partially ignored for tumor category and family classification, to being relevant for tumor type classification. Thus, while the features of the relevant diagnostic regions are the same across classification tasks (when using the same feature extractor), models learn to selectively focus on those that are relevant based on the classification granularity.

\subsubsection*{\textit{Limitations}}
Despite the promising results, this study has some limitations. First, to allow for a larger training set and include more tumor classes, which would provide a more comprehensive evaluation of models' generalization, cases with unclear diagnoses in the current dataset should be re-evaluated incorporating molecular information and datasets from multinational centers should be included. Another aspect of the dataset that influences the evaluation of model generalization presented here is that glass slides from one of the training sites and two of the out-of-site testing sites were scanned at the same location with the same scanner. Digitizing slides using the same scanner reduces domain shift and can lead to models performing well on the in-site data, but limits the generalization on out-of-site data. Consequently, the results presented reflect the domain shift caused by the slide preparation protocol and not the differences between scanners. Moreover, the results also reflect only variability in slide preparation in Sweden, not across different countries. Testing on additional datasets should be considered. Additionally, the generalization analysis presented does not show each out-of-site individually, and class distribution shift was not accounted for. \\
Interpretability of the attention maps is limited by the lack of detailed (pixel or region-level) annotations in the dataset. While single-case evaluations are beneficial, the alignment between the histomorphological features pathologists use for diagnosis should be comprehensively and quantitatively compared to regions utilized by the model for classification. 
The clinical utility of this study is hampered by the lack of accurate differentiation between high-grade gliomas and circumscribed astrocytic gliomas, which have significant prognostic implications. Despite the routine use of histological features such as cellularity, mitotic activity, necrosis, and pleomorphism for differentiating high-grade and low-grade gliomas, the visual features extracted by these histology-pretrained models are not specific to glioma brain tumors, missing important clues that are routinely used for grade classification. Fine-tuning these models to be specific to pediatric tumors would be preferable; however, the unavailable extensive data required for such pre-training/fine-tuning, this remains a challenge and avenue for future work.
Finally, this study focused only on the classification of tumor category, families, and types, while analysis such as survival rate prediction and molecular biomarker identification (e.g., BRAF mutation) would additionally provide a more comprehensive description of the potential of the unique dataset used in the study. Furthermore, only H\&E WSIs were used for model training, while multimodal approaches such as fusion of age \cite{tampu_2024_pediatric} or other healthcare record data, immunohistochemistry WSIs \cite{spyretos_early_2024}, and molecular data \cite{steyaert_multimodal_2023, diaz_de_stahl_swedish_2023} could potentially improve the performance of the models and open possibilities for predicting the diagnosis independent of the molecular testing. 

\section*{Translational utility}
In light of the results of this work, the translational utility of a DL model for pediatric brain tumor classification using digital histopathology images is both promising and limited. The model's reliance solely on histopathology data restricts its clinical applicability, given that brain tumor diagnosis typically integrates imaging, histopathology, and molecular analysis. Nonetheless, this study represents a significant first step in identifying potentials and limitations, setting a baseline for future research.

\section*{Conclusion}
In this work, we implemented attention-based multiple-instance learning on WSI features extracted by histology foundation models to diagnose pediatric brain tumors on a multicenter Swedish cohort. Classification and model generalization results demonstrate the flexibility of histology-specific encoders over generic models to transfer to new tasks and datasets. In all the experiments, UNI-derived features, aggregated through ABMIL, resulted in the highest classification performance. On model generalization, performance was maintained for tumor category classification but decreased for fine-grained classification, with a similar drop across models. Overall, these findings expand the advancement of computational pathology in pediatric brain cancer diagnosis. 

\section*{Author contribution}
I.E.T. curated the dataset, performed the formal analysis, interpreted the results, drafted and revised the manuscript. P.N. contributed to the dataset collection and curation, and results interpretations. C.S. contributed to the definition of the analysis methods. I.B. contributed to the dataset curation, clinical revision and funding. A.S. contributed to the dataset curation and histopathological interpretation of selected examples. G.P. and T.D.S. contributed to dataset collection and curation, and clinical revision. J.S. contributed to the dataset collection. P.L. contributed to the dataset collection, curation and funding. N.H.H. provided funding, contributed to the dataset collection and curation, result interpretation, and manuscript editing. All authors reviewed and approved the final manuscript.

\section*{Acknowledgments}
The authors acknowledge The Swedish Childhood Tumor Biobank, supported by the Swedish Childhood Fund for access to data and samples. The Department of Pathology at Linköping University Hospital is acknowledged for assistance in the digitalization of the slides. The study was financed by the Swedish Childhood Cancer Fund (MT2021-0011, MT2022-0013), Joanna Cocozza's Foundation for Children’s Medical Research (2023-2024, 2025-2026), Linköping University’s Cancer Strength Area (2022, 2024), Vinnova project 2017 02447 via Medtech4Health and Analytic Imaging Diagnostics Arena (1908, 2022-2222), and ALF Grants, Region Östergötland (Ida Blystad, 974566). Additionally, co-author Ida Blystad holds a grant as an associated clinical fellow with the Wallenberg Center for Molecular Medicine.

\section*{Declaration of interests}
The authors declare no conflicts of interest.

\section*{Code and data availability}
The minimal data and codes for result interpretation can be made available on reasonable request to the corresponding authors. Weights for UNI and CONCH pre-trained models are available at \url{https://github.com/mahmoodlab/UNI} and \url{https://github.com/mahmoodlab/CONCH}, respectively. The original code for CLAM is available at \url{https://github.com/mahmoodlab/CLAM}. 

\section*{Ethics Statement and Patient Consent}
Ethical approval was obtained from the Swedish Ethical Review Authority (Dnr 2021–03985 and Dnr 2022-00065-02). The study was additionally approved by BTB, as well as Karolinska University Hospital and Stockholm Medical Biobank, the medico-legal authority for BTB's personal data and tissue samples, respectively. 

\printbibliography

\makeatletter
\renewcommand \thesection{S\@arabic\c@section}
\renewcommand\thetable{S\@arabic\c@table}
\renewcommand \thefigure{S\@arabic\c@figure}
\makeatother

\setcounter{table}{0}
\setcounter{figure}{0} 

\section*{Supplementary materials}

\subsection*{Tissue area distribution over tumor entities}
\begin{figure}[ht!]
    \centering
        \includegraphics[width=\textwidth]{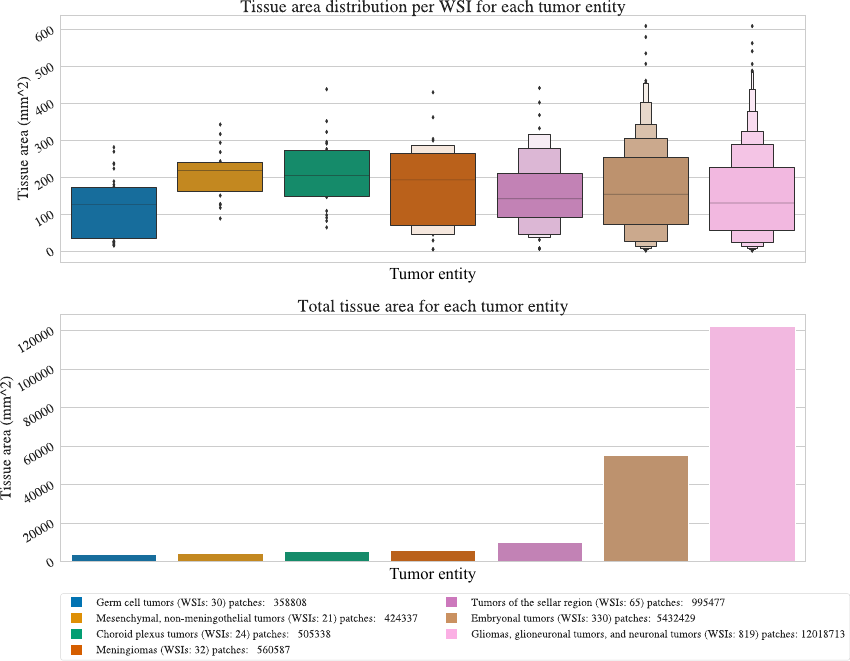}
    \caption{Tissue area distribution per WSI and cumulative tissue area for each tumor category included in the analysis.}
    \label{fig:tissue_area_stats_tumor_category}
\end{figure}
\begin{figure}[ht!]
    \centering
        \includegraphics[width=\textwidth]{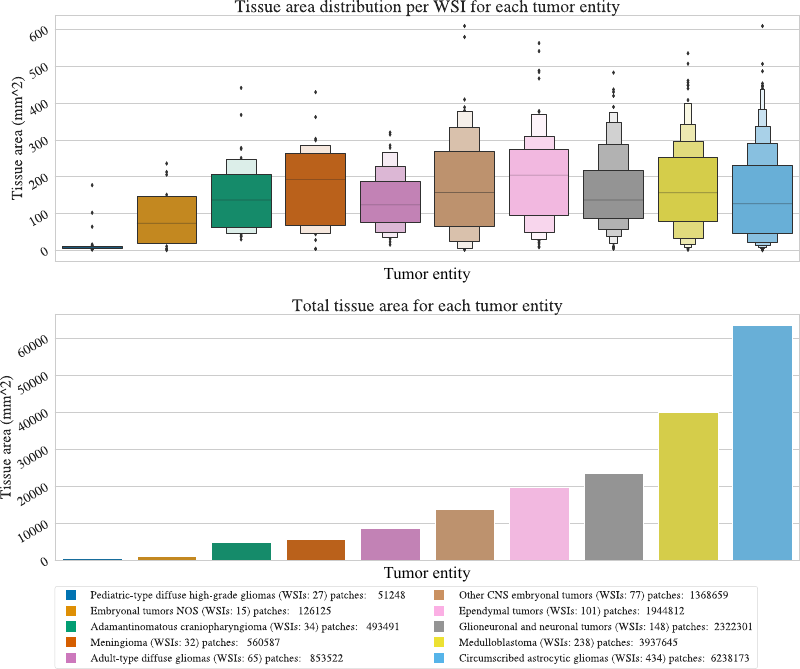}
    \caption{Tissue area distribution per WSI and cumulative tissue area for each tumor family included in the analysis.}
    \label{fig:tissue_area_stats_tumor_family}
\end{figure}
\begin{figure}[ht!]
    \centering
        \includegraphics[width=\textwidth]{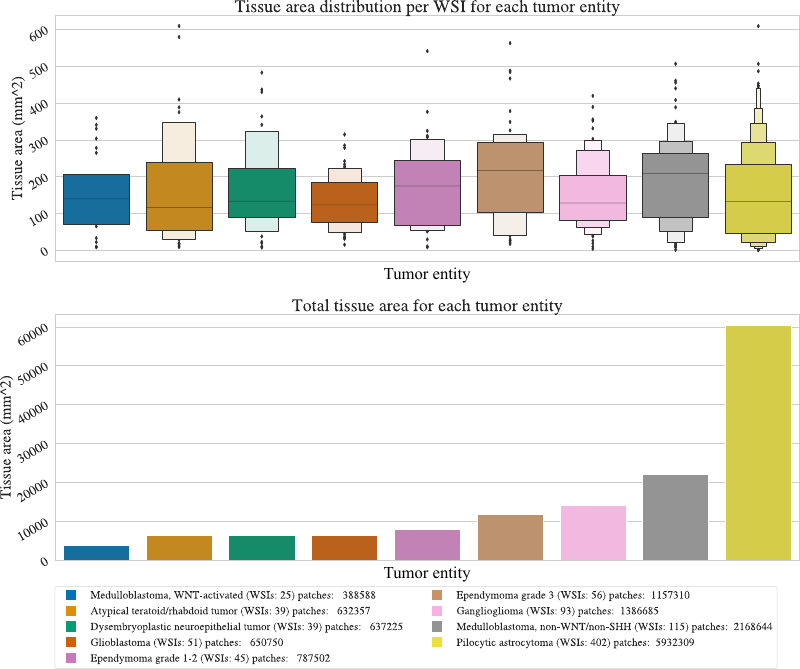}
    \caption{Tissue area distribution per WSI and cumulative tissue area for each tumor type included in the analysis.}
    \label{fig:tissue_area_stats_tumor_type}
\end{figure}

\clearpage

\subsection*{Results}
\subsubsection*{Class-wise F1 scores}

\begin{table}[ht!]
\small
\centering
\caption{Class-wise F1 scores obtained from the best model (UNI instance-level features and ABMIL aggregation) for all the classification tasks. F1 scores are presented as mean and standard deviation with 95\% confidence intervals (CI) computed over 150 model runs.}\label{tab:f1_score_all_classification}
\begin{NiceTabular}{m[c]{31mm} m[c]{16mm} m[c]{31mm} m[c]{16mm} m[c]{31mm} m[c]{16mm}}[]
\toprule
\Block{1-2}{\textbf{Tumor Category}} & & \Block{1-2}{\textbf{Tumor Family}} & & \Block{1-2}{\textbf{Tumor Type}} \\
\Block{1-1}{\textbf{Class}} & \Block{1-1}{\textbf{F1 score}\\mean$\pm$std\\ [95\% CI]} & \Block{1-1}{\textbf{Class}} & \Block{1-1}{\textbf{F1 score}\\mean$\pm$std\\ [95\% CI]} & \Block{1-1}{\textbf{Class}} & \Block{1-1}{\textbf{F1 score}\\mean$\pm$std\\ [95\% CI]} \\
\midrule
\Block{7-1}{Gliomas, glioneuronal tumors, and neuronal tumors} & \Block{7-1}{0.93$\pm$0.01\\ [0.93, 0.93]} & \Block{1-1}{Circumscribed astrocytic gliomas} & \Block{1-1}{0.81$\pm$0.04\\ [0.80, 0.81]} & \Block{1-1}{Pilocytic astrocytoma} & \Block{1-1}{0.84$\pm$0.03\\ [0.83, 0.84]}\\
\cmidrule{3-6}
  &   & \Block{2-1}{Glioneuronal and neuronal tumors} & \Block{2-1}{0.57$\pm$0.08\\ [0.56, 0.58]} & \Block{1-1}{Ganglioglioma} & \Block{1-1}{0.45$\pm$0.11\\ [0.43, 0.47]}\\
\cmidrule{5-6}
  &   &   &   & \Block{1-1}{Dysembryoplastic neuroepithelial tumor} & \Block{1-1}{0.56$\pm$0.15\\ [0.53, 0.58]}\\
\cmidrule{3-6}
  &   & \Block{2-1}{Ependymal tumors} & \Block{2-1}{0.78$\pm$0.07\\ [0.77, 0.79]} & \Block{1-1}{Ependymoma grade 1-2} & \Block{1-1}{0.41$\pm$0.17\\ [0.39, 0.44]}\\
\cmidrule{5-6}
  &   &   &   & \Block{1-1}{Ependymoma grade 3} & \Block{1-1}{0.66$\pm$0.12\\ [0.64, 0.68]}\\
\cmidrule{3-6}
  &   & \Block{1-1}{Adult-type diffuse gliomas} & \Block{1-1}{0.14$\pm$0.14\\ [0.12, 0.16]} & \Block{1-1}{Glioblastoma} & \Block{1-1}{0.27$\pm$0.19\\ [0.24, 0.30]}\\
\cmidrule{3-6}
  &   & \Block{1-1}{Pediatric-type diffuse high-grade gliomas} & \Block{1-1}{0.15$\pm$0.21\\ [0.11, 0.18]} & \Block{1-1}{/} & \Block{1-1}{/}\\
\cmidrule{1-6}
\Block{4-1}{Embryonal tumors} & \Block{4-1}{0.83$\pm$0.04\\ [0.82, 0.83]} & \Block{2-1}{Medulloblastoma} & \Block{2-1}{0.83$\pm$0.05\\ [0.82, 0.83]} & \Block{1-1}{Medulloblastoma, non-WNT/non-SHH} & \Block{1-1}{0.83$\pm$0.06\\ [0.81, 0.84]}\\
\cmidrule{5-6}
  &   &   &   & \Block{1-1}{Medulloblastoma, WNT-activated} & \Block{1-1}{0.45$\pm$0.29\\ [0.41, 0.50]}\\
\cmidrule{3-6}
  &   & \Block{1-1}{Other CNS embryonal tumors} & \Block{1-1}{0.48$\pm$0.13\\ [0.46, 0.50]} & \Block{1-1}{Atypical teratoid/rhabdoid tumor} & \Block{1-1}{0.59$\pm$0.14\\ [0.57, 0.62]}\\
\cmidrule{3-6}
  &   & \Block{1-1}{Embryonal tumors NOS} & \Block{1-1}{0.20$\pm$0.24\\ [0.16, 0.24]} & \Block{1-1}{/} & \Block{1-1}{/}\\
\cmidrule{1-6}
\Block{1-1}{Tumors of the sellar region} & \Block{1-1}{0.91$\pm$0.06\\ [0.90, 0.92]} & \Block{1-1}{Adamantinomatous craniopharyngioma} & \Block{1-1}{0.96$\pm$0.08\\ [0.95, 0.98]} & \Block{1-1}{/} & \Block{1-1}{/}\\
\cmidrule{1-6}
\Block{1-1}{Meningiomas} & \Block{1-1}{0.65$\pm$0.25\\ [0.61, 0.69]} & \Block{1-1}{Meningioma} & \Block{1-1}{0.66$\pm$0.20\\ [0.63, 0.69]} & \Block{1-1}{/} & \Block{1-1}{/}\\
\cmidrule{1-6}
\Block{1-1}{Germ cell tumors} & \Block{1-1}{0.10$\pm$0.18\\ [0.07, 0.13]} & \Block{1-1}{/} & \Block{1-1}{/} & \Block{1-1}{/} & \Block{1-1}{/}\\
\cmidrule{1-6}
\Block{1-1}{Choroid plexus tumors} & \Block{1-1}{0.61$\pm$0.23\\ [0.57, 0.64]} & \Block{1-1}{/} & \Block{1-1}{/} & \Block{1-1}{/} & \Block{1-1}{/}\\
\cmidrule{1-6}
\Block{1-1}{Mesenchymal, non-meningothelial tumors} & \Block{1-1}{0.11$\pm$0.17\\ [0.08, 0.14]} & \Block{1-1}{/} & \Block{1-1}{/} & \Block{1-1}{/} & \Block{1-1}{/}\\
\bottomrule
\end{NiceTabular}
\end{table}
\begin{table}[h!]
\small
\centering
\caption{Class-wise F1 scores obtained from the best model (UNI instance-level features and ABMIL aggregation) for gliomas, glioneuronal and neuronal tumors category classification only. F1 scores are presented as mean and standard deviation with 95\% confidence intervals (CI) computed over 150 model runs.}\label{tab:f1_score_GNN_category_only}
\begin{NiceTabular}{m[c]{50mm} m[c]{60mm} m[c]{16mm}}[]
\toprule
\Block{2-1}{\textbf{Tumor Category}} & \Block{1-2}{\textbf{Tumor Family}} \\
\Block{1-1}{} & \Block{1-1}{\textbf{Class}} & \Block{1-1}{\textbf{F1 score}\\mean$\pm$std\\ [95\% CI]} \\
\midrule
\Block{5-1}{Gliomas, glioneuronal tumors, and neuronal tumors} & \Block{1-1}{Circumscribed astrocytic gliomas} & \Block{1-1}{0.83$\pm$0.03\\ [0.82, 0.83]}\\
\cmidrule{2-3}
    & \Block{1-1}{Glioneuronal and neuronal tumors} & \Block{1-1}{0.61$\pm$0.07\\ [0.60, 0.62]}\\
\cmidrule{2-3}
    & \Block{1-1}{Ependymal tumors} & \Block{1-1}{0.81$\pm$0.07\\ [0.80, 0.82]}\\
\cmidrule{2-3}
    & \Block{1-1}{Adult-type diffuse gliomas} & \Block{1-1}{0.29$\pm$0.18\\ [0.27, 0.32]}\\
\cmidrule{2-3}
    & \Block{1-1}{Pediatric-type diffuse high-grade gliomas} & \Block{1-1}{0.15$\pm$0.23\\ [0.11, 0.19]}\\
\bottomrule
\end{NiceTabular}
\end{table}

\begin{table}[h!]
\small
\centering
\caption{Class-wise F1 scores obtained from the best model (UNI instance-level features and ABMIL aggregation) for embryonal tumors category classification only. F1 scores are presented as mean and standard deviation with 95\% confidence intervals (CI) computed over 150 model runs.}\label{tab:f1_score_ET_category_only}
\begin{NiceTabular}{m[c]{50mm} m[c]{60mm} m[c]{16mm}}[]
\toprule
\Block{2-1}{\textbf{Tumor Category}} & \Block{1-2}{\textbf{Tumor Family}} \\
\Block{1-1}{} & \Block{1-1}{\textbf{Class}} & \Block{1-1}{\textbf{F1 score}\\mean$\pm$std\\ [95\% CI]} \\
\midrule
\Block{3-1}{Embryonal tumors} & \Block{1-1}{Medulloblastoma} & \Block{1-1}{0.85$\pm$0.04\\ [0.85, 0.86]}\\
\cmidrule{2-3}
  & \Block{1-1}{Other CNS embryonal tumors} & \Block{1-1}{0.61$\pm$0.10\\ [0.60, 0.63]}\\
\cmidrule{2-3}
  & \Block{1-1}{Embryonal tumors NOS} & \Block{1-1}{0.16$\pm$0.20\\ [0.12, 0.19]}\\
\bottomrule
\end{NiceTabular}
\end{table}

\begin{figure}[ht!]
    \centering
        \includegraphics[width=\textwidth]{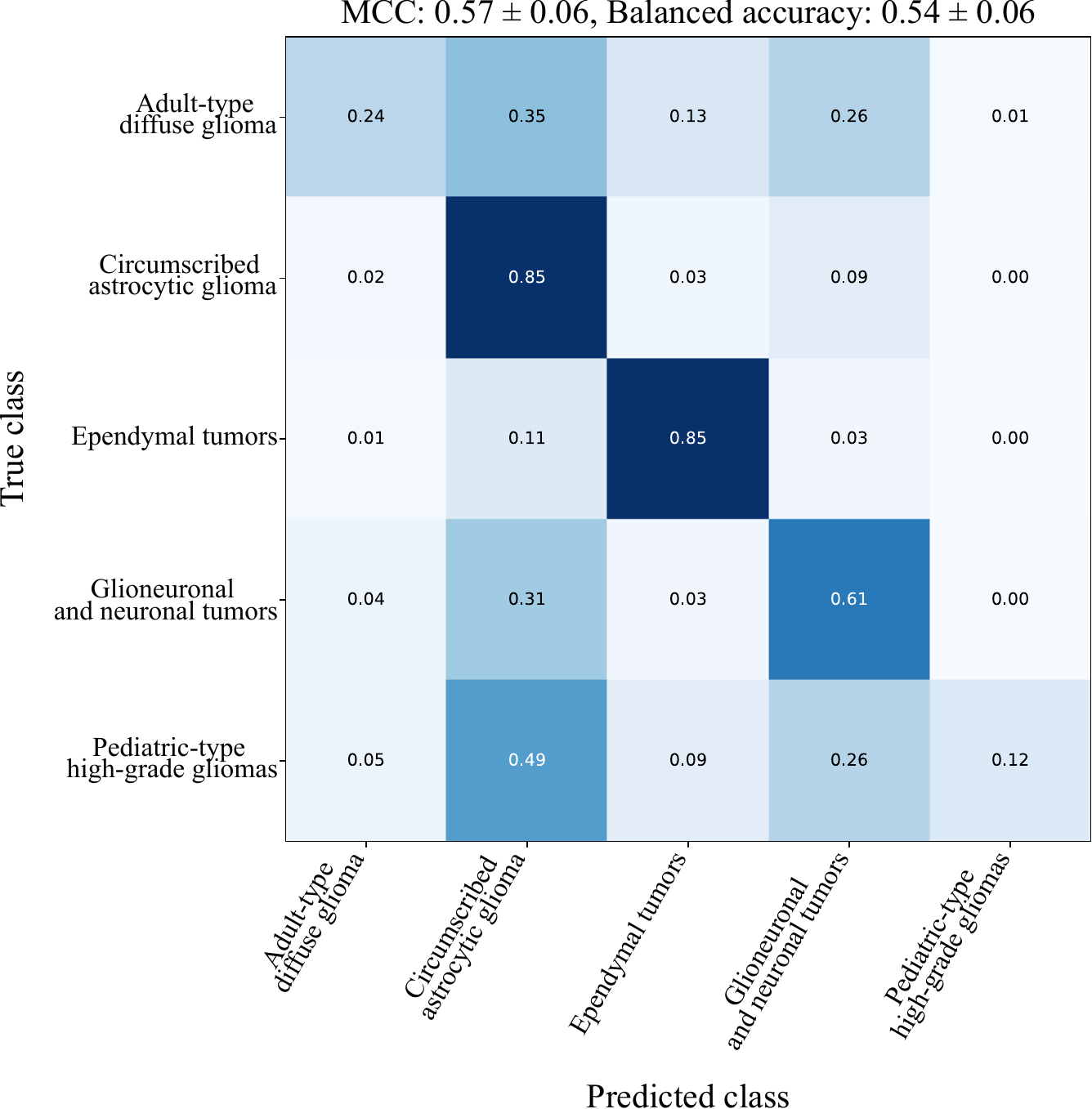}
        \caption{Average confusion matrix for the best performing model (UNI instance-level features and ABMIL aggregation) for the \textbf{gliomas/glioneuronal/neuronal tumors category-specific classifier}. The values in each cell of the confusion matrix are computed as the average cell value across 150 model runs divided by the sum of the values in each row. The resulting values describe the fraction [0, 1] of the predictions for each of the ground truth classes. MCC, Matthew’s correlation coefficient.}
        \label{fig:cm_GGN_category_only}
\end{figure}

\begin{figure}[ht!]
    \centering
        \includegraphics[width=\textwidth]{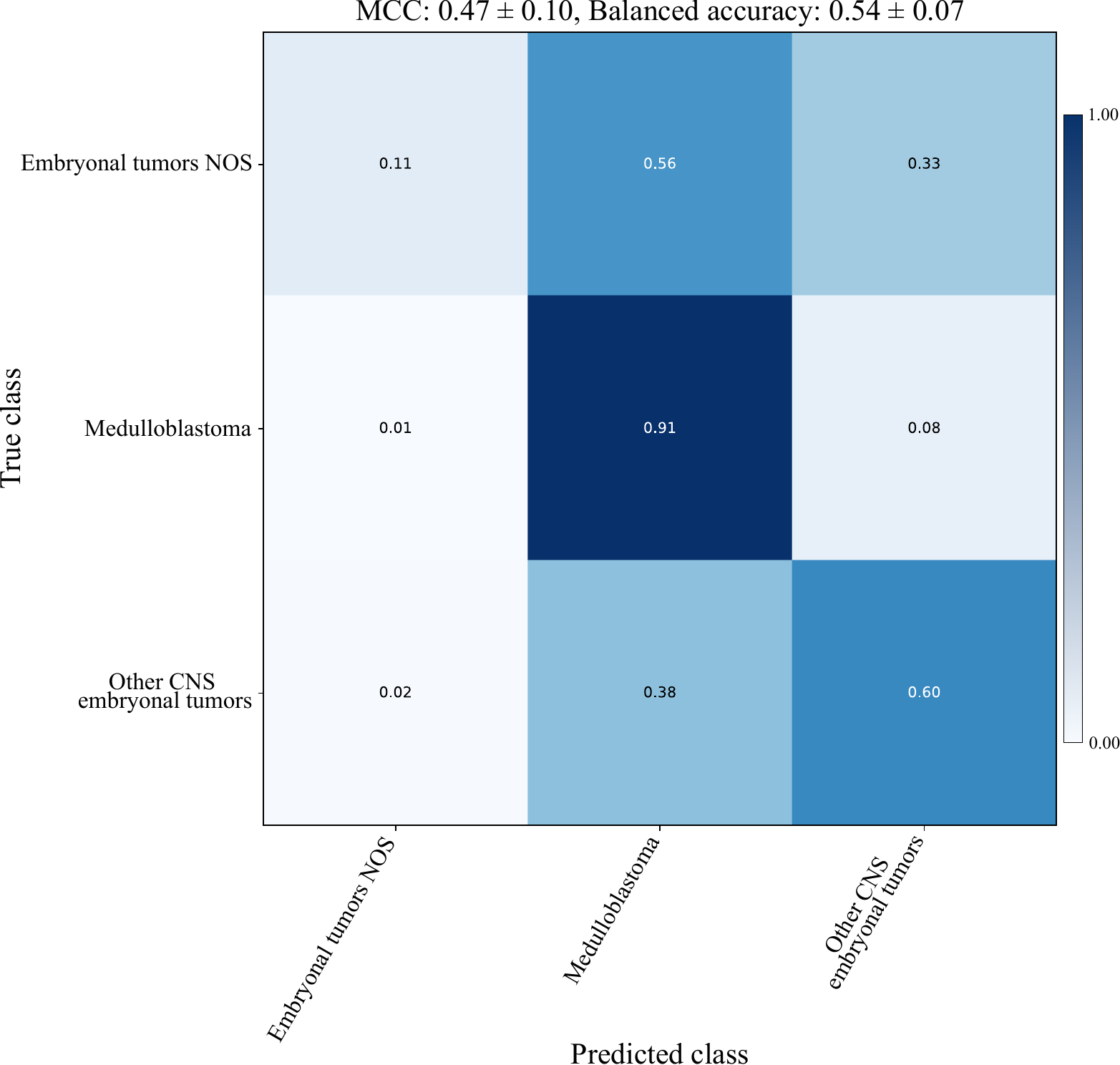}
        \caption{Average confusion matrix for the best performing model (UNI instance-level features and ABMIL aggregation) for the \textbf{embryonal tumors category-specific classifier}. The values in each cell of the confusion matrix are computed as the average cell value across 150 model runs divided by the sum of the values in each row. The resulting values describe the fraction [0, 1] of the predictions for each of the ground truth classes. MCC, Matthew’s correlation coefficient.}
        \label{fig:cm_ET_category_only}
\end{figure}

\clearpage

\subsubsection*{Cumulative confusion matrices}
\begin{figure}[ht!]
    \centering
        \includegraphics[width=\textwidth]{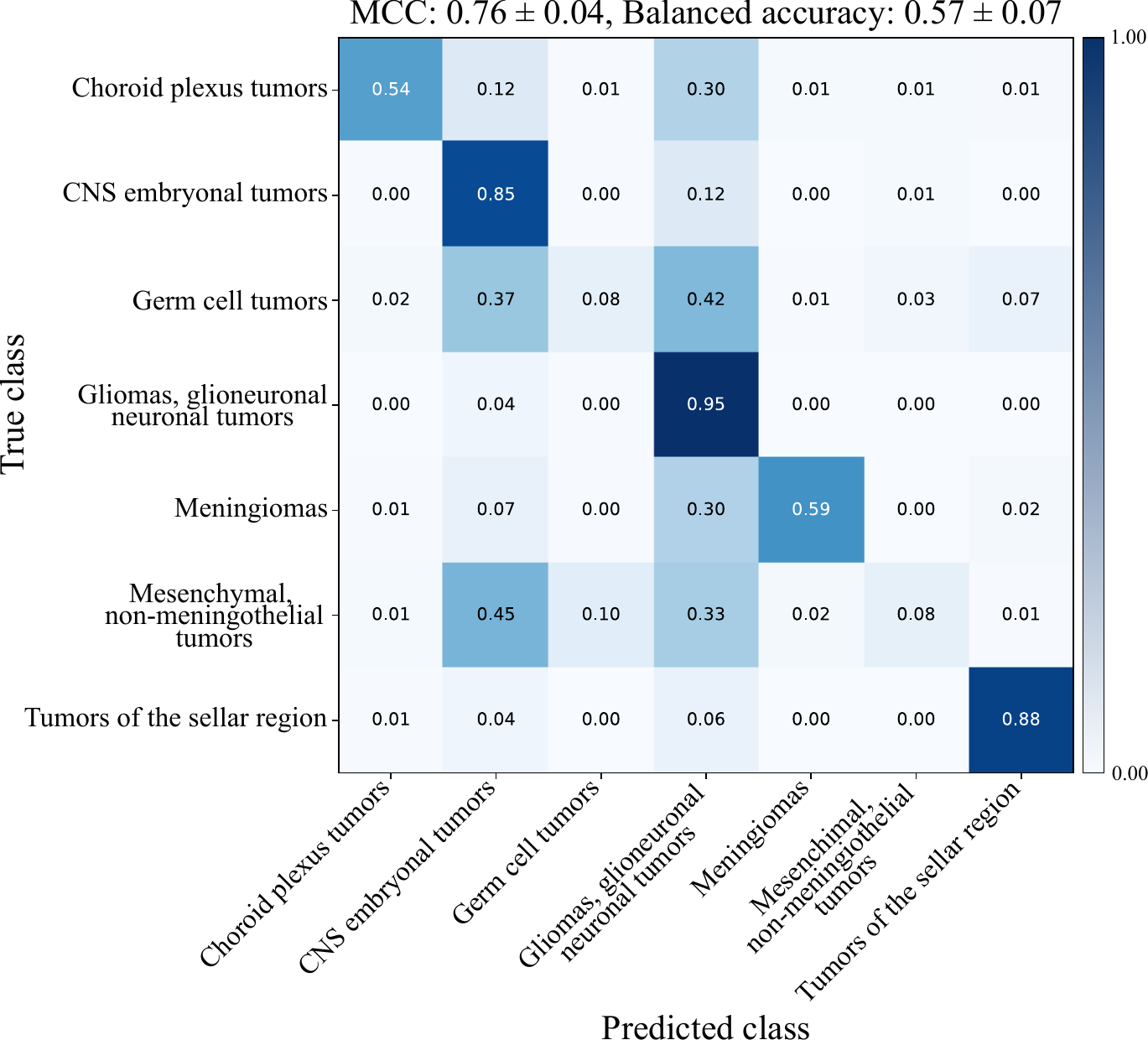}
        \caption{Average confusion matrix for the best performing model (UNI instance-level features and ABMIL aggregation) for the tumor category classification. The values in each cell of the confusion matrix are computed as the average cell value across 150 model runs divided by the sum of the values in each row. The resulting values describe the fraction [0, 1] of the predictions for each of the ground truth classes. MCC, Matthew’s correlation coefficient.}
        \label{fig:cm_tumor_category}
\end{figure}

\begin{figure}[ht!]
    \centering
        \includegraphics[width=\textwidth]{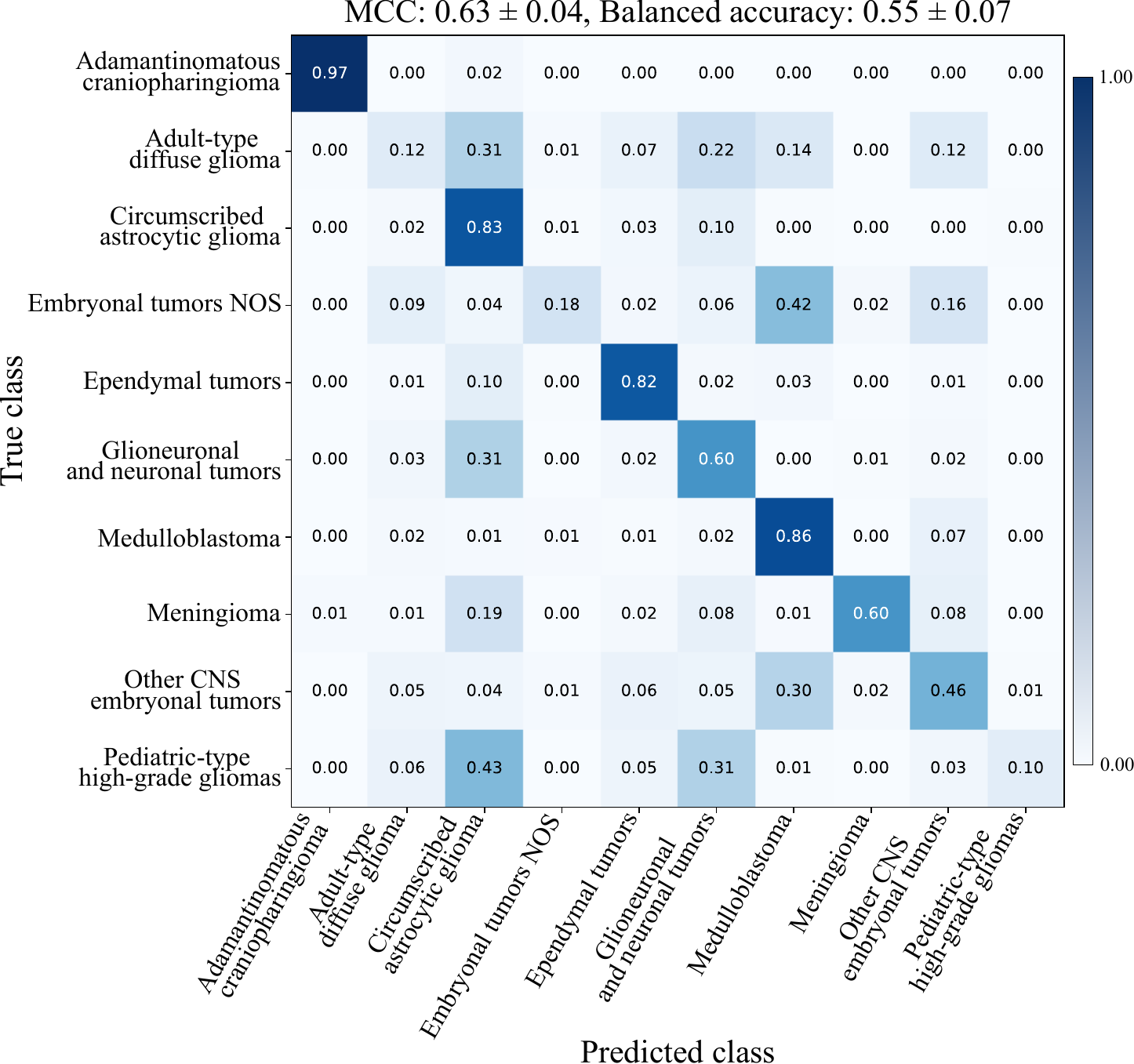}
        \caption{Average confusion matrix for the best performing model (UNI instance-level features and ABMIL aggregation) for the tumor family classification. The values in each cell of the confusion matrix are computed as the average cell value across 150 model runs divided by the sum of the values in each row. The resulting values describe the fraction [0, 1] of the predictions for each of the ground truth classes. MCC: Matthew’s correlation coefficient.}
        \label{fig:cm_tumor_family}
\end{figure}

\begin{figure}[ht!]
    \centering
        \includegraphics[width=\textwidth]{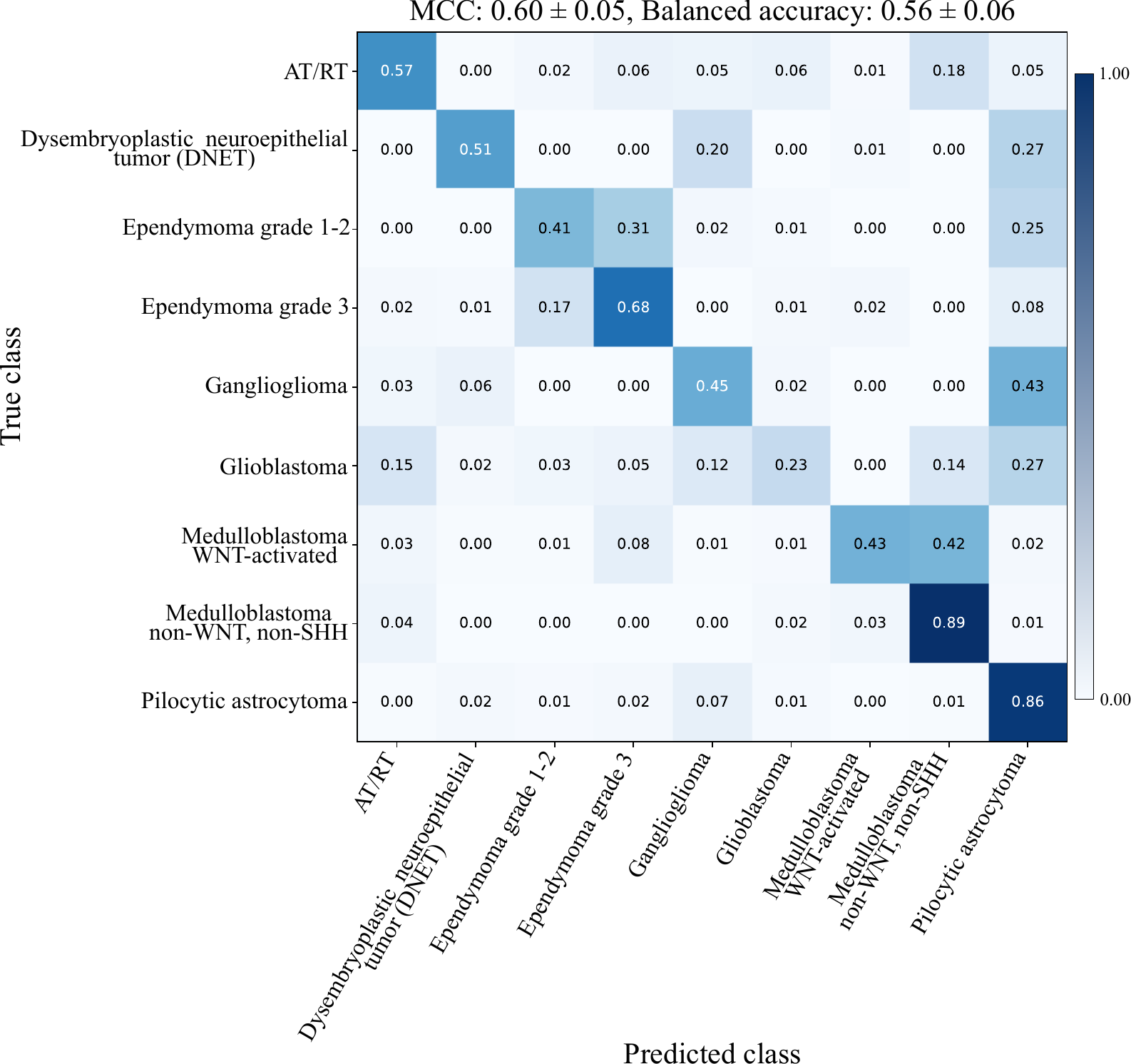}
        \caption{Average confusion matrix for the best performing model (UNI instance-level features and ABMIL aggregation) for the tumor type classification. The values in each cell of the confusion matrix are computed as the average cell value across 150 model runs divided by the sum of the values in each row. The resulting values describe the fraction [0, 1] of the predictions for each of the ground truth classes. MCC: Matthew’s correlation coefficient.}
        \label{fig:cm_tumor_type}
\end{figure}

\clearpage

\subsubsection*{Additional examples of attention maps}
\begin{figure}[ht!]
    \centering
        \includegraphics[width=\textwidth]{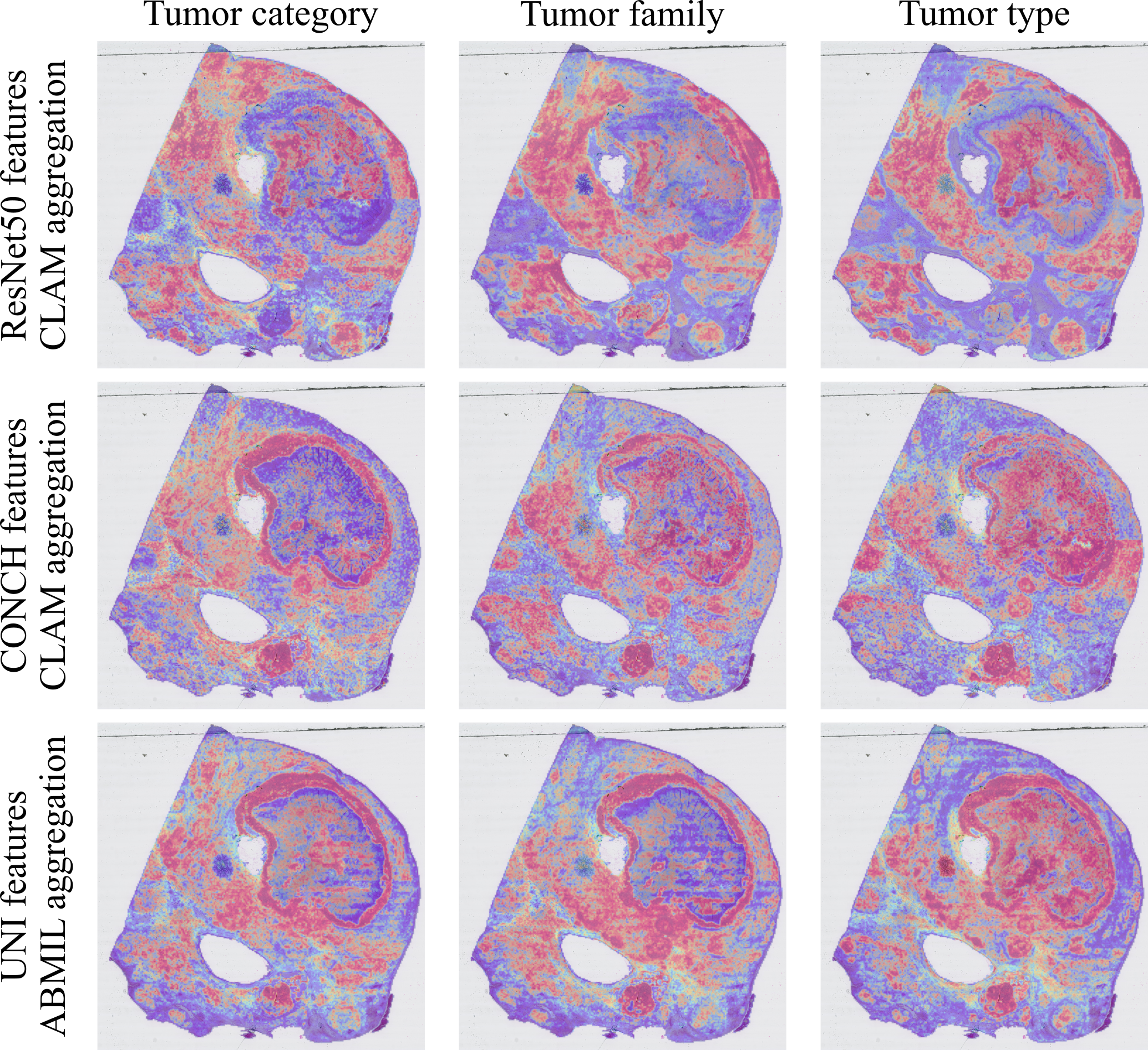}
        \caption{Attention maps for the best-performing aggregation method across feature extractors and the three classification tasks. Attention scores are obtained from the model whose performance is closest to the median value across the 150 replicates. Warm colors identify regions that mostly contribute to the classification. Models predicted correctly the example, which is a case diagnosed as ependymoma grade 3 tumor.}
        \label{fig:sup_attention_ependymoma_g3}
\end{figure}

\begin{figure}[ht!]
    \centering
        \includegraphics[width=\textwidth]{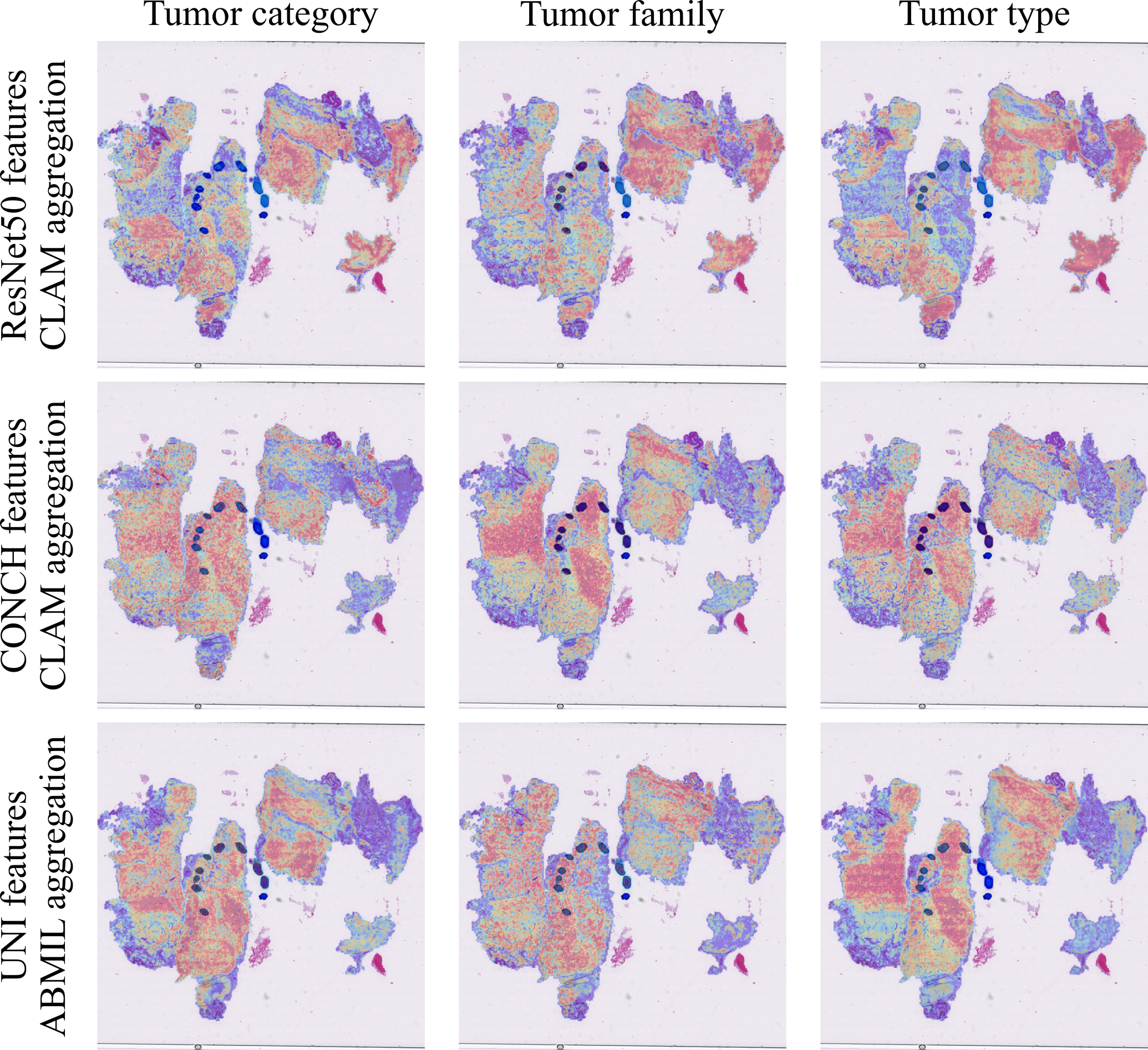}
        \caption{Attention maps for the best-performing aggregation method across feature extractors and the three classification tasks. Attention scores are obtained from the model whose performance is closest to the median value across the 150 replicates. Warm colors identify regions that mostly contribute to the classification. Models predicted correctly the example, which is a case diagnosed as pilocytic astrocytoma.}
     \label{fig:sup_attention_pilocytic_astrocytoma_1}
\end{figure}

\begin{figure}[ht!]
    \centering
        \includegraphics[width=\textwidth]{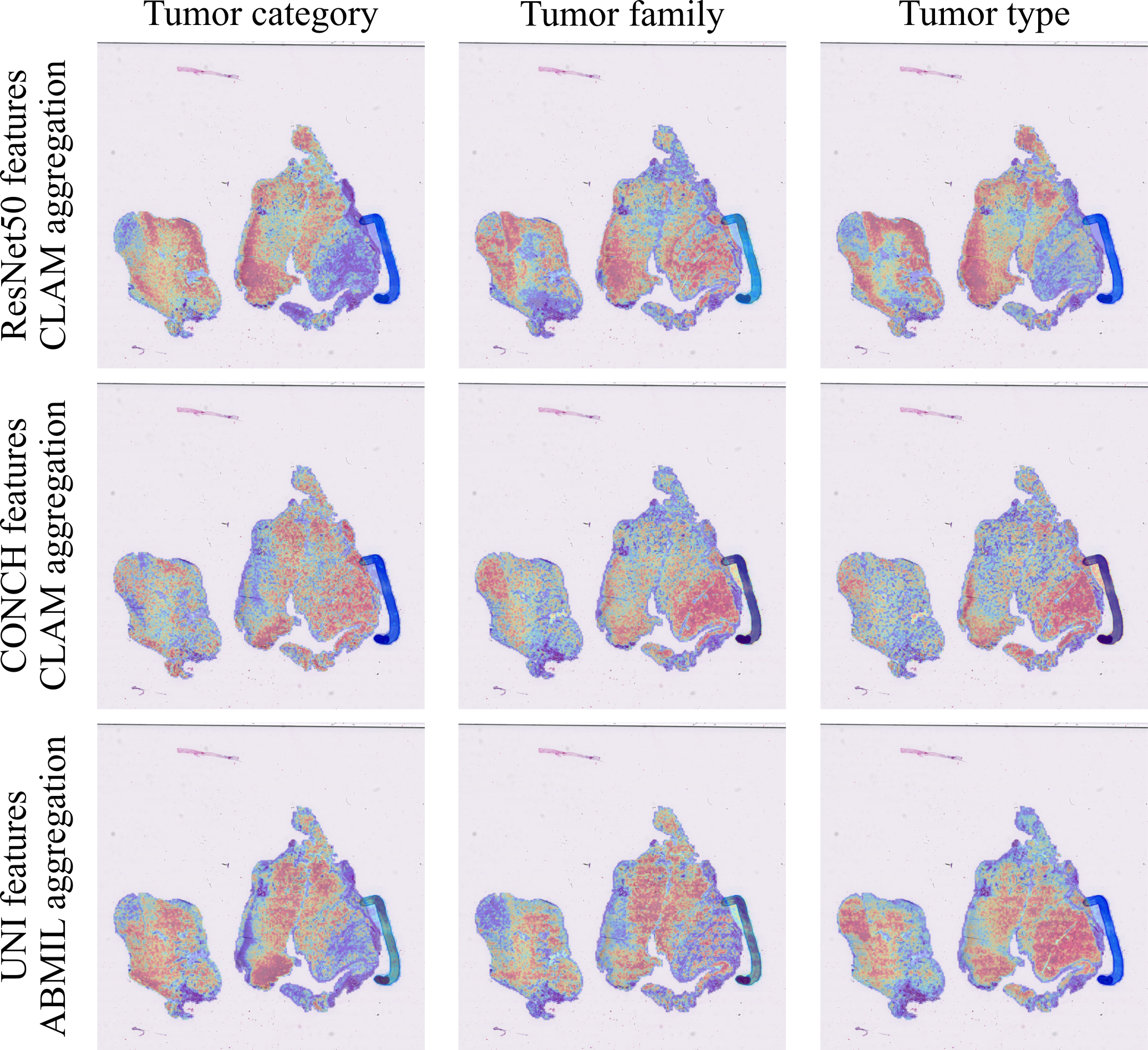}
    \caption{Attention maps for the best-performing aggregation method across feature extractors and the three classification tasks. Attention scores are obtained from the model whose performance is closest to the median value across the 150 replicates. Warm colors identify regions that mostly contribute to the classification. Models predicted correctly the example, which is a case diagnosed as pilocytic astrocytoma.}
    \label{fig:sup_attention_pilocytic_astrocytoma_2}
\end{figure}

\clearpage

\subsubsection*{Model generalization detailed results with a minimum of 10 cases}
\begin{table}[!h]
\small
\centering
\caption{Classification performance when evaluating on \textbf{in-site data} (two sites), with a minimum number of cases set to 10 for class inclusion. Metrics are presented as mean and standard deviation with 95\% confidence intervals (CI) computed over 5 model runs. Abbreviations: ABMIL, attention-based multiple-instance learning; CLAM, clustering-constrained attention multiple instance-learning, MCC, Matthew's correlation coefficient; ResNet50, convolutional neural network pretrained on ImageNet; UNI and CONCH, vision transformer foundation models pretrained on histopathology data.}\label{tab:classification_performance_generalization_in_site_data_min_10}
\begin{NiceTabular}{p[c]{19mm} p[c]{19mm} p[c]{19mm} p[c]{19mm} m[c]{19mm} p[c]{19mm} p[c]{19mm}}[]
\toprule
\Block{1-1}{Classification\\granularity} & \Block{1-1}{Instance feature\\extractor} & \Block{1-1}{MIL aggregation\\method} & \Block{1-1}{MCC\\mean$\pm$std\\ [95\% CI]} & \Block{1-1}{Balanced\\accuracy\\ mean$\pm$std\\ [95\% CI]} & \Block{1-1}{F1 score\\ mean$\pm$std\\ [95\% CI]} & \Block{1-1}{AUROC\\ mean$\pm$std\\ [95\% CI]} \\
\midrule
\Block{6-1}{Tumor Category} & \Block{2-1}{ResNet50} & \Block{1-1}{ABMIL} & \Block{1-1}{0.42$\pm$0.09\\ [0.30, 0.55]} & \Block{1-1}{0.67$\pm$0.07\\ [0.58, 0.77]} & \Block{1-1}{0.54$\pm$0.07\\ [0.45, 0.64]} & \Block{1-1}{0.83$\pm$0.04\\ [0.77, 0.89]}\\
\cmidrule{3-7}
  &   & \Block{1-1}{CLAM} & \Block{1-1}{0.45$\pm$0.06\\ [0.36, 0.53]} & \Block{1-1}{0.69$\pm$0.07\\ [0.59, 0.80]} & \Block{1-1}{0.56$\pm$0.05\\ [0.49, 0.64]} & \Block{1-1}{0.85$\pm$0.04\\ [0.80, 0.90]}\\
\cmidrule{2-7}
  & \Block{2-1}{CONCH} & \Block{1-1}{ABMIL} & \Block{1-1}{0.78$\pm$0.04\\ [0.73, 0.83]} & \Block{1-1}{0.88$\pm$0.03\\ [0.84, 0.93]} & \Block{1-1}{0.88$\pm$0.02\\ [0.85, 0.90]} & \Block{1-1}{0.97$\pm$0.01\\ [0.96, 0.99]}\\
\cmidrule{3-7}
  &   & \Block{1-1}{CLAM} & \Block{1-1}{0.80$\pm$0.07\\ [0.70, 0.90]} & \Block{1-1}{0.89$\pm$0.05\\ [0.82, 0.95]} & \Block{1-1}{0.88$\pm$0.06\\ [0.80, 0.96]} & \Block{1-1}{0.97$\pm$0.01\\ [0.96, 0.99]}\\
\cmidrule{2-7}
  & \Block{2-1}{UNI} & \Block{1-1}{ABMIL} & \Block{1-1}{0.81$\pm$0.04\\ [0.75, 0.87]} & \Block{1-1}{0.87$\pm$0.02\\ [0.84, 0.90]} & \Block{1-1}{0.89$\pm$0.02\\ [0.86, 0.92]} & \Block{1-1}{0.98$\pm$0.01\\ [0.97, 0.99]}\\
\cmidrule{3-7}
  &   & \Block{1-1}{CLAM} & \Block{1-1}{0.81$\pm$0.04\\ [0.76, 0.86]} & \Block{1-1}{0.87$\pm$0.02\\ [0.84, 0.91]} & \Block{1-1}{0.89$\pm$0.02\\ [0.86, 0.92]} & \Block{1-1}{0.98$\pm$0.01\\ [0.97, 0.98]}\\
\cmidrule{1-7}
\Block[fill=gray!10]{6-1}{Tumor Family} & \Block[fill=gray!10]{2-1}{ResNet50} & \Block[fill=gray!10]{1-1}{ABMIL} & \Block[fill=gray!10]{1-1}{0.28$\pm$0.04\\ [0.23, 0.34]} & \Block[fill=gray!10]{1-1}{0.40$\pm$0.03\\ [0.36, 0.44]} & \Block[fill=gray!10]{1-1}{0.29$\pm$0.04\\ [0.25, 0.34]} & \Block[fill=gray!10]{1-1}{0.77$\pm$0.02\\ [0.74, 0.79]}\\
\cmidrule{3-7}
  &   & \Block[fill=gray!10]{1-1}{CLAM} & \Block[fill=gray!10]{1-1}{0.31$\pm$0.10\\ [0.17, 0.44]} & \Block[fill=gray!10]{1-1}{0.43$\pm$0.09\\ [0.31, 0.54]} & \Block[fill=gray!10]{1-1}{0.34$\pm$0.12\\ [0.17, 0.52]} & \Block[fill=gray!10]{1-1}{0.77$\pm$0.02\\ [0.75, 0.79]}\\
\cmidrule{2-7}
  & \Block[fill=gray!10]{2-1}{CONCH} & \Block[fill=gray!10]{1-1}{ABMIL} & \Block[fill=gray!10]{1-1}{0.60$\pm$0.04\\ [0.54, 0.66]} & \Block[fill=gray!10]{1-1}{0.67$\pm$0.05\\ [0.61, 0.74]} & \Block[fill=gray!10]{1-1}{0.66$\pm$0.04\\ [0.60, 0.71]} & \Block[fill=gray!10]{1-1}{0.90$\pm$0.02\\ [0.88, 0.92]}\\
\cmidrule{3-7}
  &   & \Block[fill=gray!10]{1-1}{CLAM} & \Block[fill=gray!10]{1-1}{0.65$\pm$0.02\\ [0.62, 0.68]} & \Block[fill=gray!10]{1-1}{0.72$\pm$0.02\\ [0.68, 0.75]} & \Block[fill=gray!10]{1-1}{0.70$\pm$0.03\\ [0.66, 0.74]} & \Block[fill=gray!10]{1-1}{0.91$\pm$0.02\\ [0.89, 0.93]}\\
\cmidrule{2-7}
  & \Block[fill=gray!10]{2-1}{UNI} & \Block[fill=gray!10]{1-1}{ABMIL} & \Block[fill=gray!10]{1-1}{0.67$\pm$0.05\\ [0.60, 0.74]} & \Block[fill=gray!10]{1-1}{0.72$\pm$0.05\\ [0.65, 0.79]} & \Block[fill=gray!10]{1-1}{0.71$\pm$0.06\\ [0.63, 0.79]} & \Block[fill=gray!10]{1-1}{0.93$\pm$0.00\\ [0.93, 0.94]}\\
\cmidrule{3-7}
  &   & \Block[fill=gray!10]{1-1}{CLAM} & \Block[fill=gray!10]{1-1}{0.68$\pm$0.07\\ [0.58, 0.78]} & \Block[fill=gray!10]{1-1}{0.73$\pm$0.06\\ [0.64, 0.81]} & \Block[fill=gray!10]{1-1}{0.72$\pm$0.07\\ [0.62, 0.81]} & \Block[fill=gray!10]{1-1}{0.92$\pm$0.01\\ [0.90, 0.94]}\\
\cmidrule{1-7}
\Block{6-1}{Tumor Type} & \Block{2-1}{ResNet50} & \Block{1-1}{ABMIL} & \Block{1-1}{0.40$\pm$0.13\\ [0.23, 0.58]} & \Block{1-1}{0.56$\pm$0.04\\ [0.51, 0.61]} & \Block{1-1}{0.43$\pm$0.09\\ [0.30, 0.56]} & \Block{1-1}{0.73$\pm$0.02\\ [0.70, 0.76]}\\
\cmidrule{3-7}
  &   & \Block{1-1}{CLAM} & \Block{1-1}{0.39$\pm$0.09\\ [0.27, 0.51]} & \Block{1-1}{0.58$\pm$0.07\\ [0.48, 0.67]} & \Block{1-1}{0.46$\pm$0.10\\ [0.32, 0.60]} & \Block{1-1}{0.77$\pm$0.03\\ [0.72, 0.82]}\\
\cmidrule{2-7}
  & \Block{2-1}{CONCH} & \Block{1-1}{ABMIL} & \Block{1-1}{0.67$\pm$0.04\\ [0.62, 0.72]} & \Block{1-1}{0.78$\pm$0.03\\ [0.75, 0.82]} & \Block{1-1}{0.77$\pm$0.03\\ [0.74, 0.81]} & \Block{1-1}{0.94$\pm$0.01\\ [0.93, 0.96]}\\
\cmidrule{3-7}
  &   & \Block{1-1}{CLAM} & \Block{1-1}{0.69$\pm$0.06\\ [0.61, 0.77]} & \Block{1-1}{0.79$\pm$0.05\\ [0.72, 0.86]} & \Block{1-1}{0.79$\pm$0.04\\ [0.72, 0.85]} & \Block{1-1}{0.95$\pm$0.01\\ [0.93, 0.97]}\\
\cmidrule{2-7}
  & \Block{2-1}{UNI} & \Block{1-1}{ABMIL} & \Block{1-1}{0.61$\pm$0.12\\ [0.45, 0.78]} & \Block{1-1}{0.73$\pm$0.10\\ [0.59, 0.86]} & \Block{1-1}{0.72$\pm$0.10\\ [0.58, 0.86]} & \Block{1-1}{0.90$\pm$0.02\\ [0.87, 0.93]}\\
\cmidrule{3-7}
  &   & \Block{1-1}{CLAM} & \Block{1-1}{0.60$\pm$0.10\\ [0.46, 0.74]} & \Block{1-1}{0.69$\pm$0.09\\ [0.56, 0.81]} & \Block{1-1}{0.68$\pm$0.11\\ [0.53, 0.83]} & \Block{1-1}{0.89$\pm$0.03\\ [0.85, 0.93]}\\
\bottomrule
\end{NiceTabular}
\end{table}
\begin{table}[!h]
\small
\centering
\caption{Classification performance when evaluating on \textbf{out-of-site data} (4 sites), with a minimum number of cases set to 10 for class inclusion. Metrics are presented as mean and standard deviation with 95\% confidence intervals (CI) computed over 5 model runs. Abbreviations: ABMIL, attention-based multiple-instance learning; CLAM, clustering-constrained attention multiple instance-learning, MCC, Matthew's correlation coefficient; ResNet50, convolutional neural network pretrained on ImageNet; UNI and CONCH, vision transformer foundation models pretrained on histopathology data.}\label{tab:classification_performance_generalization_out_of_site_data_min_10}
\begin{NiceTabular}{p[c]{19mm} p[c]{19mm} p[c]{19mm} p[c]{19mm} m[c]{19mm} p[c]{19mm} p[c]{19mm}}[]
\toprule
\Block{1-1}{Classification\\granularity} & \Block{1-1}{Instance feature\\extractor} & \Block{1-1}{MIL aggregation\\method} & \Block{1-1}{MCC\\mean$\pm$std\\ [95\% CI]} & \Block{1-1}{Balanced\\accuracy\\ mean$\pm$std\\ [95\% CI]} & \Block{1-1}{F1 score\\ mean$\pm$std\\ [95\% CI]} & \Block{1-1}{AUROC\\ mean$\pm$std\\ [95\% CI]} \\
\midrule
\Block{6-1}{Tumor Category} & \Block{2-1}{ResNet50} & \Block{1-1}{ABMIL} & \Block{1-1}{0.51$\pm$0.08\\ [0.40, 0.61]} & \Block{1-1}{0.74$\pm$0.04\\ [0.68, 0.80]} & \Block{1-1}{0.60$\pm$0.06\\ [0.52, 0.69]} & \Block{1-1}{0.88$\pm$0.05\\ [0.82, 0.95]}\\
\cmidrule{3-7}
  &   & \Block{1-1}{CLAM} & \Block{1-1}{0.54$\pm$0.07\\ [0.44, 0.64]} & \Block{1-1}{0.76$\pm$0.07\\ [0.66, 0.86]} & \Block{1-1}{0.62$\pm$0.06\\ [0.54, 0.71]} & \Block{1-1}{0.89$\pm$0.05\\ [0.81, 0.96]}\\
\cmidrule{2-7}
  & \Block{2-1}{CONCH} & \Block{1-1}{ABMIL} & \Block{1-1}{0.79$\pm$0.05\\ [0.73, 0.86]} & \Block{1-1}{0.89$\pm$0.04\\ [0.83, 0.94]} & \Block{1-1}{0.87$\pm$0.03\\ [0.83, 0.91]} & \Block{1-1}{0.98$\pm$0.01\\ [0.97, 0.99]}\\
\cmidrule{3-7}
  &   & \Block{1-1}{CLAM} & \Block{1-1}{0.82$\pm$0.02\\ [0.79, 0.86]} & \Block{1-1}{0.92$\pm$0.02\\ [0.90, 0.94]} & \Block{1-1}{0.89$\pm$0.01\\ [0.87, 0.91]} & \Block{1-1}{0.99$\pm$0.00\\ [0.98, 0.99]}\\
\cmidrule{2-7}
  & \Block{2-1}{UNI} & \Block{1-1}{ABMIL} & \Block{1-1}{0.85$\pm$0.02\\ [0.82, 0.87]} & \Block{1-1}{0.92$\pm$0.01\\ [0.91, 0.93]} & \Block{1-1}{0.92$\pm$0.01\\ [0.91, 0.93]} & \Block{1-1}{0.98$\pm$0.00\\ [0.98, 0.99]}\\
\cmidrule{3-7}
  &   & \Block{1-1}{CLAM} & \Block{1-1}{0.84$\pm$0.04\\ [0.79, 0.90]} & \Block{1-1}{0.91$\pm$0.03\\ [0.86, 0.95]} & \Block{1-1}{0.91$\pm$0.02\\ [0.89, 0.94]} & \Block{1-1}{0.99$\pm$0.01\\ [0.98, 0.99]}\\
\cmidrule{1-7}
\Block[fill=gray!10]{6-1}{Tumor Family} & \Block[fill=gray!10]{2-1}{ResNet50} & \Block[fill=gray!10]{1-1}{ABMIL} & \Block[fill=gray!10]{1-1}{0.26$\pm$0.02\\ [0.24, 0.29]} & \Block[fill=gray!10]{1-1}{0.40$\pm$0.02\\ [0.38, 0.43]} & \Block[fill=gray!10]{1-1}{0.26$\pm$0.01\\ [0.24, 0.28]} & \Block[fill=gray!10]{1-1}{0.76$\pm$0.02\\ [0.73, 0.78]}\\
\cmidrule{3-7}
  &   & \Block[fill=gray!10]{1-1}{CLAM} & \Block[fill=gray!10]{1-1}{0.29$\pm$0.06\\ [0.21, 0.37]} & \Block[fill=gray!10]{1-1}{0.41$\pm$0.06\\ [0.33, 0.49]} & \Block[fill=gray!10]{1-1}{0.30$\pm$0.08\\ [0.19, 0.41]} & \Block[fill=gray!10]{1-1}{0.76$\pm$0.04\\ [0.70, 0.81]}\\
\cmidrule{2-7}
  & \Block[fill=gray!10]{2-1}{CONCH} & \Block[fill=gray!10]{1-1}{ABMIL} & \Block[fill=gray!10]{1-1}{0.58$\pm$0.03\\ [0.54, 0.62]} & \Block[fill=gray!10]{1-1}{0.59$\pm$0.02\\ [0.57, 0.62]} & \Block[fill=gray!10]{1-1}{0.58$\pm$0.03\\ [0.55, 0.62]} & \Block[fill=gray!10]{1-1}{0.90$\pm$0.03\\ [0.87, 0.94]}\\
\cmidrule{3-7}
  &   & \Block[fill=gray!10]{1-1}{CLAM} & \Block[fill=gray!10]{1-1}{0.59$\pm$0.02\\ [0.56, 0.62]} & \Block[fill=gray!10]{1-1}{0.64$\pm$0.01\\ [0.63, 0.65]} & \Block[fill=gray!10]{1-1}{0.62$\pm$0.02\\ [0.59, 0.64]} & \Block[fill=gray!10]{1-1}{0.91$\pm$0.02\\ [0.88, 0.94]}\\
\cmidrule{2-7}
  & \Block[fill=gray!10]{2-1}{UNI} & \Block[fill=gray!10]{1-1}{ABMIL} & \Block[fill=gray!10]{1-1}{0.66$\pm$0.02\\ [0.63, 0.69]} & \Block[fill=gray!10]{1-1}{0.65$\pm$0.01\\ [0.64, 0.67]} & \Block[fill=gray!10]{1-1}{0.66$\pm$0.01\\ [0.64, 0.67]} & \Block[fill=gray!10]{1-1}{0.90$\pm$0.01\\ [0.89, 0.91]}\\
\cmidrule{3-7}
  &   & \Block[fill=gray!10]{1-1}{CLAM} & \Block[fill=gray!10]{1-1}{0.66$\pm$0.03\\ [0.62, 0.69]} & \Block[fill=gray!10]{1-1}{0.66$\pm$0.01\\ [0.64, 0.68]} & \Block[fill=gray!10]{1-1}{0.66$\pm$0.01\\ [0.64, 0.67]} & \Block[fill=gray!10]{1-1}{0.89$\pm$0.01\\ [0.87, 0.91]}\\
\cmidrule{1-7}
\Block{6-1}{Tumor Type} & \Block{2-1}{ResNet50} & \Block{1-1}{ABMIL} & \Block{1-1}{0.26$\pm$0.14\\ [0.07, 0.45]} & \Block{1-1}{0.45$\pm$0.06\\ [0.36, 0.53]} & \Block{1-1}{0.27$\pm$0.09\\ [0.14, 0.40]} & \Block{1-1}{0.75$\pm$0.04\\ [0.69, 0.81]}\\
\cmidrule{3-7}
  &   & \Block{1-1}{CLAM} & \Block{1-1}{0.29$\pm$0.05\\ [0.22, 0.37]} & \Block{1-1}{0.51$\pm$0.04\\ [0.46, 0.57]} & \Block{1-1}{0.33$\pm$0.07\\ [0.24, 0.43]} & \Block{1-1}{0.76$\pm$0.04\\ [0.70, 0.82]}\\
\cmidrule{2-7}
  & \Block{2-1}{CONCH} & \Block{1-1}{ABMIL} & \Block{1-1}{0.56$\pm$0.07\\ [0.46, 0.66]} & \Block{1-1}{0.63$\pm$0.02\\ [0.60, 0.65]} & \Block{1-1}{0.59$\pm$0.04\\ [0.54, 0.64]} & \Block{1-1}{0.89$\pm$0.02\\ [0.86, 0.91]}\\
\cmidrule{3-7}
  &   & \Block{1-1}{CLAM} & \Block{1-1}{0.58$\pm$0.04\\ [0.52, 0.63]} & \Block{1-1}{0.63$\pm$0.02\\ [0.60, 0.66]} & \Block{1-1}{0.60$\pm$0.01\\ [0.58, 0.62]} & \Block{1-1}{0.88$\pm$0.02\\ [0.86, 0.90]}\\
\cmidrule{2-7}
  & \Block{2-1}{UNI} & \Block{1-1}{ABMIL} & \Block{1-1}{0.64$\pm$0.04\\ [0.58, 0.69]} & \Block{1-1}{0.63$\pm$0.03\\ [0.58, 0.68]} & \Block{1-1}{0.62$\pm$0.03\\ [0.58, 0.67]} & \Block{1-1}{0.86$\pm$0.02\\ [0.83, 0.89]}\\
\cmidrule{3-7}
  &   & \Block{1-1}{CLAM} & \Block{1-1}{0.62$\pm$0.03\\ [0.58, 0.67]} & \Block{1-1}{0.61$\pm$0.04\\ [0.55, 0.67]} & \Block{1-1}{0.60$\pm$0.04\\ [0.55, 0.65]} & \Block{1-1}{0.86$\pm$0.02\\ [0.83, 0.88]}\\
\bottomrule
\end{NiceTabular}
\end{table}

\begin{table}[ht]
\small
\centering
\caption{Class-wise F1 scores when evaluating on \textbf{in-site data} (two sites), with a minimum number of cases set to 10 for class inclusion. Metrics are presented for the best-performing model (UNI-features and ABMIL aggregation) as mean and standard deviation with 95\% confidence intervals (CI) computed over 5 model runs. }\label{tab:f1_score_generalization_in_site_data_min_10}
\begin{NiceTabular}{m[c]{31mm} m[c]{16mm} m[c]{31mm} m[c]{16mm} m[c]{31mm} m[c]{16mm}}[]
\toprule
\Block{1-2}{\textbf{Tumor Category}} & & \Block{1-2}{\textbf{Tumor Family}} & & \Block{1-2}{\textbf{Tumor Type}} \\
\Block{1-1}{\textbf{Class}} & \Block{1-1}{\textbf{F1 score}\\mean$\pm$std\\ [95\% CI]} & \Block{1-1}{\textbf{Class}} & \Block{1-1}{\textbf{F1 score}\\mean$\pm$std\\ [95\% CI]} & \Block{1-1}{\textbf{Class}} & \Block{1-1}{\textbf{F1 score}\\mean$\pm$std\\ [95\% CI]} \\
\midrule
\Block{3-1}{Gliomas, glioneuronal tumors, and neuronal tumors} & \Block{3-1}{0.94$\pm$0.01\\ [0.92, 0.95]} & \Block{1-1}{Circumscribed astrocytic gliomas} & \Block{1-1}{0.80$\pm$0.04\\ [0.75, 0.86]} & \Block{1-1}{Pilocytic astrocytoma} & \Block{1-1}{0.80$\pm$0.07\\ [0.70, 0.89]}\\
\cmidrule{3-6}
  &   & \Block{1-1}{Glioneuronal and neuronal tumors} & \Block{1-1}{0.69$\pm$0.05\\ [0.63, 0.76]} & \Block{1-1}{Ganglioglioma} & \Block{1-1}{0.56$\pm$0.12\\ [0.39, 0.73]}\\
\cmidrule{3-6}
  &   & \Block{1-1}{Ependymal tumors} & \Block{1-1}{0.83$\pm$0.09\\ [0.71, 0.96]} & \Block{1-1}{/} & \Block{1-1}{/}\\
\cmidrule{1-6}
\Block{2-1}{Embryonal tumors} & \Block{2-1}{0.84$\pm$0.05\\ [0.76, 0.91]} & \Block{1-1}{Medulloblastoma} & \Block{1-1}{0.80$\pm$0.08\\ [0.68, 0.91]} & \Block{1-1}{Medulloblastoma, non-WNT/non-SHH} & \Block{1-1}{0.88$\pm$0.04\\ [0.83, 0.94]}\\
\cmidrule{3-6}
  &   & \Block{1-1}{Other CNS embryonal tumors} & \Block{1-1}{0.43$\pm$0.18\\ [0.18, 0.68]} & \Block{1-1}{Atypical teratoid/rhabdoid tumor} & \Block{1-1}{0.64$\pm$0.18\\ [0.38, 0.89]}\\
\bottomrule
\end{NiceTabular}
\end{table}

\begin{table}[ht]
\small
\centering
\caption{Class-wise F1 scores when evaluating on \textbf{out-of-site} data (four sites), with a minimum number of cases set to 10 for class inclusion. Metrics are presented for the best-performing model (UNI-features and ABMIL aggregation) as mean and standard deviation with 95\% confidence intervals (CI) computed over 5 model runs.}\label{tab:f1_score_generalization_out_of_site_data_min_10}
\begin{NiceTabular}{m[c]{31mm} m[c]{16mm} m[c]{31mm} m[c]{16mm} m[c]{31mm} m[c]{16mm}}[]
\toprule
\Block{1-2}{\textbf{Tumor Category}} & & \Block{1-2}{\textbf{Tumor Family}} & & \Block{1-2}{\textbf{Tumor Type}} \\
\Block{1-1}{\textbf{Class}} & \Block{1-1}{\textbf{F1 score}\\mean$\pm$std\\ [95\% CI]} & \Block{1-1}{\textbf{Class}} & \Block{1-1}{\textbf{F1 score}\\mean$\pm$std\\ [95\% CI]} & \Block{1-1}{\textbf{Class}} & \Block{1-1}{\textbf{F1 score}\\mean$\pm$std\\ [95\% CI]} \\
\midrule
\Block{3-1}{Gliomas, glioneuronal tumors, and neuronal tumors} & \Block{3-1}{0.95$\pm$0.01\\ [0.94, 0.96]} & \Block{1-1}{Circumscribed astrocytic gliomas} & \Block{1-1}{0.86$\pm$0.02\\ [0.83, 0.89]} & \Block{1-1}{Pilocytic astrocytoma} & \Block{1-1}{0.90$\pm$0.02\\ [0.87, 0.93]}\\
\cmidrule{3-6}
  &   & \Block{1-1}{Glioneuronal and neuronal tumors} & \Block{1-1}{0.52$\pm$0.04\\ [0.46, 0.58]} & \Block{1-1}{Ganglioglioma} & \Block{1-1}{0.29$\pm$0.06\\ [0.21, 0.37]}\\
\cmidrule{3-6}
  &   & \Block{1-1}{Ependymal tumors} & \Block{1-1}{0.68$\pm$0.09\\ [0.56, 0.80]} & \Block{1-1}{/} & \Block{1-1}{/}\\
\cmidrule{1-6}
\Block{2-1}{Embryonal tumors} & \Block{2-1}{0.87$\pm$0.02\\ [0.84, 0.89]} & \Block{1-1}{Medulloblastoma} & \Block{1-1}{0.84$\pm$0.01\\ [0.83, 0.86]} & \Block{1-1}{Medulloblastoma, non-WNT/non-SHH} & \Block{1-1}{0.84$\pm$0.03\\ [0.80, 0.89]}\\
\cmidrule{3-6}
  &   & \Block{1-1}{Other CNS embryonal tumors} & \Block{1-1}{0.37$\pm$0.08\\ [0.25, 0.48]} & \Block{1-1}{Atypical teratoid/rhabdoid tumor} & \Block{1-1}{0.46$\pm$0.06\\ [0.38, 0.55]}\\
\bottomrule
\end{NiceTabular}
\end{table}

\clearpage

\subsubsection*{Model generalization detailed results a minimum of 8 cases}
\begin{table}[ht]
\small
\centering
\caption{Classification performance when evaluating on \textbf{in-site data} (two sites), with a minimum number of cases set to 8 for class inclusion. Metrics are presented as mean and standard deviation with 95\% confidence intervals (CI) computed over 5 model runs. ABMIL: attention-based multiple-instance learning, CLAM: clustering-constrained attention multiple instance-learning, MCC: Matthew's correlation coefficient. ResNet50: convolutional neural network pretrained on ImageNet, UNI and CONCH: vision transformer foundation models pretrained on histopathology data.}\label{tab:classification_performance_generalization_in_site_data_min_8}
\begin{NiceTabular}{p[c]{19mm} p[c]{19mm} p[c]{19mm} p[c]{19mm} m[c]{19mm} p[c]{19mm} p[c]{19mm}}[]
\toprule
\Block{1-1}{Classification\\granularity} & \Block{1-1}{Instance feature\\extractor} & \Block{1-1}{MIL aggregation\\method} & \Block{1-1}{MCC\\mean$\pm$std\\ [95\% CI]} & \Block{1-1}{Balanced\\accuracy\\ mean$\pm$std\\ [95\% CI]} & \Block{1-1}{F1 score\\ mean$\pm$std\\ [95\% CI]} & \Block{1-1}{AUROC\\ mean$\pm$std\\ [95\% CI]} \\

\midrule

\Block{6-1}{Tumor Category} & \Block{2-1}{ResNet50} & \Block{1-1}{ABMIL} & \Block{1-1}{0.35$\pm$0.07\\ [0.25, 0.45]} & \Block{1-1}{0.48$\pm$0.06\\ [0.40, 0.56]} & \Block{1-1}{0.38$\pm$0.08\\ [0.27, 0.48]} & \Block{1-1}{0.77$\pm$0.03\\ [0.72, 0.81]}\\
\cmidrule{3-7}
  &   & \Block{1-1}{CLAM} & \Block{1-1}{0.44$\pm$0.07\\ [0.34, 0.54]} & \Block{1-1}{0.58$\pm$0.08\\ [0.46, 0.69]} & \Block{1-1}{0.47$\pm$0.08\\ [0.36, 0.58]} & \Block{1-1}{0.80$\pm$0.06\\ [0.72, 0.88]}\\
\cmidrule{2-7}
  & \Block{2-1}{CONCH} & \Block{1-1}{ABMIL} & \Block{1-1}{0.77$\pm$0.06\\ [0.69, 0.85]} & \Block{1-1}{0.72$\pm$0.03\\ [0.68, 0.76]} & \Block{1-1}{0.72$\pm$0.05\\ [0.64, 0.79]} & \Block{1-1}{0.96$\pm$0.01\\ [0.95, 0.97]}\\
\cmidrule{3-7}
  &   & \Block{1-1}{CLAM} & \Block{1-1}{0.75$\pm$0.05\\ [0.69, 0.82]} & \Block{1-1}{0.74$\pm$0.02\\ [0.71, 0.78]} & \Block{1-1}{0.75$\pm$0.05\\ [0.68, 0.82]} & \Block{1-1}{0.95$\pm$0.01\\ [0.95, 0.96]}\\
\cmidrule{2-7}
  & \Block{2-1}{UNI} & \Block{1-1}{ABMIL} & \Block{1-1}{0.78$\pm$0.06\\ [0.69, 0.86]} & \Block{1-1}{0.70$\pm$0.06\\ [0.62, 0.78]} & \Block{1-1}{0.73$\pm$0.07\\ [0.63, 0.82]} & \Block{1-1}{0.97$\pm$0.01\\ [0.96, 0.98]}\\
\cmidrule{3-7}
  &   & \Block{1-1}{CLAM} & \Block{1-1}{0.73$\pm$0.06\\ [0.65, 0.81]} & \Block{1-1}{0.67$\pm$0.04\\ [0.62, 0.72]} & \Block{1-1}{0.67$\pm$0.04\\ [0.62, 0.72]} & \Block{1-1}{0.96$\pm$0.01\\ [0.95, 0.97]}\\
\cmidrule{1-7}
\Block[fill=gray!10]{6-1}{Tumor Family} & \Block[fill=gray!10]{2-1}{ResNet50} & \Block[fill=gray!10]{1-1}{ABMIL} & \Block[fill=gray!10]{1-1}{0.14$\pm$0.08\\ [0.03, 0.26]} & \Block[fill=gray!10]{1-1}{0.23$\pm$0.07\\ [0.14, 0.32]} & \Block[fill=gray!10]{1-1}{0.13$\pm$0.06\\ [0.05, 0.22]} & \Block[fill=gray!10]{1-1}{0.69$\pm$0.04\\ [0.62, 0.75]}\\
\cmidrule{3-7}
  &   & \Block[fill=gray!10]{1-1}{CLAM} & \Block[fill=gray!10]{1-1}{0.23$\pm$0.03\\ [0.19, 0.28]} & \Block[fill=gray!10]{1-1}{0.29$\pm$0.05\\ [0.22, 0.36]} & \Block[fill=gray!10]{1-1}{0.20$\pm$0.03\\ [0.17, 0.24]} & \Block[fill=gray!10]{1-1}{0.70$\pm$0.03\\ [0.66, 0.74]}\\
\cmidrule{2-7}
  & \Block[fill=gray!10]{2-1}{CONCH} & \Block[fill=gray!10]{1-1}{ABMIL} & \Block[fill=gray!10]{1-1}{0.56$\pm$0.07\\ [0.46, 0.65]} & \Block[fill=gray!10]{1-1}{0.53$\pm$0.06\\ [0.45, 0.61]} & \Block[fill=gray!10]{1-1}{0.52$\pm$0.06\\ [0.44, 0.60]} & \Block[fill=gray!10]{1-1}{0.88$\pm$0.03\\ [0.85, 0.92]}\\
\cmidrule{3-7}
  &   & \Block[fill=gray!10]{1-1}{CLAM} & \Block[fill=gray!10]{1-1}{0.54$\pm$0.06\\ [0.46, 0.62]} & \Block[fill=gray!10]{1-1}{0.53$\pm$0.06\\ [0.44, 0.61]} & \Block[fill=gray!10]{1-1}{0.52$\pm$0.06\\ [0.43, 0.60]} & \Block[fill=gray!10]{1-1}{0.87$\pm$0.03\\ [0.83, 0.91]}\\
\cmidrule{2-7}
  & \Block[fill=gray!10]{2-1}{UNI} & \Block[fill=gray!10]{1-1}{ABMIL} & \Block[fill=gray!10]{1-1}{0.61$\pm$0.04\\ [0.55, 0.67]} & \Block[fill=gray!10]{1-1}{0.53$\pm$0.04\\ [0.47, 0.59]} & \Block[fill=gray!10]{1-1}{0.53$\pm$0.03\\ [0.48, 0.58]} & \Block[fill=gray!10]{1-1}{0.89$\pm$0.03\\ [0.84, 0.93]}\\
\cmidrule{3-7}
  &   & \Block[fill=gray!10]{1-1}{CLAM} & \Block[fill=gray!10]{1-1}{0.62$\pm$0.07\\ [0.52, 0.72]} & \Block[fill=gray!10]{1-1}{0.52$\pm$0.07\\ [0.43, 0.61]} & \Block[fill=gray!10]{1-1}{0.52$\pm$0.06\\ [0.44, 0.60]} & \Block[fill=gray!10]{1-1}{0.89$\pm$0.02\\ [0.85, 0.92]}\\
\cmidrule{1-7}
\Block{6-1}{Tumor Type} & \Block{2-1}{ResNet50} & \Block{1-1}{ABMIL} & \Block{1-1}{0.19$\pm$0.05\\ [0.11, 0.26]} & \Block{1-1}{0.27$\pm$0.06\\ [0.19, 0.36]} & \Block{1-1}{0.17$\pm$0.03\\ [0.12, 0.21]} & \Block{1-1}{0.74$\pm$0.05\\ [0.68, 0.81]}\\
\cmidrule{3-7}
  &   & \Block{1-1}{CLAM} & \Block{1-1}{0.17$\pm$0.03\\ [0.12, 0.22]} & \Block{1-1}{0.28$\pm$0.05\\ [0.22, 0.35]} & \Block{1-1}{0.18$\pm$0.02\\ [0.15, 0.21]} & \Block{1-1}{0.75$\pm$0.05\\ [0.67, 0.82]}\\
\cmidrule{2-7}
  & \Block{2-1}{CONCH} & \Block{1-1}{ABMIL} & \Block{1-1}{0.51$\pm$0.03\\ [0.47, 0.55]} & \Block{1-1}{0.53$\pm$0.04\\ [0.47, 0.59]} & \Block{1-1}{0.49$\pm$0.03\\ [0.45, 0.54]} & \Block{1-1}{0.89$\pm$0.02\\ [0.86, 0.91]}\\
\cmidrule{3-7}
  &   & \Block{1-1}{CLAM} & \Block{1-1}{0.52$\pm$0.10\\ [0.38, 0.65]} & \Block{1-1}{0.52$\pm$0.10\\ [0.37, 0.66]} & \Block{1-1}{0.50$\pm$0.10\\ [0.36, 0.64]} & \Block{1-1}{0.89$\pm$0.03\\ [0.85, 0.92]}\\
\cmidrule{2-7}
  & \Block{2-1}{UNI} & \Block{1-1}{ABMIL} & \Block{1-1}{0.57$\pm$0.03\\ [0.52, 0.62]} & \Block{1-1}{0.51$\pm$0.03\\ [0.47, 0.55]} & \Block{1-1}{0.52$\pm$0.04\\ [0.47, 0.58]} & \Block{1-1}{0.87$\pm$0.05\\ [0.80, 0.94]}\\
\cmidrule{3-7}
  &   & \Block{1-1}{CLAM} & \Block{1-1}{0.57$\pm$0.04\\ [0.51, 0.63]} & \Block{1-1}{0.49$\pm$0.03\\ [0.45, 0.53]} & \Block{1-1}{0.49$\pm$0.04\\ [0.44, 0.54]} & \Block{1-1}{0.88$\pm$0.03\\ [0.84, 0.92]}\\
\bottomrule
\end{NiceTabular}
\end{table}

\begin{table}[ht]
\small
\centering
\caption{Classification performance when evaluating on \textbf{out-of-site data} (4 sites), with a minimum number of cases set to 8 for class inclusion. Metrics are presented as mean and standard deviation with 95\% confidence intervals (CI) computed over 5 model runs. ABMIL: attention-based multiple-instance learning, CLAM: clustering-constrained attention multiple instance-learning, MCC: Matthew's correlation coefficient. ResNet50: convolutional neural network pretrained on ImageNet, UNI and CONCH: vision transformer foundation models pretrained on histopathology data.}\label{tab:classification_performance_generalization_out_of_site_data_min_8}
\begin{NiceTabular}{p[c]{19mm} p[c]{19mm} p[c]{19mm} p[c]{19mm} m[c]{19mm} p[c]{19mm} p[c]{19mm}}[]
\toprule
\Block{1-1}{Classification\\granularity} & \Block{1-1}{Instance feature\\extractor} & \Block{1-1}{MIL aggregation\\method} & \Block{1-1}{MCC\\mean$\pm$std\\ [95\% CI]} & \Block{1-1}{Balanced\\accuracy\\ mean$\pm$std\\ [95\% CI]} & \Block{1-1}{F1 score\\ mean$\pm$std\\ [95\% CI]} & \Block{1-1}{AUROC\\ mean$\pm$std\\ [95\% CI]} \\

\midrule
\Block{6-1}{Tumor Category} & \Block{2-1}{ResNet50} & \Block{1-1}{ABMIL} & \Block{1-1}{0.37$\pm$0.08\\ [0.26, 0.48]} & \Block{1-1}{0.51$\pm$0.05\\ [0.43, 0.58]} & \Block{1-1}{0.38$\pm$0.06\\ [0.30, 0.47]} & \Block{1-1}{0.83$\pm$0.07\\ [0.73, 0.93]}\\
\cmidrule{3-7}
  &   & \Block{1-1}{CLAM} & \Block{1-1}{0.49$\pm$0.03\\ [0.44, 0.54]} & \Block{1-1}{0.64$\pm$0.05\\ [0.56, 0.71]} & \Block{1-1}{0.48$\pm$0.03\\ [0.44, 0.52]} & \Block{1-1}{0.88$\pm$0.03\\ [0.83, 0.93]}\\
\cmidrule{2-7}
  & \Block{2-1}{CONCH} & \Block{1-1}{ABMIL} & \Block{1-1}{0.78$\pm$0.03\\ [0.74, 0.82]} & \Block{1-1}{0.73$\pm$0.05\\ [0.67, 0.80]} & \Block{1-1}{0.72$\pm$0.06\\ [0.63, 0.81]} & \Block{1-1}{0.98$\pm$0.01\\ [0.97, 0.99]}\\
\cmidrule{3-7}
  &   & \Block{1-1}{CLAM} & \Block{1-1}{0.78$\pm$0.02\\ [0.74, 0.81]} & \Block{1-1}{0.75$\pm$0.03\\ [0.70, 0.79]} & \Block{1-1}{0.73$\pm$0.05\\ [0.66, 0.80]} & \Block{1-1}{0.98$\pm$0.01\\ [0.97, 0.99]}\\
\cmidrule{2-7}
  & \Block{2-1}{UNI} & \Block{1-1}{ABMIL} & \Block{1-1}{0.84$\pm$0.03\\ [0.81, 0.88]} & \Block{1-1}{0.81$\pm$0.06\\ [0.72, 0.89]} & \Block{1-1}{0.82$\pm$0.08\\ [0.72, 0.93]} & \Block{1-1}{0.99$\pm$0.00\\ [0.98, 0.99]}\\
\cmidrule{3-7}
  &   & \Block{1-1}{CLAM} & \Block{1-1}{0.82$\pm$0.03\\ [0.78, 0.86]} & \Block{1-1}{0.75$\pm$0.06\\ [0.68, 0.83]} & \Block{1-1}{0.78$\pm$0.07\\ [0.68, 0.88]} & \Block{1-1}{0.98$\pm$0.01\\ [0.98, 0.99]}\\
\cmidrule{1-7}
\Block[fill=gray!10]{6-1}{Tumor Family} & \Block[fill=gray!10]{2-1}{ResNet50} & \Block[fill=gray!10]{1-1}{ABMIL} & \Block[fill=gray!10]{1-1}{0.14$\pm$0.06\\ [0.05, 0.23]} & \Block[fill=gray!10]{1-1}{0.29$\pm$0.04\\ [0.23, 0.35]} & \Block[fill=gray!10]{1-1}{0.13$\pm$0.04\\ [0.07, 0.19]} & \Block[fill=gray!10]{1-1}{0.73$\pm$0.01\\ [0.72, 0.74]}\\
\cmidrule{3-7}
  &   & \Block[fill=gray!10]{1-1}{CLAM} & \Block[fill=gray!10]{1-1}{0.21$\pm$0.04\\ [0.15, 0.26]} & \Block[fill=gray!10]{1-1}{0.29$\pm$0.03\\ [0.25, 0.34]} & \Block[fill=gray!10]{1-1}{0.17$\pm$0.02\\ [0.15, 0.20]} & \Block[fill=gray!10]{1-1}{0.75$\pm$0.01\\ [0.73, 0.76]}\\
\cmidrule{2-7}
  & \Block[fill=gray!10]{2-1}{CONCH} & \Block[fill=gray!10]{1-1}{ABMIL} & \Block[fill=gray!10]{1-1}{0.53$\pm$0.02\\ [0.50, 0.56]} & \Block[fill=gray!10]{1-1}{0.51$\pm$0.04\\ [0.44, 0.57]} & \Block[fill=gray!10]{1-1}{0.45$\pm$0.03\\ [0.40, 0.49]} & \Block[fill=gray!10]{1-1}{0.90$\pm$0.00\\ [0.89, 0.90]}\\
\cmidrule{3-7}
  &   & \Block[fill=gray!10]{1-1}{CLAM} & \Block[fill=gray!10]{1-1}{0.50$\pm$0.02\\ [0.47, 0.53]} & \Block[fill=gray!10]{1-1}{0.49$\pm$0.06\\ [0.40, 0.57]} & \Block[fill=gray!10]{1-1}{0.44$\pm$0.04\\ [0.39, 0.49]} & \Block[fill=gray!10]{1-1}{0.88$\pm$0.02\\ [0.86, 0.90]}\\
\cmidrule{2-7}
  & \Block[fill=gray!10]{2-1}{UNI} & \Block[fill=gray!10]{1-1}{ABMIL} & \Block[fill=gray!10]{1-1}{0.59$\pm$0.03\\ [0.56, 0.63]} & \Block[fill=gray!10]{1-1}{0.59$\pm$0.02\\ [0.56, 0.62]} & \Block[fill=gray!10]{1-1}{0.57$\pm$0.02\\ [0.54, 0.60]} & \Block[fill=gray!10]{1-1}{0.90$\pm$0.01\\ [0.89, 0.90]}\\
\cmidrule{3-7}
  &   & \Block[fill=gray!10]{1-1}{CLAM} & \Block[fill=gray!10]{1-1}{0.59$\pm$0.03\\ [0.55, 0.64]} & \Block[fill=gray!10]{1-1}{0.56$\pm$0.06\\ [0.48, 0.64]} & \Block[fill=gray!10]{1-1}{0.56$\pm$0.06\\ [0.47, 0.64]} & \Block[fill=gray!10]{1-1}{0.89$\pm$0.01\\ [0.87, 0.91]}\\
\cmidrule{1-7}
\Block{6-1}{Tumor Type} & \Block{2-1}{ResNet50} & \Block{1-1}{ABMIL} & \Block{1-1}{0.11$\pm$0.03\\ [0.06, 0.15]} & \Block{1-1}{0.21$\pm$0.03\\ [0.16, 0.25]} & \Block{1-1}{0.10$\pm$0.03\\ [0.05, 0.14]} & \Block{1-1}{0.68$\pm$0.05\\ [0.61, 0.74]}\\
\cmidrule{3-7}
  &   & \Block{1-1}{CLAM} & \Block{1-1}{0.14$\pm$0.05\\ [0.07, 0.21]} & \Block{1-1}{0.23$\pm$0.05\\ [0.16, 0.30]} & \Block{1-1}{0.14$\pm$0.05\\ [0.06, 0.21]} & \Block{1-1}{0.66$\pm$0.03\\ [0.62, 0.71]}\\
\cmidrule{2-7}
  & \Block{2-1}{CONCH} & \Block{1-1}{ABMIL} & \Block{1-1}{0.46$\pm$0.03\\ [0.42, 0.50]} & \Block{1-1}{0.44$\pm$0.02\\ [0.41, 0.47]} & \Block{1-1}{0.41$\pm$0.02\\ [0.38, 0.44]} & \Block{1-1}{0.86$\pm$0.03\\ [0.82, 0.90]}\\
\cmidrule{3-7}
  &   & \Block{1-1}{CLAM} & \Block{1-1}{0.47$\pm$0.02\\ [0.44, 0.51]} & \Block{1-1}{0.46$\pm$0.05\\ [0.40, 0.53]} & \Block{1-1}{0.40$\pm$0.03\\ [0.36, 0.45]} & \Block{1-1}{0.85$\pm$0.02\\ [0.82, 0.88]}\\
\cmidrule{2-7}
  & \Block{2-1}{UNI} & \Block{1-1}{ABMIL} & \Block{1-1}{0.51$\pm$0.05\\ [0.44, 0.58]} & \Block{1-1}{0.43$\pm$0.03\\ [0.39, 0.47]} & \Block{1-1}{0.40$\pm$0.03\\ [0.35, 0.44]} & \Block{1-1}{0.86$\pm$0.02\\ [0.82, 0.89]}\\
\cmidrule{3-7}
  &   & \Block{1-1}{CLAM} & \Block{1-1}{0.57$\pm$0.04\\ [0.52, 0.62]} & \Block{1-1}{0.45$\pm$0.05\\ [0.39, 0.51]} & \Block{1-1}{0.42$\pm$0.04\\ [0.37, 0.48]} & \Block{1-1}{0.86$\pm$0.04\\ [0.81, 0.92]}\\
\bottomrule
\end{NiceTabular}
\end{table}

\begin{table}[ht]
 \centering
 \caption{Classification performance when training on two sites and testing on the remaining four sites, with a minimum of 8 cases per tumor entity. Metrics are presented as mean and standard deviation with 95\% confidence intervals (CI) computed over 5 model runs. The difference between testing on the in-site and out-of-site data is also presented, with arrows showing if performance increased ($\uparrow$) or decreased ($\downarrow$). Abbreviations: ABMIL, attention-based multiple-instance learning; CLAM, clustering-constrained attention multiple instance-learning, MCC, Matthew's correlation coefficient; ResNet50, convolutional neural network pretrained on ImageNet; UNI and CONCH, vision transformer foundation models pretrained on histopathology data.}\label{tab:generalization_summary_min_8}

    \small
    \begin{NiceTabular}{m[c]{16mm} m[c]{16mm} m[c]{16mm} m[c]{16mm} m[c]{16mm} m[c]{16mm} m[c]{16mm} m[c]{16mm}}[]

    \toprule

    \Block{2-1}{Classification\\granularity} & \Block{2-1}{Instance\\feature\\extractor} & \Block{2-1}{Aggregation\\method} & \Block{1-2}{In-site testing} & & \Block{1-2}{Out-of-site testing} & & \Block{2-1}{Drop in\\performance\\(MCC difference)} \\

    \cmidrule{4-7}
    
    & & & \Block{1-1}{\textbf{MCC}\\mean$\pm$std\\ [95\% CI]} & \Block{1-1}{\textbf{Balanced}\\\textbf{accuracy}\\mean$\pm$std\\ [95\% CI]} & \Block{1-1}{\textbf{MCC}\\mean$\pm$std\\ [95\% CI]} & \Block{1-1}{\textbf{Balanced}\\\textbf{accuracy}\\mean$\pm$std\\ [95\% CI]} & \\

    \midrule

    \Block{6-1}{Tumor\\Category} & \Block{2-1}{ResNet50} & \Block{1-1}{ABMIL} & \Block{1-1}{0.35$\pm$0.07\\ [0.25, 0.45]} & \Block{1-1}{0.48$\pm$0.06\\ [0.40, 0.56]} & \Block{1-1}{0.37$\pm$0.08\\ [0.26, 0.48]} & \Block{1-1}{0.51$\pm$0.05\\ [0.43, 0.58]} & \Block{1-1}{0.02$\uparrow$} \\

    \cmidrule{3-8}

    & & \Block{1-1}{CLAM} & \Block{1-1}{0.44$\pm$0.07\\ [0.34, 0.54]} & \Block{1-1}{0.58$\pm$0.08\\ [0.46, 0.69]} & \Block{1-1}{0.49$\pm$0.03\\ [0.44, 0.54]} & \Block{1-1}{0.64$\pm$0.05\\ [0.56, 0.71]} & \Block{1-1}{0.05$\uparrow$} \\

    \cmidrule{2-8}
    
    & \Block{2-1}{CONCH} & \Block{1-1}{ABMIL} & \Block{1-1}{0.77$\pm$0.06\\ [0.69, 0.85]} & \Block{1-1}{0.72$\pm$0.03\\ [0.68, 0.76]} & \Block{1-1}{0.78$\pm$0.03\\ [0.74, 0.82]} & \Block{1-1}{0.73$\pm$0.05\\ [0.67, 0.80]} & \Block{1-1}{0.01$\uparrow$} \\

    \cmidrule{3-8}
    
    & & \Block{1-1}{CLAM} & \Block{1-1}{0.75$\pm$0.05\\ [0.69, 0.82]} & \Block{1-1}{0.74$\pm$0.02\\ [0.71, 0.78]} & \Block{1-1}{0.78$\pm$0.02\\ [0.74, 0.81]} & \Block{1-1}{0.75$\pm$0.03\\ [0.70, 0.79]} & \Block{1-1}{0.03$\uparrow$} \\

    \cmidrule{2-8}
    
    & \Block{2-1}{UNI} & \Block{1-1}{ABMIL} & \Block{1-1}{0.78$\pm$0.06\\ [0.69, 0.86]} & \Block{1-1}{0.70$\pm$0.06\\ [0.62, 0.78]} & \Block{1-1}{0.84$\pm$0.03\\ [0.81, 0.88]} & \Block{1-1}{0.81$\pm$0.06\\ [0.72, 0.89]} & \Block{1-1}{0.06$\uparrow$} \\
    \cmidrule{3-8}
    
    & & \Block{1-1}{CLAM} & \Block{1-1}{0.73$\pm$0.06\\ [0.65, 0.81]} & \Block{1-1}{0.67$\pm$0.04\\ [0.62, 0.72]} & \Block{1-1}{0.82$\pm$0.03\\ [0.78, 0.86]} & \Block{1-1}{0.75$\pm$0.06\\ [0.68, 0.83]} & \Block{1-1}{0.09$\uparrow$} \\

    \midrule\midrule

    \Block[fill=gray!10]{6-1}{Tumor\\family} & \Block[fill=gray!10]{2-1}{ResNet50} & \Block[fill=gray!10]{1-1}{ABMIL} & \Block[fill=gray!10]{1-1}{0.14$\pm$0.08\\ [0.03, 0.26]} & \Block[fill=gray!10]{1-1}{0.23$\pm$0.07\\ [0.14, 0.32]} & \Block[fill=gray!10]{1-1}{0.14$\pm$0.06\\ [0.05, 0.23]} & \Block[fill=gray!10]{1-1}{0.29$\pm$0.04\\ [0.23, 0.35]} & \Block[fill=gray!10]{1-1}{0.00} \\

    \cmidrule{3-8}

    & & \Block[fill=gray!10]{1-1}{CLAM} & \Block[fill=gray!10]{1-1}{0.23$\pm$0.03\\ [0.19, 0.28]} & \Block[fill=gray!10]{1-1}{0.29$\pm$0.05\\ [0.22, 0.36]} & \Block[fill=gray!10]{1-1}{0.21$\pm$0.04\\ [0.15, 0.26]} & \Block[fill=gray!10]{1-1}{0.29$\pm$0.03\\ [0.25, 0.34]} & \Block[fill=gray!10]{1-1}{-0.02$\downarrow$} \\

    \cmidrule{2-8}
    
    & \Block[fill=gray!10]{2-1}{CONCH} & \Block[fill=gray!10]{1-1}{ABMIL} & \Block[fill=gray!10]{1-1}{0.56$\pm$0.07\\ [0.46, 0.65]} & \Block[fill=gray!10]{1-1}{0.53$\pm$0.06\\ [0.45, 0.61]} & \Block[fill=gray!10]{1-1}{0.53$\pm$0.02\\ [0.50, 0.56]} & \Block[fill=gray!10]{1-1}{0.51$\pm$0.04\\ [0.44, 0.57]} & \Block[fill=gray!10]{1-1}{-0.03$\downarrow$} \\

    \cmidrule{3-8}
    
    & & \Block[fill=gray!10]{1-1}{CLAM} & \Block[fill=gray!10]{1-1}{0.54$\pm$0.06\\ [0.46, 0.62]} & \Block[fill=gray!10]{1-1}{0.53$\pm$0.06\\ [0.44, 0.61]} & \Block[fill=gray!10]{1-1}{0.50$\pm$0.02\\ [0.47, 0.53]} & \Block[fill=gray!10]{1-1}{0.49$\pm$0.06\\ [0.40, 0.57]} & \Block[fill=gray!10]{1-1}{-0.04$\downarrow$} \\

    \cmidrule{2-8}
    
    & \Block[fill=gray!10]{2-1}{UNI} & \Block[fill=gray!10]{1-1}{ABMIL} & \Block[fill=gray!10]{1-1}{0.61$\pm$0.04\\ [0.55, 0.67]} & \Block[fill=gray!10]{1-1}{0.53$\pm$0.04\\ [0.47, 0.59]} & \Block[fill=gray!10]{1-1}{0.59$\pm$0.03\\ [0.56, 0.63]} & \Block[fill=gray!10]{1-1}{0.59$\pm$0.02\\ [0.56, 0.62]} & \Block[fill=gray!10]{1-1}{-0.02$\downarrow$} \\

    \cmidrule{3-8}
    
    & & \Block[fill=gray!10]{1-1}{CLAM} & \Block[fill=gray!10]{1-1}{0.62$\pm$0.07\\ [0.52, 0.72]} & \Block[fill=gray!10]{1-1}{0.52$\pm$0.07\\ [0.43, 0.61]} & \Block[fill=gray!10]{1-1}{0.59$\pm$0.03\\ [0.55, 0.64]} & \Block[fill=gray!10]{1-1}{0.56$\pm$0.06\\ [0.48, 0.64]} & \Block[fill=gray!10]{1-1}{-0.03$\downarrow$} \\

    \midrule\midrule

       \Block{6-1}{Tumor\\type} & \Block{2-1}{ResNet50} & \Block{1-1}{ABMIL} & \Block{1-1}{0.19$\pm$0.05\\ [0.11, 0.26]} & \Block{1-1}{0.27$\pm$0.06\\ [0.19, 0.36]} & \Block{1-1}{0.11$\pm$0.03\\ [0.06, 0.15]} & \Block{1-1}{0.21$\pm$0.03\\ [0.16, 0.25]} & \Block{1-1}{-0.08$\downarrow$} \\

    \cmidrule{3-8}

    & & \Block{1-1}{CLAM} & \Block{1-1}{0.17$\pm$0.03\\ [0.12, 0.22]} & \Block{1-1}{0.28$\pm$0.05\\ [0.22, 0.35]} & \Block{1-1}{0.14$\pm$0.05\\ [0.07, 0.21]} & \Block{1-1}{0.23$\pm$0.05\\ [0.16, 0.30]} & \Block{1-1}{-0.03$\downarrow$} \\

    \cmidrule{2-8}
    
    & \Block{2-1}{CONCH} & \Block{1-1}{ABMIL} & \Block{1-1}{0.51$\pm$0.03\\ [0.47, 0.55]} & \Block{1-1}{0.53$\pm$0.04\\ [0.47, 0.59]} & \Block{1-1}{0.46$\pm$0.03\\ [0.42, 0.50]} & \Block{1-1}{0.44$\pm$0.02\\ [0.41, 0.47]} & \Block{1-1}{-0.05$\downarrow$} \\

    \cmidrule{3-8}
    
    & & \Block{1-1}{CLAM} & \Block{1-1}{0.52$\pm$0.10\\ [0.38, 0.65]} & \Block{1-1}{0.52$\pm$0.10\\ [0.37, 0.66]} & \Block{1-1}{0.47$\pm$0.02\\ [0.44, 0.51]} & \Block{1-1}{0.46$\pm$0.05\\ [0.40, 0.53]} & \Block{1-1}{-0.05$\downarrow$} \\

    \cmidrule{2-8}
    
    & \Block{2-1}{UNI} & \Block{1-1}{ABMIL} & \Block{1-1}{0.57$\pm$0.03\\ [0.52, 0.62]} & \Block{1-1}{0.51$\pm$0.03\\ [0.47, 0.55]} & \Block{1-1}{0.51$\pm$0.05\\ [0.44, 0.58]} & \Block{1-1}{0.43$\pm$0.03\\ [0.39, 0.47]} & \Block{1-1}{-0.06$\downarrow$} \\

    \cmidrule{3-8}
    
    & & \Block{1-1}{CLAM} & \Block{1-1}{0.57$\pm$0.04\\ [0.51, 0.63]} & \Block{1-1}{0.49$\pm$0.03\\ [0.45, 0.53]} & \Block{1-1}{0.57$\pm$0.04\\ [0.52, 0.62]} & \Block{1-1}{0.45$\pm$0.05\\ [0.39, 0.51]} & \Block{1-1}{0.00} \\

    \bottomrule
    \end{NiceTabular}
\end{table}
\begin{table}[ht]
\small
\centering
\caption{Class-wise F1 scores when evaluating on \textbf{in-site data} (two sites), with a minimum number of cases set to 8 for class inclusion. Metrics are presented for the best-performing model (UNI-features and ABMIL aggregation) as mean and standard deviation with 95\% confidence intervals (CI) computed over 5 model runs. }\label{tab:f1_score_generalization_in_site_data_min_8}
\begin{NiceTabular}{m[c]{31mm} m[c]{16mm} m[c]{31mm} m[c]{16mm} m[c]{31mm} m[c]{16mm}}[]
\toprule
\Block{1-2}{\textbf{Tumor Category}} & & \Block{1-2}{\textbf{Tumor Family}} & & \Block{1-2}{\textbf{Tumor Type}} \\
\Block{1-1}{\textbf{Class}} & \Block{1-1}{\textbf{F1 score}\\mean$\pm$std\\ [95\% CI]} & \Block{1-1}{\textbf{Class}} & \Block{1-1}{\textbf{F1 score}\\mean$\pm$std\\ [95\% CI]} & \Block{1-1}{\textbf{Class}} & \Block{1-1}{\textbf{F1 score}\\mean$\pm$std\\ [95\% CI]} \\
\midrule
\Block{6-1}{Gliomas, glioneuronal tumors, and neuronal tumors} & \Block{6-1}{0.93$\pm$0.02\\ [0.89, 0.96]} & \Block{1-1}{Circumscribed astrocytic gliomas} & \Block{1-1}{0.75$\pm$0.04\\ [0.70, 0.80]} & \Block{1-1}{Pilocytic astrocytoma} & \Block{1-1}{0.77$\pm$0.06\\ [0.69, 0.84]}\\
\cmidrule{3-6}
  &   & \Block{2-1}{Glioneuronal and neuronal tumors} & \Block{2-1}{0.64$\pm$0.08\\ [0.53, 0.76]} & \Block{1-1}{Ganglioglioma} & \Block{1-1}{0.48$\pm$0.11\\ [0.32, 0.63]}\\
\cmidrule{5-6}
  &   &   &   & \Block{1-1}{Dysembryoplastic neuroepithelial tumor} & \Block{1-1}{0.46$\pm$0.16\\ [0.24, 0.68]}\\
\cmidrule{3-6}
  &   & \Block{1-1}{Ependymal tumors} & \Block{1-1}{0.78$\pm$0.12\\ [0.62, 0.95]} & \Block{1-1}{Ependymoma grade 3} & \Block{1-1}{0.79$\pm$0.09\\ [0.67, 0.91]}\\
\cmidrule{3-6}
  &   & \Block{1-1}{Adult-type diffuse gliomas} & \Block{1-1}{0.00$\pm$0.00\\ [nan, nan]} & \Block{1-1}{Glioblastoma} & \Block{1-1}{0.00$\pm$0.00\\ [nan, nan]}\\
\cmidrule{3-6}
  &   & \Block{1-1}{Pediatric-type diffuse high-grade gliomas} & \Block{1-1}{0.00$\pm$0.00\\ [nan, nan]} & \Block{1-1}{/} & \Block{1-1}{/}\\
\cmidrule{1-6}
\Block{2-1}{Embryonal tumors} & \Block{2-1}{0.84$\pm$0.05\\ [0.77, 0.91]} & \Block{1-1}{Medulloblastoma} & \Block{1-1}{0.85$\pm$0.05\\ [0.79, 0.92]} & \Block{1-1}{Medulloblastoma, non-WNT/non-SHH} & \Block{1-1}{0.90$\pm$0.07\\ [0.81, 1.00]}\\
\cmidrule{3-6}
  &   & \Block{1-1}{Other CNS embryonal tumors} & \Block{1-1}{0.50$\pm$0.15\\ [0.30, 0.71]} & \Block{1-1}{Atypical teratoid/rhabdoid tumor} & \Block{1-1}{0.55$\pm$0.30\\ [0.13, 0.97]}\\
\cmidrule{1-6}
\Block{1-1}{Meningiomas} & \Block{1-1}{0.23$\pm$0.29\\ [-0.17, 0.64]} & \Block{1-1}{Meningioma} & \Block{1-1}{0.71$\pm$0.12\\ [0.55, 0.88]} & \Block{1-1}{/} & \Block{1-1}{/}\\
\bottomrule
\end{NiceTabular}
\end{table}

\begin{table}[ht]
\small
\centering
\caption{Class-wise F1 scores when evaluating on \textbf{out-of-site} data (four sites), with a minimum number of cases set to 8 for class inclusion. Metrics are presented for the best-performing model (UNI-features and ABMIL aggregation) as mean and standard deviation with 95\% confidence intervals (CI) computed over 5 model runs.}\label{tab:f1_score_generalization_out_of_site_data_min_8}
\begin{NiceTabular}{m[c]{31mm} m[c]{16mm} m[c]{31mm} m[c]{16mm} m[c]{31mm} m[c]{16mm}}[]
\toprule
\Block{1-2}{\textbf{Tumor Category}} & & \Block{1-2}{\textbf{Tumor Family}} & & \Block{1-2}{\textbf{Tumor Type}} \\
\Block{1-1}{\textbf{Class}} & \Block{1-1}{\textbf{F1 score}\\mean$\pm$std\\ [95\% CI]} & \Block{1-1}{\textbf{Class}} & \Block{1-1}{\textbf{F1 score}\\mean$\pm$std\\ [95\% CI]} & \Block{1-1}{\textbf{Class}} & \Block{1-1}{\textbf{F1 score}\\mean$\pm$std\\ [95\% CI]} \\
\midrule
\Block{6-1}{Gliomas, glioneuronal tumors, and neuronal tumors} & \Block{6-1}{0.95$\pm$0.01\\ [0.93, 0.97]} & \Block{1-1}{Circumscribed astrocytic gliomas} & \Block{1-1}{0.85$\pm$0.01\\ [0.83, 0.87]} & \Block{1-1}{Pilocytic astrocytoma} & \Block{1-1}{0.83$\pm$0.04\\ [0.77, 0.89]}\\
\cmidrule{3-6}
  &   & \Block{2-1}{Glioneuronal and neuronal tumors} & \Block{2-1}{0.49$\pm$0.03\\ [0.45, 0.53]} & \Block{1-1}{Ganglioglioma} & \Block{1-1}{0.25$\pm$0.06\\ [0.17, 0.34]}\\
\cmidrule{5-6}
  &   &   &   & \Block{1-1}{Dysembryoplastic neuroepithelial tumor} & \Block{1-1}{0.16$\pm$0.14\\ [-0.04, 0.36]}\\
\cmidrule{3-6}
  &   & \Block{1-1}{Ependymal tumors} & \Block{1-1}{0.70$\pm$0.06\\ [0.62, 0.78]} & \Block{1-1}{Ependymoma grade 3} & \Block{1-1}{0.65$\pm$0.07\\ [0.55, 0.75]}\\
\cmidrule{3-6}
  &   & \Block{1-1}{Adult-type diffuse gliomas} & \Block{1-1}{0.08$\pm$0.07\\ [-0.02, 0.19]} & \Block{1-1}{Glioblastoma} & \Block{1-1}{0.06$\pm$0.08\\ [-0.05, 0.17]}\\
\cmidrule{3-6}
  &   & \Block{1-1}{Pediatric-type diffuse high-grade gliomas} & \Block{1-1}{0.39$\pm$0.25\\ [0.04, 0.74]} & \Block{1-1}{/} & \Block{1-1}{/}\\
\cmidrule{1-6}
\Block{2-1}{Embryonal tumors} & \Block{2-1}{0.87$\pm$0.03\\ [0.84, 0.91]} & \Block{1-1}{Medulloblastoma} & \Block{1-1}{0.78$\pm$0.07\\ [0.67, 0.88]} & \Block{1-1}{Medulloblastoma, non-WNT/non-SHH} & \Block{1-1}{0.77$\pm$0.01\\ [0.76, 0.79]}\\
\cmidrule{3-6}
  &   & \Block{1-1}{Other CNS embryonal tumors} & \Block{1-1}{0.33$\pm$0.09\\ [0.21, 0.45]} & \Block{1-1}{Atypical teratoid/rhabdoid tumor} & \Block{1-1}{0.39$\pm$0.03\\ [0.36, 0.43]}\\
\cmidrule{1-6}
\Block{1-1}{Meningiomas} & \Block{1-1}{0.53$\pm$0.28\\ [0.14, 0.92]} & \Block{1-1}{Meningioma} & \Block{1-1}{0.93$\pm$0.13\\ [0.75, 1.12]} & \Block{1-1}{/} & \Block{1-1}{/}\\
\bottomrule
\end{NiceTabular}
\end{table}

\clearpage

\subsubsection*{Classification results using balanced classes - 10 cases for each tumor entity}
\begin{table}[ht]
\small
\centering
\caption{Classification performance when training on data from all contributing sites and using balanced classes (10 cases for each tumor entity). Metrics are presented as mean and standard deviation with 95\% confidence intervals (CI) computed over 150 model runs. Abbreviations: ABMIL, attention-based multiple-instance learning; CLAM, clustering-constrained attention multiple instance-learning, MCC, Matthew's correlation coefficient; ResNet50, convolutional neural network pretrained on ImageNet; UNI and CONCH, vision transformer foundation models pretrained on histopathology data.}\label{tab:classification_results_balanced_classes}
\begin{NiceTabular}{p[c]{19mm} p[c]{19mm} p[c]{19mm} p[c]{19mm} m[c]{19mm} p[c]{19mm} p[c]{19mm}}[]
\toprule
\Block{1-1}{Classification\\granularity} & \Block{1-1}{Instance feature\\extractor} & \Block{1-1}{MIL aggregation\\method} & \Block{1-1}{MCC\\mean$\pm$std\\ [95\% CI]} & \Block{1-1}{Balanced\\accuracy\\ mean$\pm$std\\ [95\% CI]} & \Block{1-1}{F1 score\\ mean$\pm$std\\ [95\% CI]} & \Block{1-1}{AUROC\\ mean$\pm$std\\ [95\% CI]} \\
\midrule
\Block{4-1}{Tumor Category} & \Block{2-1}{CONCH} & \Block{1-1}{ABMIL} & \Block{1-1}{0.60$\pm$0.10\\ [0.58, 0.61]} & \Block{1-1}{0.62$\pm$0.09\\ [0.61, 0.64]} & \Block{1-1}{0.60$\pm$0.09\\ [0.58, 0.62]} & \Block{1-1}{0.90$\pm$0.04\\ [0.90, 0.91]}\\
\cmidrule{3-7}
  &   & \Block{1-1}{CLAM} & \Block{1-1}{0.58$\pm$0.10\\ [0.56, 0.60]} & \Block{1-1}{0.61$\pm$0.09\\ [0.60, 0.63]} & \Block{1-1}{0.59$\pm$0.10\\ [0.57, 0.60]} & \Block{1-1}{0.90$\pm$0.05\\ [0.89, 0.90]}\\
\cmidrule{2-7}
  & \Block{2-1}{UNI} & \Block{1-1}{ABMIL} & \Block{1-1}{0.51$\pm$0.11\\ [0.49, 0.53]} & \Block{1-1}{0.56$\pm$0.10\\ [0.54, 0.57]} & \Block{1-1}{0.54$\pm$0.10\\ [0.52, 0.55]} & \Block{1-1}{0.88$\pm$0.05\\ [0.87, 0.89]}\\
\cmidrule{3-7}
  &   & \Block{1-1}{CLAM} & \Block{1-1}{0.46$\pm$0.12\\ [0.44, 0.48]} & \Block{1-1}{0.51$\pm$0.10\\ [0.49, 0.53]} & \Block{1-1}{0.49$\pm$0.10\\ [0.48, 0.51]} & \Block{1-1}{0.85$\pm$0.05\\ [0.84, 0.86]}\\
\cmidrule{1-7}
\Block[fill=gray!10]{4-1}{Tumor Family} & \Block[fill=gray!10]{2-1}{CONCH} & \Block[fill=gray!10]{1-1}{ABMIL} & \Block[fill=gray!10]{1-1}{0.44$\pm$0.09\\ [0.43, 0.45]} & \Block[fill=gray!10]{1-1}{0.48$\pm$0.08\\ [0.47, 0.50]} & \Block[fill=gray!10]{1-1}{0.45$\pm$0.08\\ [0.44, 0.47]} & \Block[fill=gray!10]{1-1}{0.86$\pm$0.03\\ [0.86, 0.87]}\\
\cmidrule{3-7}
  &   & \Block[fill=gray!10]{1-1}{CLAM} & \Block[fill=gray!10]{1-1}{0.45$\pm$0.08\\ [0.44, 0.46]} & \Block[fill=gray!10]{1-1}{0.49$\pm$0.07\\ [0.48, 0.50]} & \Block[fill=gray!10]{1-1}{0.46$\pm$0.07\\ [0.45, 0.48]} & \Block[fill=gray!10]{1-1}{0.86$\pm$0.03\\ [0.85, 0.86]}\\
\cmidrule{2-7}
  & \Block[fill=gray!10]{2-1}{UNI} & \Block[fill=gray!10]{1-1}{ABMIL} & \Block[fill=gray!10]{1-1}{0.43$\pm$0.08\\ [0.41, 0.44]} & \Block[fill=gray!10]{1-1}{0.47$\pm$0.07\\ [0.46, 0.48]} & \Block[fill=gray!10]{1-1}{0.45$\pm$0.07\\ [0.44, 0.46]} & \Block[fill=gray!10]{1-1}{0.84$\pm$0.04\\ [0.83, 0.85]}\\
\cmidrule{3-7}
  &   & \Block[fill=gray!10]{1-1}{CLAM} & \Block[fill=gray!10]{1-1}{0.40$\pm$0.09\\ [0.38, 0.41]} & \Block[fill=gray!10]{1-1}{0.45$\pm$0.08\\ [0.44, 0.46]} & \Block[fill=gray!10]{1-1}{0.43$\pm$0.08\\ [0.42, 0.44]} & \Block[fill=gray!10]{1-1}{0.82$\pm$0.04\\ [0.82, 0.83]}\\
\cmidrule{1-7}
\Block{4-1}{Tumor Type} & \Block{2-1}{CONCH} & \Block{1-1}{ABMIL} & \Block{1-1}{0.49$\pm$0.08\\ [0.48, 0.50]} & \Block{1-1}{0.53$\pm$0.07\\ [0.52, 0.54]} & \Block{1-1}{0.49$\pm$0.08\\ [0.48, 0.51]} & \Block{1-1}{0.90$\pm$0.02\\ [0.90, 0.91]}\\
\cmidrule{3-7}
  &   & \Block{1-1}{CLAM} & \Block{1-1}{0.49$\pm$0.09\\ [0.47, 0.50]} & \Block{1-1}{0.53$\pm$0.08\\ [0.51, 0.54]} & \Block{1-1}{0.50$\pm$0.09\\ [0.48, 0.51]} & \Block{1-1}{0.90$\pm$0.03\\ [0.89, 0.90]}\\
\cmidrule{2-7}
  & \Block{2-1}{UNI} & \Block{1-1}{ABMIL} & \Block{1-1}{0.42$\pm$0.11\\ [0.40, 0.44]} & \Block{1-1}{0.48$\pm$0.10\\ [0.46, 0.49]} & \Block{1-1}{0.45$\pm$0.10\\ [0.43, 0.46]} & \Block{1-1}{0.87$\pm$0.04\\ [0.86, 0.87]}\\
\cmidrule{3-7}
  &   & \Block{1-1}{CLAM} & \Block{1-1}{0.38$\pm$0.10\\ [0.36, 0.39]} & \Block{1-1}{0.44$\pm$0.09\\ [0.43, 0.45]} & \Block{1-1}{0.41$\pm$0.09\\ [0.40, 0.43]} & \Block{1-1}{0.84$\pm$0.04\\ [0.83, 0.85]}\\
\bottomrule
\end{NiceTabular}
\end{table}

\begin{table}[ht]
\small
\centering
\caption{Class-wise F1 scores obtained from the best model (CONCH instance-level features and CLAM aggregation) for all the classification tasks when using balanced classes (10 cases for each of the tumor entities). F1 scores are presented as mean and standard deviation with 95\% confidence intervals (CI) computed over 150 model runs.}\label{tab:f1_score_balanced_classes}
\begin{NiceTabular}{m[c]{31mm} m[c]{16mm} m[c]{31mm} m[c]{16mm} m[c]{31mm} m[c]{16mm}}[]
\toprule
\Block{1-2}{\textbf{Tumor Category}} & & \Block{1-2}{\textbf{Tumor Family}} & & \Block{1-2}{\textbf{Tumor Type}} \\
\Block{1-1}{\textbf{Class}} & \Block{1-1}{\textbf{F1 score}\\mean$\pm$std\\ [95\% CI]} & \Block{1-1}{\textbf{Class}} & \Block{1-1}{\textbf{F1 score}\\mean$\pm$std\\ [95\% CI]} & \Block{1-1}{\textbf{Class}} & \Block{1-1}{\textbf{F1 score}\\mean$\pm$std\\ [95\% CI]} \\
\midrule
\Block{7-1}{Gliomas, glioneuronal tumors, and neuronal tumors} & \Block{7-1}{0.68$\pm$0.21\\ [0.65, 0.72]} & \Block{1-1}{Circumscribed astrocytic gliomas} & \Block{1-1}{0.44$\pm$0.22\\ [0.41, 0.48]} & \Block{1-1}{Pilocytic astrocytoma} & \Block{1-1}{0.47$\pm$0.27\\ [0.43, 0.52]}\\
\cmidrule{3-6}
  &   & \Block{2-1}{Glioneuronal and neuronal tumors} & \Block{2-1}{0.38$\pm$0.26\\ [0.34, 0.42]} & \Block{1-1}{Ganglioglioma} & \Block{1-1}{0.41$\pm$0.23\\ [0.37, 0.45]}\\
\cmidrule{5-6}
  &   &   &   & \Block{1-1}{Dysembryoplastic neuroepithelial tumor} & \Block{1-1}{0.54$\pm$0.25\\ [0.50, 0.59]}\\
\cmidrule{3-6}
  &   & \Block{2-1}{Ependymal tumors} & \Block{2-1}{0.69$\pm$0.20\\ [0.65, 0.72]} & \Block{1-1}{Ependymoma grade 1-2} & \Block{1-1}{0.47$\pm$0.23\\ [0.43, 0.50]}\\
\cmidrule{5-6}
  &   &   &   & \Block{1-1}{Ependymoma grade 3} & \Block{1-1}{0.32$\pm$0.23\\ [0.28, 0.36]}\\
\cmidrule{3-6}
  &   & \Block{1-1}{Adult-type diffuse gliomas} & \Block{1-1}{0.12$\pm$0.18\\ [0.09, 0.15]} & \Block{1-1}{Glioblastoma} & \Block{1-1}{0.23$\pm$0.21\\ [0.20, 0.27]}\\
\cmidrule{3-6}
  &   & \Block{1-1}{Pediatric-type diffuse high-grade gliomas} & \Block{1-1}{0.33$\pm$0.23\\ [0.29, 0.36]} & \Block{1-1}{/} & \Block{1-1}{/}\\
\cmidrule{1-6}
\Block{4-1}{Embryonal tumors} & \Block{4-1}{0.51$\pm$0.22\\ [0.48, 0.55]} & \Block{2-1}{Medulloblastoma} & \Block{2-1}{0.55$\pm$0.20\\ [0.51, 0.58]} & \Block{1-1}{Medulloblastoma, non-WNT/non-SHH} & \Block{1-1}{0.62$\pm$0.19\\ [0.59, 0.65]}\\
\cmidrule{5-6}
  &   &   &   & \Block{1-1}{Medulloblastoma, WNT-activated} & \Block{1-1}{0.70$\pm$0.20\\ [0.67, 0.74]}\\
\cmidrule{3-6}
  &   & \Block{1-1}{Other CNS embryonal tumors} & \Block{1-1}{0.28$\pm$0.21\\ [0.24, 0.31]} & \Block{1-1}{Atypical teratoid/rhabdoid tumor} & \Block{1-1}{0.73$\pm$0.18\\ [0.70, 0.76]}\\
\cmidrule{3-6}
  &   & \Block{1-1}{Embryonal tumors NOS} & \Block{1-1}{0.32$\pm$0.25\\ [0.28, 0.36]} & \Block{1-1}{/} & \Block{1-1}{/}\\
\cmidrule{1-6}
\Block{1-1}{Tumors of the sellar region} & \Block{1-1}{0.78$\pm$0.19\\ [0.75, 0.81]} & \Block{1-1}{Adamantinomatous craniopharyngioma} & \Block{1-1}{0.90$\pm$0.12\\ [0.89, 0.92]} & \Block{1-1}{/} & \Block{1-1}{/}\\
\cmidrule{1-6}
\Block{1-1}{Meningiomas} & \Block{1-1}{0.64$\pm$0.19\\ [0.60, 0.67]} & \Block{1-1}{Meningioma} & \Block{1-1}{0.65$\pm$0.22\\ [0.61, 0.69]} & \Block{1-1}{/} & \Block{1-1}{/}\\
\cmidrule{1-6}
\Block{1-1}{Germ cell tumors} & \Block{1-1}{0.40$\pm$0.25\\ [0.36, 0.44]} & \Block{1-1}{/} & \Block{1-1}{/} & \Block{1-1}{/} & \Block{1-1}{/}\\
\cmidrule{1-6}
\Block{1-1}{Choroid plexus tumors} & \Block{1-1}{0.79$\pm$0.15\\ [0.77, 0.82]} & \Block{1-1}{/} & \Block{1-1}{/} & \Block{1-1}{/} & \Block{1-1}{/}\\
\cmidrule{1-6}
\Block{1-1}{Mesenchymal, non-meningothelial tumors} & \Block{1-1}{0.31$\pm$0.26\\ [0.27, 0.35]} & \Block{1-1}{/} & \Block{1-1}{/} & \Block{1-1}{/} & \Block{1-1}{/}\\
\bottomrule
\end{NiceTabular}
\end{table}

\clearpage

\subsubsection*{Classification results using balanced tissue area}
\begin{table}[ht]
\small
\centering
\caption{Classification performance when training on data from all contributing sites and using balanced tissue areas. Metrics are presented as mean and standard deviation with 95\% confidence intervals (CI) computed over 150 model runs. Abbreviations: ABMIL, attention-based multiple-instance learning; CLAM, clustering-constrained attention multiple instance-learning, MCC, Matthew's correlation coefficient; UNI and CONCH, vision transformer foundation models pretrained on histopathology data.}\label{tab:classification_results_balanced_tissue_area}
\begin{NiceTabular}{p[c]{19mm} p[c]{19mm} p[c]{19mm} p[c]{19mm} m[c]{19mm} p[c]{19mm} p[c]{19mm}}[]
\toprule
\Block{1-1}{Classification\\granularity} & \Block{1-1}{Instance feature\\extractor} & \Block{1-1}{MIL aggregation\\method} & \Block{1-1}{MCC\\mean$\pm$std\\ [95\% CI]} & \Block{1-1}{Balanced\\accuracy\\ mean$\pm$std\\ [95\% CI]} & \Block{1-1}{F1 score\\ mean$\pm$std\\ [95\% CI]} & \Block{1-1}{AUROC\\ mean$\pm$std\\ [95\% CI]} \\
\midrule
\Block{4-1}{Tumor Category} & \Block{2-1}{CONCH} & \Block{1-1}{ABMIL} & \Block{1-1}{0.55$\pm$0.07\\ [0.54, 0.56]} & \Block{1-1}{0.62$\pm$0.06\\ [0.61, 0.63]} & \Block{1-1}{0.50$\pm$0.05\\ [0.49, 0.51]} & \Block{1-1}{0.89$\pm$0.03\\ [0.88, 0.89]}\\
\cmidrule{3-7}
  &   & \Block{1-1}{CLAM} & \Block{1-1}{0.54$\pm$0.07\\ [0.53, 0.55]} & \Block{1-1}{0.62$\pm$0.07\\ [0.61, 0.63]} & \Block{1-1}{0.49$\pm$0.05\\ [0.48, 0.50]} & \Block{1-1}{0.89$\pm$0.03\\ [0.88, 0.89]}\\
\cmidrule{2-7}
  & \Block{2-1}{UNI} & \Block{1-1}{ABMIL} & \Block{1-1}{0.51$\pm$0.09\\ [0.50, 0.52]} & \Block{1-1}{0.59$\pm$0.07\\ [0.58, 0.61]} & \Block{1-1}{0.48$\pm$0.06\\ [0.47, 0.49]} & \Block{1-1}{0.88$\pm$0.03\\ [0.88, 0.89]}\\
\cmidrule{3-7}
  &   & \Block{1-1}{CLAM} & \Block{1-1}{0.45$\pm$0.09\\ [0.44, 0.46]} & \Block{1-1}{0.54$\pm$0.08\\ [0.53, 0.55]} & \Block{1-1}{0.42$\pm$0.06\\ [0.41, 0.43]} & \Block{1-1}{0.86$\pm$0.04\\ [0.85, 0.87]}\\
\cmidrule{1-7}
\Block[fill=gray!10]{4-1}{Tumor Family} & \Block[fill=gray!10]{2-1}{CONCH} & \Block[fill=gray!10]{1-1}{ABMIL} & \Block[fill=gray!10]{1-1}{0.38$\pm$0.07\\ [0.37, 0.39]} & \Block[fill=gray!10]{1-1}{0.48$\pm$0.05\\ [0.47, 0.49]} & \Block[fill=gray!10]{1-1}{0.41$\pm$0.05\\ [0.40, 0.42]} & \Block[fill=gray!10]{1-1}{0.86$\pm$0.02\\ [0.85, 0.86]}\\
\cmidrule{3-7}
  &   & \Block[fill=gray!10]{1-1}{CLAM} & \Block[fill=gray!10]{1-1}{0.36$\pm$0.08\\ [0.35, 0.37]} & \Block[fill=gray!10]{1-1}{0.46$\pm$0.06\\ [0.45, 0.47]} & \Block[fill=gray!10]{1-1}{0.40$\pm$0.06\\ [0.39, 0.41]} & \Block[fill=gray!10]{1-1}{0.85$\pm$0.03\\ [0.84, 0.85]}\\
\cmidrule{2-7}
  & \Block[fill=gray!10]{2-1}{UNI} & \Block[fill=gray!10]{1-1}{ABMIL} & \Block[fill=gray!10]{1-1}{0.37$\pm$0.06\\ [0.36, 0.38]} & \Block[fill=gray!10]{1-1}{0.48$\pm$0.06\\ [0.47, 0.49]} & \Block[fill=gray!10]{1-1}{0.41$\pm$0.05\\ [0.40, 0.42]} & \Block[fill=gray!10]{1-1}{0.84$\pm$0.02\\ [0.84, 0.84]}\\
\cmidrule{3-7}
  &   & \Block[fill=gray!10]{1-1}{CLAM} & \Block[fill=gray!10]{1-1}{0.33$\pm$0.07\\ [0.32, 0.34]} & \Block[fill=gray!10]{1-1}{0.45$\pm$0.06\\ [0.44, 0.46]} & \Block[fill=gray!10]{1-1}{0.37$\pm$0.05\\ [0.36, 0.38]} & \Block[fill=gray!10]{1-1}{0.82$\pm$0.03\\ [0.82, 0.83]}\\
\cmidrule{1-7}
\Block{4-1}{Tumor Type} & \Block{2-1}{CONCH} & \Block{1-1}{ABMIL} & \Block{1-1}{0.45$\pm$0.07\\ [0.44, 0.46]} & \Block{1-1}{0.53$\pm$0.05\\ [0.52, 0.54]} & \Block{1-1}{0.47$\pm$0.05\\ [0.46, 0.48]} & \Block{1-1}{0.90$\pm$0.02\\ [0.90, 0.90]}\\
\cmidrule{3-7}
  &   & \Block{1-1}{CLAM} & \Block{1-1}{0.43$\pm$0.07\\ [0.42, 0.44]} & \Block{1-1}{0.53$\pm$0.06\\ [0.52, 0.54]} & \Block{1-1}{0.46$\pm$0.06\\ [0.45, 0.47]} & \Block{1-1}{0.89$\pm$0.02\\ [0.89, 0.89]}\\
\cmidrule{2-7}
  & \Block{2-1}{UNI} & \Block{1-1}{ABMIL} & \Block{1-1}{0.43$\pm$0.06\\ [0.42, 0.44]} & \Block{1-1}{0.52$\pm$0.06\\ [0.51, 0.53]} & \Block{1-1}{0.46$\pm$0.06\\ [0.45, 0.47]} & \Block{1-1}{0.88$\pm$0.02\\ [0.88, 0.88]}\\
\cmidrule{3-7}
  &   & \Block{1-1}{CLAM} & \Block{1-1}{0.40$\pm$0.07\\ [0.39, 0.41]} & \Block{1-1}{0.49$\pm$0.06\\ [0.48, 0.50]} & \Block{1-1}{0.43$\pm$0.06\\ [0.42, 0.44]} & \Block{1-1}{0.86$\pm$0.03\\ [0.86, 0.87]}\\
\bottomrule
\end{NiceTabular}
\end{table}

\begin{table}[ht]
\small
\centering
\caption{Class-wise F1 scores obtained from the best model (UNI instance-level features and ABMIL aggregation) for all the classification tasks when using balanced tissue areas. F1 scores are presented as mean and standard deviation with 95\% confidence intervals (CI) computed over 150 model runs.}\label{tab:f1_score_balanced_tissue_area}
\begin{NiceTabular}{m[c]{31mm} m[c]{16mm} m[c]{31mm} m[c]{16mm} m[c]{31mm} m[c]{16mm}}[]
\toprule
\Block{1-2}{\textbf{Tumor Category}} & & \Block{1-2}{\textbf{Tumor Family}} & & \Block{1-2}{\textbf{Tumor Type}} \\
\Block{1-1}{\textbf{Class}} & \Block{1-1}{\textbf{F1 score}\\mean$\pm$std\\ [95\% CI]} & \Block{1-1}{\textbf{Class}} & \Block{1-1}{\textbf{F1 score}\\mean$\pm$std\\ [95\% CI]} & \Block{1-1}{\textbf{Class}} & \Block{1-1}{\textbf{F1 score}\\mean$\pm$std\\ [95\% CI]} \\
\midrule
\Block{7-1}{Gliomas, glioneuronal tumors, and neuronal tumors} & \Block{7-1}{0.77$\pm$0.09\\ [0.76, 0.79]} & \Block{1-1}{Circumscribed astrocytic gliomas} & \Block{1-1}{0.50$\pm$0.18\\ [0.47, 0.53]} & \Block{1-1}{Pilocytic astrocytoma} & \Block{1-1}{0.63$\pm$0.13\\ [0.60, 0.65]}\\
\cmidrule{3-6}
  &   & \Block{2-1}{Glioneuronal and neuronal tumors} & \Block{2-1}{0.36$\pm$0.16\\ [0.33, 0.38]} & \Block{1-1}{Ganglioglioma} & \Block{1-1}{0.33$\pm$0.13\\ [0.31, 0.35]}\\
\cmidrule{5-6}
  &   &   &   & \Block{1-1}{Dysembryoplastic neuroepithelial tumor} & \Block{1-1}{0.39$\pm$0.15\\ [0.37, 0.42]}\\
\cmidrule{3-6}
  &   & \Block{2-1}{Ependymal tumors} & \Block{2-1}{0.52$\pm$0.14\\ [0.50, 0.55]} & \Block{1-1}{Ependymoma grade 1-2} & \Block{1-1}{0.29$\pm$0.13\\ [0.26, 0.31]}\\
\cmidrule{5-6}
  &   &   &   & \Block{1-1}{Ependymoma grade 3} & \Block{1-1}{0.56$\pm$0.16\\ [0.54, 0.59]}\\
\cmidrule{3-6}
  &   & \Block{1-1}{Adult-type diffuse gliomas} & \Block{1-1}{0.16$\pm$0.11\\ [0.14, 0.18]} & \Block{1-1}{Glioblastoma} & \Block{1-1}{0.27$\pm$0.14\\ [0.25, 0.30]}\\
\cmidrule{3-6}
  &   & \Block{1-1}{Pediatric-type diffuse high-grade gliomas} & \Block{1-1}{0.11$\pm$0.09\\ [0.10, 0.13]} & \Block{1-1}{/} & \Block{1-1}{/}\\
\cmidrule{1-6}
\Block{4-1}{Embryonal tumors} & \Block{4-1}{0.73$\pm$0.09\\ [0.72, 0.75]} & \Block{2-1}{Medulloblastoma} & \Block{2-1}{0.59$\pm$0.19\\ [0.55, 0.62]} & \Block{1-1}{Medulloblastoma, non-WNT/non-SHH} & \Block{1-1}{0.67$\pm$0.13\\ [0.65, 0.69]}\\
\cmidrule{5-6}
  &   &   &   & \Block{1-1}{Medulloblastoma, WNT-activated} & \Block{1-1}{0.48$\pm$0.18\\ [0.45, 0.50]}\\
\cmidrule{3-6}
  &   & \Block{1-1}{Other CNS embryonal tumors} & \Block{1-1}{0.29$\pm$0.13\\ [0.27, 0.31]} & \Block{1-1}{Atypical teratoid/rhabdoid tumor} & \Block{1-1}{0.49$\pm$0.15\\ [0.46, 0.51]}\\
\cmidrule{3-6}
  &   & \Block{1-1}{Embryonal tumors NOS} & \Block{1-1}{0.17$\pm$0.14\\ [0.15, 0.20]} & \Block{1-1}{/} & \Block{1-1}{/}\\
\cmidrule{1-6}
\Block{1-1}{Tumors of the sellar region} & \Block{1-1}{0.77$\pm$0.12\\ [0.76, 0.79]} & \Block{1-1}{Adamantinomatous craniopharyngioma} & \Block{1-1}{0.87$\pm$0.16\\ [0.84, 0.89]} & \Block{1-1}{/} & \Block{1-1}{/}\\
\cmidrule{1-6}
\Block{1-1}{Meningiomas} & \Block{1-1}{0.28$\pm$0.12\\ [0.26, 0.30]} & \Block{1-1}{Meningioma} & \Block{1-1}{0.41$\pm$0.15\\ [0.38, 0.43]} & \Block{1-1}{/} & \Block{1-1}{/}\\
\cmidrule{1-6}
\Block{1-1}{Germ cell tumors} & \Block{1-1}{0.13$\pm$0.11\\ [0.11, 0.15]} & \Block{1-1}{/} & \Block{1-1}{/} & \Block{1-1}{/} & \Block{1-1}{/}\\
\cmidrule{1-6}
\Block{1-1}{Choroid plexus tumors} & \Block{1-1}{0.60$\pm$0.17\\ [0.57, 0.63]} & \Block{1-1}{/} & \Block{1-1}{/} & \Block{1-1}{/} & \Block{1-1}{/}\\
\cmidrule{1-6}
\Block{1-1}{Mesenchymal, non-meningothelial tumors} & \Block{1-1}{0.16$\pm$0.15\\ [0.13, 0.18]} & \Block{1-1}{/} & \Block{1-1}{/} & \Block{1-1}{/} & \Block{1-1}{/}\\
\bottomrule
\end{NiceTabular}
\end{table}

\end{document}